\documentclass{article} 
\usepackage[preprint]{colm2026_conference}

\usepackage{microtype}
\usepackage{hyperref}
\usepackage{url}
\usepackage{booktabs}

\usepackage{lineno}

\definecolor{darkblue}{rgb}{0, 0, 0.5}
\hypersetup{colorlinks=true, citecolor=darkblue, linkcolor=darkblue, urlcolor=darkblue}

\usepackage{subcaption}
\usepackage{wrapfig}

\usepackage{tcolorbox}
\usepackage{xcolor}
\usepackage{graphicx}
\usepackage{caption}
\tcbuselibrary{skins}
\tcbuselibrary{breakable}
\usepackage{multirow}
\usepackage{makecell}
\usepackage{booktabs}
\usepackage{cleveref}

\usepackage{enumitem}

\usepackage[colorinlistoftodos,textwidth=3cm]{todonotes}

\usepackage{soul}

\usepackage{amssymb}
\usepackage{newunicodechar}
\newunicodechar{☆}{$\star$}

\usepackage{xcolor}
\usepackage{pifont}
\newcommand{\cmark}{\textcolor{green!60!black}{\ding{51}}}
\newcommand{\xmark}{\textcolor{red!70!black}{\ding{55}}}
\definecolor{myblue}{RGB}{0,90,170}
\newcommand{\qmark}{{\textcolor{myblue}{\textbf{\normalsize ?}}}}

\definecolor{bblue}{RGB}{120,200,248}
\definecolor{bpetrol}{RGB}{0,111,151}
\definecolor{bpink}{RGB}{247,168,217}
\definecolor{bpurple}{RGB}{184,146,245}
\definecolor{bred}{RGB}{255,102,122}

\title{SEQUOR: A Multi-Turn Benchmark for Realistic Constraint \\ Following}

\author{%
Beatriz Canaverde\textsuperscript{1,2}, 
Duarte M. Alves\textsuperscript{1,2},
José Pombal\textsuperscript{1,2,3}, \\
\textbf{Giuseppe Attanasio\textsuperscript{1} \&}
\textbf{André F. T. Martins}\textsuperscript{1,2,4,5} \\
\\
\textsuperscript{1}Instituto de Telecomunicações, 
\textsuperscript{2}Instituto Superior Técnico, Universidade de Lisboa \\
\textsuperscript{3}Sword Health, 
\textsuperscript{4}TransPerfect, 
\textsuperscript{5}ELLIS Unit Lisbon \\
\texttt{beatriz.canaverde@tecnico.ulisboa.pt}
}

\begin{document}

\ifcolmsubmission
\linenumbers
\fi

\maketitle

\begin{abstract}
In a conversation, a helpful assistant must reliably follow user directives, even as they refine, modify, or contradict earlier requests. Yet most instruction-following benchmarks focus on single-turn or short multi-turn scenarios, leaving open how well models handle long-horizon instruction-following tasks. To bridge this gap, we present \textsc{Sequor}, an automatic benchmark for evaluating constraint adherence in long multi-turn conversations. \textsc{Sequor} consists of simulated persona-driven interactions built with constraints extracted from real-world conversations. Our results show that even when following a single constraint, instruction-following accuracy consistently decreases as the conversation grows longer, with drops exceeding $11\%$. This decline becomes larger when models have to follow multiple constraints simultaneously, reducing their accuracy by over $40\%$. In scenarios where constraints are added or replaced at arbitrary points of the conversation, model accuracy decreases by more than $9\%$. Taken together, our results reveal that current models still struggle to follow user instructions in multi-turn conversations, and provide a way for better measuring instruction-following capabilities in assistants.\footnote{Code and data are available at: \url{https://github.com/deep-spin/SEQUOR}}
\end{abstract}


\section{Introduction}

Digital assistants must consistently adhere to user instructions across the full course of a conversation. Because users often refine, modify, or even contradict earlier requests~\citep{zheng2024lmsyschatm,zhao2024wildchat,bai-etal-2024-mt,10.5555/3692070.3692401,laban2025llmslostmultiturnconversation}, this skill requires following instructions that may change over many turns. 
This makes evaluation challenging, as current instruction-following benchmarks evaluate models in single-turn or short multi-turn settings, using either programmatically verifiable or LLM-generated instructions that are not representative of real-world use cases~\citep{10.5555/3666122.3668142,zhou2023instructionfollowingevaluationlargelanguage,qin-etal-2024-infobench,he2024multiifbenchmarkingllmsmultiturn,kwan-etal-2024-mt,jiang-etal-2024-followbench,dussolle-etal-2025-ifeval,pyatkin2025generalizingverifiableinstructionfollowing,xia-etal-2024-fofo,bai-etal-2024-mt,jiang-etal-2024-followbench,deshpande-etal-2025-multichallenge}.
This context raises the question of how robustly modern large language models (LLMs) follow instructions in open-domain interactions that span multiple turns.

\begin{figure*}[ht!]
\centering
\includegraphics[width=1.0\textwidth]{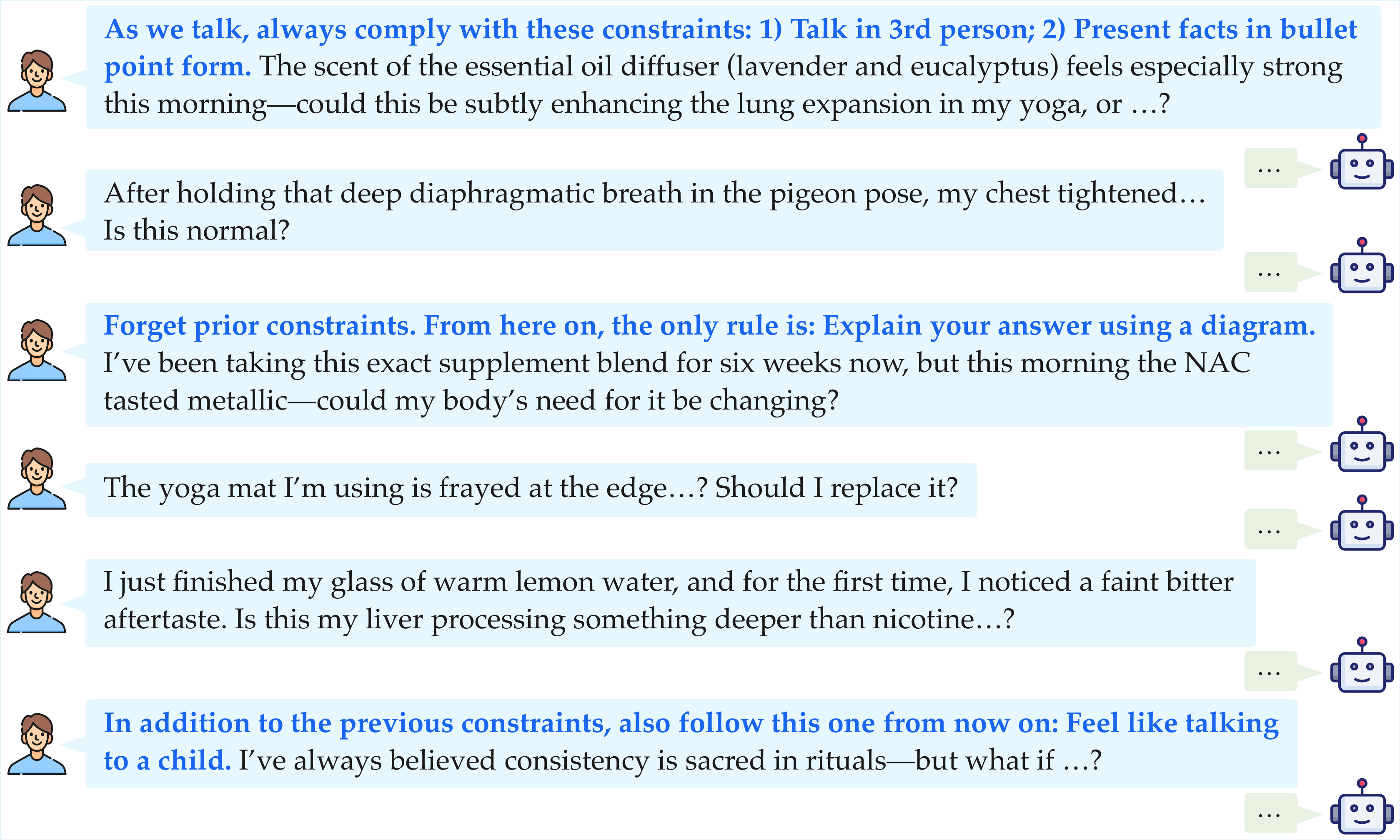}
\caption{Example snippet of a conversation from \textsc{Sequor}.}
\label{fig:example-snippet-conversation}
\end{figure*}

To bridge this gap, we present \textsc{Sequor},\footnote{\textsc{Sequor} is a Latin verb meaning ``I follow.''} an automatic benchmark that measures instruction-following capabilities in multi-turn open-domain conversations. \textsc{Sequor} is grounded in two core principles. First, constraints must be realistic, broadly applicable, challenging, and verifiable~(\S\ref{sec:collecting-realistic-constraints-wild}). Second, interactions must span many turns, allowing constraints to accumulate or be replaced in a credible way~(\S\ref{sequor-simulating-multi-turn-conversations}; see \Cref{fig:example-snippet-conversation}). Accordingly, \textsc{Sequor} comprises simulated persona-driven interactions built from constraints extracted from real-world conversations. It systematically varies how and when constraints are introduced, from initial constraints that remain unchanged over the conversation, to constraints that are incrementally added or replaced over time.
As a result, it captures a broad range of long-horizon instruction-following scenarios.

We evaluated several modern LLMs on \textsc{Sequor}, revealing consistent limitations in long-horizon instruction-following (\S\ref{sec:experiments}). Across all regimes, constraint-following accuracy degrades as conversations become longer. Even when following a single constraint, accuracy drops by more than $11\%$ between the first and last turns. The decline is substantially larger, exceeding $38\%$, when multiple constraints need to be satisfied simultaneously, and is most pronounced, with losses above $40\%$, when constraints are introduced sequentially rather than all at once. Resetting constraints mid-conversation allows models to recover their initial performance, although accuracy declines more rapidly afterward. Finally, when constraints are randomly added or replaced at arbitrary points in the conversation, accuracy decreases by more than $9\%$. Taken together, these results show that current models still struggle to reliably follow user directives over long multi-turn interactions.

Our main contributions are summarized as follows:
\begin{itemize}[leftmargin=2em]
    \item We propose an automated pipeline for extracting and curating broadly applicable, non-trivial, and objectively verifiable constraints from real-world conversations~(\S\ref{sec:collecting-realistic-constraints-wild}).\footnote{Accompanying our benchmark, we will also release the curated pool of $1{,}446$ realistic constraints.}
    \item We introduce \textsc{Sequor}, a multi-turn benchmark for constraint-following, comprising $1,400$ conversations of $50$ turns each (\S\ref{sequor-simulating-multi-turn-conversations}). It systematically varies how constraints are introduced in a conversation, ranging from static initial constraints to incremental addition and replacement over time.
    \item We empirically demonstrate that current LLMs experience substantial degradation in constraint adherence as the number of turns increases and constraints accumulate (\S\ref{sec:experiments}).
\end{itemize}


\section{Collecting Realistic Constraints in the Wild}
\label{sec:collecting-realistic-constraints-wild}

\begin{figure*}[ht!]
\centering
\includegraphics[width=1.0\textwidth]{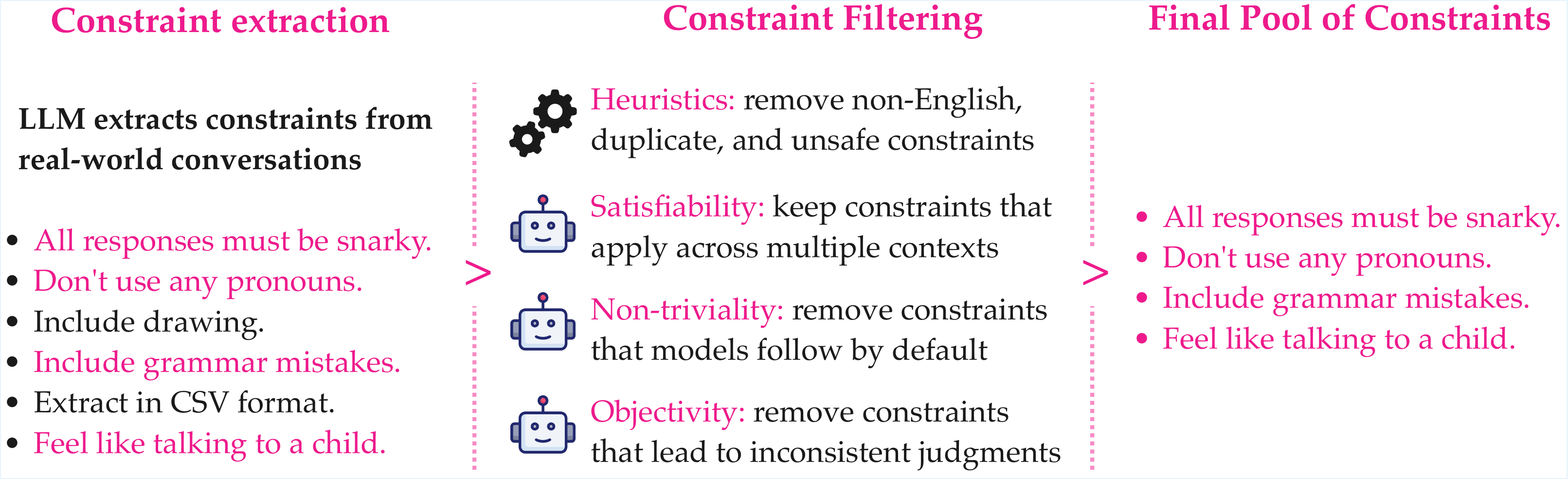}
\caption{Pipeline to collect constraints from real-world conversations.}
\label{fig:pipeline-collecting-constraints-in-the-wild}
\end{figure*}

To evaluate instruction-following, we test whether an assistant adheres to constraints shaping its output's form, style, or structure. We collect constraints from real-world conversational data and automatically filter them using heuristics and LLMs-as-judges \citep{10.5555/3666122.3668142}. Our pipeline, shown in \Cref{fig:pipeline-collecting-constraints-in-the-wild}, produced a pool of $1{,}446$ realistic constraints.

\paragraph{Extracting constraints.}
Starting from lmsys-chat-1m \citep{zheng2024lmsyschatm}, a dataset of real-world user-assistant conversations,\footnote{It contains $1$M conversations collected from Chatbot Arena and Vicuna demo (April-August $2023$).} we prompt Qwen3-Next-80B-A3B-Instruct-FP8 \citep{qwen2.5-1m,qwen3technicalreport} with each English conversation to extract all constraints expressed in the user prompts.
Following \cite{qin-etal-2024-infobench}, we categorize the constraints into four main categories: linguistic guidelines, style rules, format specifications, and number limitations.\footnote{\cite{qin-etal-2024-infobench} consider an additional fifth category, content constraints, which define the topics or details that should be addressed in the LLM's response. We exclude this category from our experiments because: 1) we found them harder to extract from conversations, and 2) they are less aligned with our goal of collecting constraints that are broadly applicable across diverse contexts.}

\paragraph{Automatic filtering.}
Using the Datatrove library \citep{penedo2024datatrove}, we remove non-English constraints with the fastText language identification model \citep{joulin2016fasttextzipcompressingtextclassification,joulin2016bagtricksefficienttext}, discarding all entries with a confidence score below $0.65$. We then remove similar constraints using MinHash deduplication with $50$ buckets, $4$ hashes per bucket, and $3$-grams. Finally, we exclude constraints containing words from the English subset of a predefined list of bad words,\footnote{\url{https://github.com/LDNOOBW/List-of-Dirty-Naughty-Obscene-and-Otherwise-Bad-Words}} or sequences of characters from a custom list.\footnote{Custom list: sex, porn, nud}

\paragraph{Ensuring constraints are satisfiable.}
We retain only constraints that are satisfiable across multiple contexts. For example, \textit{``Answer in at most $100$ words.''} is broadly applicable, whereas \textit{``Your answer must include a Python function definition.''} is only meaningful for programming-related tasks.
To identify satisfiable constraints, we pair each one with $100$ randomly sampled tasks, and evaluate each pair using various judges and the rubrics presented in~\autoref{table:rubrics-satisfiable-constraints}. A constraint is satisfiable for a given task if a judge assigns positive scores to rubrics \textbf{1}, \textbf{3}, and \textbf{4}, and a negative response to rubric \textbf{2}. To pass this filter, a constraint must be deemed satisfiable by every judge in at least $70\%$ of the analyzed contexts.

\begin{figure*}[t]
\centering
\begin{tcolorbox}[
  enhanced,
  colback=bblue!10, 
  colframe=bblue!100, 
  colbacktitle=bblue!100, 
  coltitle=white,
  fonttitle=\bfseries,
  title={Rubrics used to identify non-satisfiable constraints},
  boxrule=0.8pt,
  left=2mm,right=2mm,top=1mm,bottom=1mm
]
\textbf{1.} Is the constraint actually a restriction or condition that limits how the model should generate its output to the task? \\
\textbf{2.} Does the constraint target a different question, topic, or domain than the task itself? \\
\textbf{3.} Is the constraint applicable to the type of output the task requires? \\
\textbf{4.} Does the constraint fall into one of the following four categories: linguistic guidelines, style rules, format specifications, or number limitations?
\end{tcolorbox}
\caption{Rubrics used by LLM judges to identify non-satisfiable constraints. The complete prompt template, including the definitions of the constraint categories, is shown in \Cref{fig:satisfiable-constraints-prompt-template}.}
\label{table:rubrics-satisfiable-constraints}
\end{figure*}

\begin{figure*}[t!]
\centering
\includegraphics[width=1.0\textwidth]{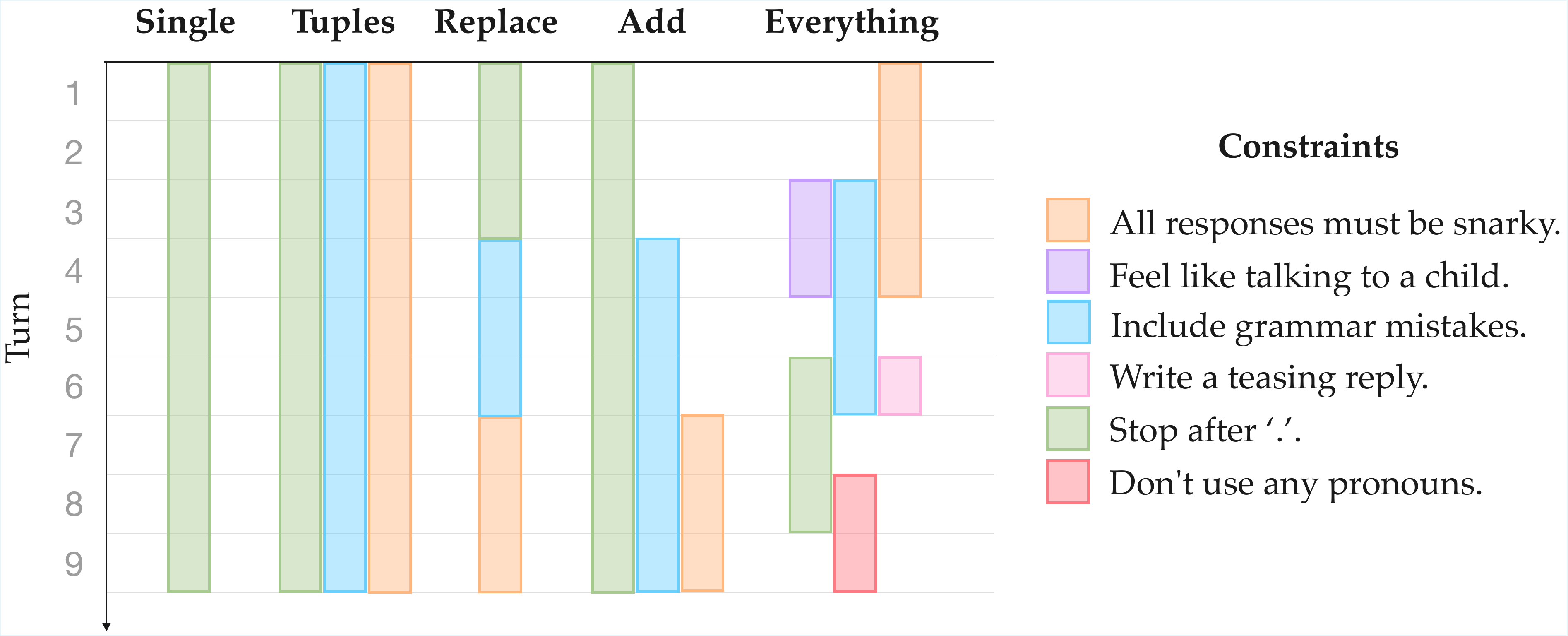}
\caption{\textsc{Sequor} simulates persona-driven interactions, varying how constraints are introduced across five systematic regimes.}
\label{fig:sequor-simulating-multi-turn-conversations}
\end{figure*}

\paragraph{Avoiding trivially satisfiable constraints.}
Although we prioritize broadly applicable constraints, we exclude ones that are likely to be satisfied even when not explicitly specified in the prompt. For example, \textit{``Answer in proper English.''} is typically followed by most language models when responding in English. To identify such trivial constraints, we sample model responses to $100$ tasks without specifying any constraint and test whether the constraint is nevertheless satisfied. A constraint is non-trivial if each judge classifies at least $70\%$ of the responses as not satisfying the constraint.

\paragraph{Removing subjective constraints.} 
Some constraints are subjective and may lead to inconsistent evaluations across judges (e.g., \textit{``Write a creative response.''}). To identify and remove such cases, we pair each constraint with $100$ tasks and sample model responses to these constraint-task pairs. We then use multiple judges to independently assess whether the constraint has been followed in each response. A constraint is non-subjective if all judges agree on the binary judgment in at least $70\%$ of the evaluated task contexts.

For constraint assessment, we use three judges: GPT-oss-120B \citep{openai2025gptoss120bgptoss20bmodel}, Qwen3-235B-A22B-Instruct-2507-FP8 \citep{qwen3technicalreport}, and GLM-4.7-FP8 \citep{5team2025glm45agenticreasoningcoding}. Model responses are generated using four smaller LLMs: Qwen3-4B-Instruct-2507 \citep{qwen3technicalreport}, Llama-3.2-3B-Instruct \citep{grattafiori2024llama3herdmodels}, Gemma3-4B \citep{gemmateam2025gemma3technicalreport}, and Olmo-3-7B-Instruct \citep{olmo2025olmo3}. The $70\%$ threshold balances robustness to contextual variability and constraint diversity. Further analysis and prompt templates are in \Cref{app:constraints-pipeline}.


\section{SEQUOR: Simulating and Evaluating Multi-Turn Conversations}
\label{sequor-simulating-multi-turn-conversations}

From our pool of realistic constraints, we construct \textsc{Sequor}.
In \textsc{Sequor}, user turns are generated from persona profiles and the extracted constraints, and the assistant turns are then evaluated using LLMs-as-a-Judge.

\subsection{Simulating Conversations}

\textsc{Sequor} consists of sequences of user turns that form multi-turn conversations with an assistant. Each turn specifies a task---an action or goal for the assistant to perform---and optionally updates the constraints the assistant must follow. To emulate realistic conditions, we design five test scenarios, use persona profiles, and control for conflicting constraints.

\paragraph{Test sets.}
\textsc{Sequor} includes five test sets built from a fixed collection of user-turn sequences, differing only in the constraints provided to the assistant (see \Cref{fig:sequor-simulating-multi-turn-conversations}). For each test set, the constraints are randomly sampled from our pool and introduced at specific turns using predefined templates (see \S\ref{app:templates-introducing-constraints}). The five sets are defined as follows:
\begin{itemize}
    \item \textbf{Single.} One constraint is given in the first turn and must be followed thereafter.
    \item \textbf{Tuples.} Three constraints are given in the first turn and must be followed thereafter.
    \item \textbf{Replace.} A constraint is given in the first turn and replaced every $x$ turns. Each constraint must be followed until it is replaced. We consider $x=5$ and $x=10$.
    \item \textbf{Add.} A constraint is given in the first turn, and additional ones are added every $x$ turns, up to a maximum of three. Constraints accumulate; once introduced, they must be followed thereafter. We consider $x=5$ and $x=10$.
    \item \textbf{Everything.} A mixture of the previous regimes. After a random number of turns (between $1$ and $5$), up to three constraints are given, randomly accumulating with or replacing earlier ones.
\end{itemize}

\paragraph{Tasks.}
We design tasks to simulate interactions between diverse personas and an assistant. We first sample persona profiles from Persona Hub \citep{ge2025scalingsyntheticdatacreation} and use Qwen3-Next-80B-A3B-Instruct-FP8 \citep{qwen2.5-1m,qwen3technicalreport} to generate a sequence of daily activities tailored to each persona's profession, interests, and lifestyle. Then, given a persona and an activity, the same model generates open-ended questions that the persona might naturally ask an assistant in that scenario. This process yields ordered sequences of questions that simulate a natural flow of interactions. 
See \Cref{fig:example-snippet-conversation} for an example.
Prompt templates are given in Appendix~\ref{app:synthetic-task-generation}. The final dataset contains $200$ personas, each with $50$ associated tasks.

\paragraph{Tuples of constraints.}
For evaluation scenarios in which the assistant must satisfy multiple constraints  simultaneously, we must identify tuples of compatible constraints (e.g., \textit{``Write your answer entirely in capital letters.''} vs.\ \textit{``Write your answer entirely in lowercase letters.''} are not compatible). To do so, we first sample tuples of three constraints from our pool and pair them with $100$ tasks. Then, we apply two filtering criteria to ensure the tuples contain compatible constraints. First, we only retain tuples for which all judges agree that its constraints are non-conflicting for $70\%$ of the tasks. Second, after sampling one response for each tuple-task pair, we only retain tuples for which all judges agree that at least one answer satisfies all constraints for $70\%$ of the tasks. We use three judges: GPT-oss-120B \citep{openai2025gptoss120bgptoss20bmodel}, Qwen3-235B-A22B-Instruct-2507-FP8 \citep{qwen3technicalreport}, and GLM-4.7-FP8 \citep{5team2025glm45agenticreasoningcoding}. Responses are sampled from GPT-oss-20B \citep{openai2025gptoss120bgptoss20bmodel}, Llama-3.1-8B-Instruct \citep{grattafiori2024llama3herdmodels}, Gemma3-27B \citep{gemmateam2025gemma3technicalreport}, and Olmo-3.1-32B-Instruct \citep{olmo2025olmo3}. Prompt templates are provided in \Cref{app:creating-tuples-constraints}.

\subsection{Evaluation with LLM-as-a-Judge}
\label{sec:choose-judge}

All assistant responses in \textsc{Sequor} are evaluated independently on a turn-by-turn basis using an LLM-as-a-judge \citep{10.5555/3666122.3668142,gu2025surveyllmasajudge}. For each turn, the judge determines whether the response satisfies each active constraint individually. We verify that models can reliably judge constraint adherence.

\paragraph{Data.}
We pair $500$ constraints and $500$ tasks randomly sampled from our pool. For each pair, we prompt proprietary models to generate two answers in a single-turn setup: (i) one answer explicitly instructed to satisfy the constraint, and (ii) one answer explicitly instructed to violate the constraint. We manually inspect a subset to verify that answers labeled as satisfying, or violating, the constraint behave as intended. We treat these as gold responses.

\paragraph{Evaluation.}
We evaluate four candidate judges by asking them to determine whether each answer satisfies its associated constraint (prompt template in \Cref{fig:judge-single-constraint-prompt-template}). We report the percentage of correct and incorrect classifications for: (i) gold answers satisfying constraints (Gold: Yes), (ii) gold answers violating constraints (Gold: No), and (iii) overall.

\paragraph{Models.}
Gold responses are generated using Gemini 3 Flash Preview \citep{gemini-3-flash}\footnote{At the time of testing, this model ranked among the top-performing systems on the IFBench leaderboard: \url{https://artificialanalysis.ai/evaluations/ifbench} (accessed February 25, 2026).} and GPT-5.2 \citep{introducing-gpt-5.2}, producing two independent gold sets. We compare the following models as judges: Qwen3-235B-A22B-Instruct-2507-FP8, Qwen3-235B-A22B-Thinking-2507-FP8 \citep{qwen3technicalreport}, GPT-oss-120B \citep{openai2025gptoss120bgptoss20bmodel}, and GLM-4.7-FP8 \citep{5team2025glm45agenticreasoningcoding}.

\begin{table*}[t]
\centering
\small
\begin{tabular}{lccc ccc ccc}
\toprule
& \multicolumn{3}{c}{\textbf{Gold: Yes}} 
& \multicolumn{3}{c}{\textbf{Gold: No}} 
& \multicolumn{3}{c}{\textbf{Overall}} \\

\cmidrule(lr){2-4} 
\cmidrule(lr){5-7} 
\cmidrule(lr){8-10}

\textbf{Model} 

& \cmark & \xmark & \qmark
& \cmark & \xmark & \qmark
& \cmark & \xmark & \qmark \\

\midrule

Qwen3-235B-A22B-Inst. 
& 84.90 & 15.10 & 0 
& 98.10 & 1.70 & 0.20 
& 91.50 & 8.40 & 0.10 \\



Qwen3-235B-A22B-Think. 
& 87.90 & 11.90 & 0.20 
& 98.40 & 1.60 & 0 
& 93.15 & 6.75 & 0.10 \\



GPT-oss-120B 
& 88.30 & 11.50 & 0.20 
& 98.80 & 0.70 & 0.50 
& 93.55 & 6.10 & 0.35 \\



GLM-4.7-FP8 
& 91.10 & 8.90 &  0
& 98.60 & 1.30 & 0.10 
& 94.85 & 5.10 & 0.05 \\


  
\bottomrule
\end{tabular}
\caption{Percentage of correct (\cmark), incorrect (\xmark), and unextractable (\qmark) verdicts of candidate judges in detecting constraint adherence. ``Gold: Yes'' and ``Gold: No'' denote gold responses that satisfy or violate constraints, respectively; ``Overall'' reflects the full evaluation set.}
\label{tab:best-judge-results-avg}
\end{table*}

\paragraph{Results.}
\Cref{tab:best-judge-results-avg} reports the judges' performance averaged across the two gold sets generated by Gemini 3 Flash Preview and GPT-5.2. For all models, detecting violations is easier than confirming correct adherence, and the rate of unextractable verdicts is minimal. GLM-4.7-FP8 performs best on ``Gold: Yes'' and overall. However, we select GPT-oss-120B for our main experiments (\S\ref{sec:experiments}) as it offers competitive performance with greater efficiency.


\section{Experiments}
\label{sec:experiments}

\textsc{Sequor} evaluates models' ability to follow constraints throughout multi-turn conversations. Accordingly, we consider all model responses in a conversation and assess uniquely whether they satisfy all constraints active at each turn.

\subsection{Experimental Setup}

\paragraph{Metrics.} A turn is considered successful if the model's response satisfies all constraints active at that turn. Our main metric is \textbf{per-turn accuracy}, defined as the percentage of successful responses at each turn, averaged across conversations.

\paragraph{Evaluated Models.} We benchmark $10$ open-weight models from different families and sizes: Qwen3-4B-Inst, Qwen3-30B-A3B-Inst, Qwen3-235B-A22B-Inst\footnote{We use the Instruct-2507 versions of the models, and FP8 quantization for the largest two.} \citep{qwen3technicalreport}, Gemma3 4B, 12B, and 27B \citep{gemmateam2025gemma3technicalreport}, Llama-3.3-70B-Inst \citep{grattafiori2024llama3herdmodels}, GPT-oss 20B and 120B \citep{openai2025gptoss120bgptoss20bmodel}, and GLM-4.7-Flash \citep{5team2025glm45agenticreasoningcoding}. We also report results for the proprietary Gemini 3.1 Flash Lite \citep{gemini-3.1-flash-lite}. All models are run with their default configurations. We implement a sliding-window approach when the conversation history exceeds a model's context length.


\subsection{Regime Analysis}

\begin{figure*}[t!]
\centering
\includegraphics[width=1.0\textwidth]{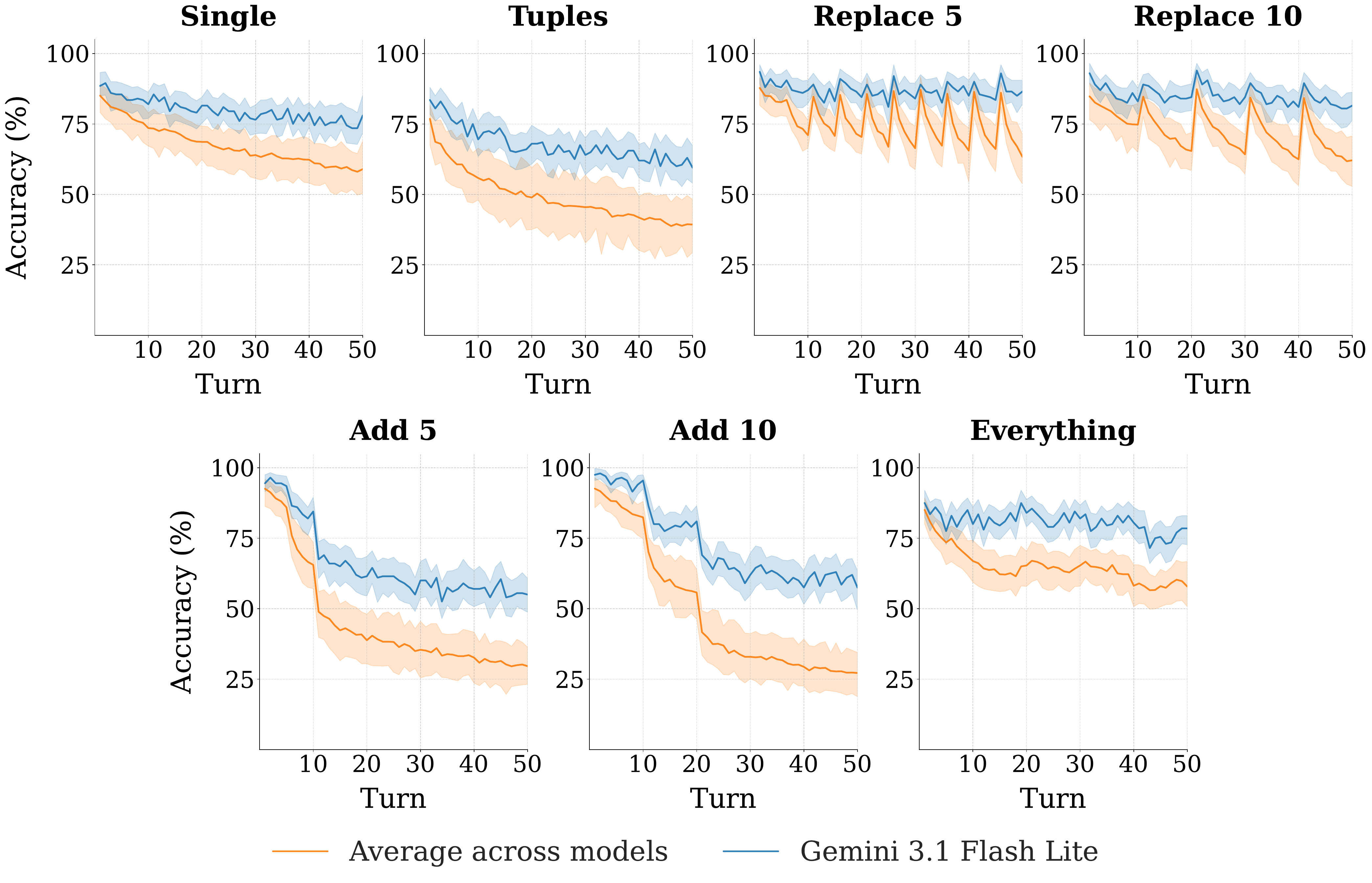}
\caption{Per-turn accuracy across regimes. Shaded regions indicate $95\%$ bootstrap confidence intervals.}
\label{fig:per_turn_accuracy_bootstrap}
\end{figure*}

\begin{figure*}[t!]
\centering
\includegraphics[width=1.0\textwidth]{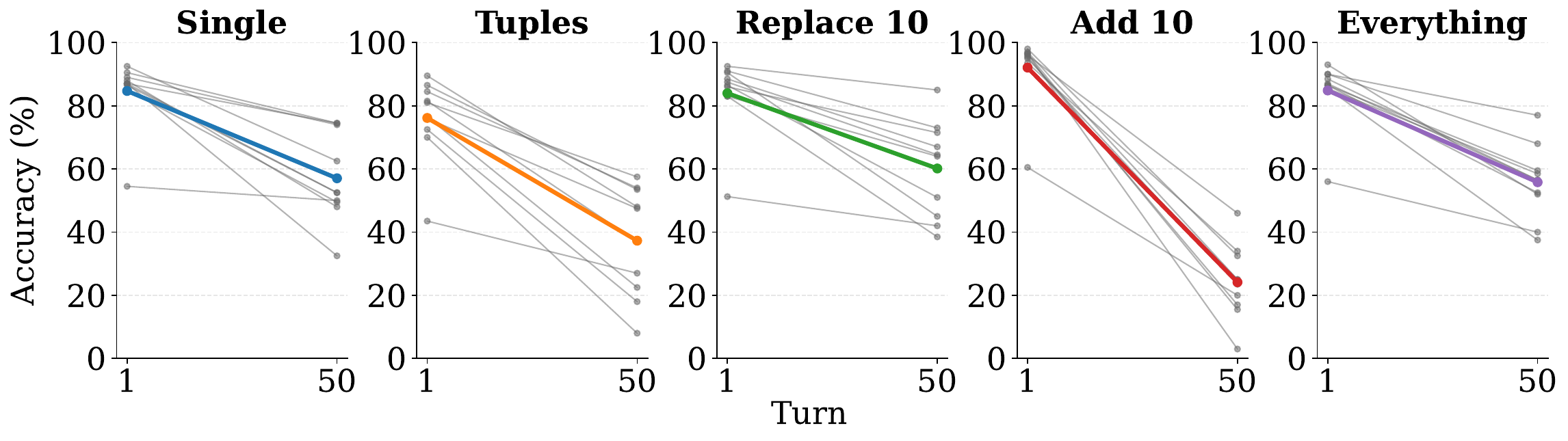}
\caption{Change in per-turn accuracy from turn 1 to turn 50 across regimes. Each gray line corresponds to a model, while the colored lines show the average across models.}
\label{fig:regime_slope_plots}
\end{figure*}

We analyze how constraint adherence evolves throughout conversations for each regime. \Cref{fig:per_turn_accuracy_bootstrap} shows the per-turn accuracy averaged across all models, as well as results for the best-performing model; shaded areas indicate the $95\%$ confidence intervals. \Cref{fig:regime_slope_plots} illustrates the drop in accuracy between the first and last turns for all regimes and models. \textit{Add 5, 10} and \textit{Replace 5, 10} denote the number of turns between each addition or replacement.

\begin{figure*}[t!]
\centering
\includegraphics[width=0.9\textwidth]{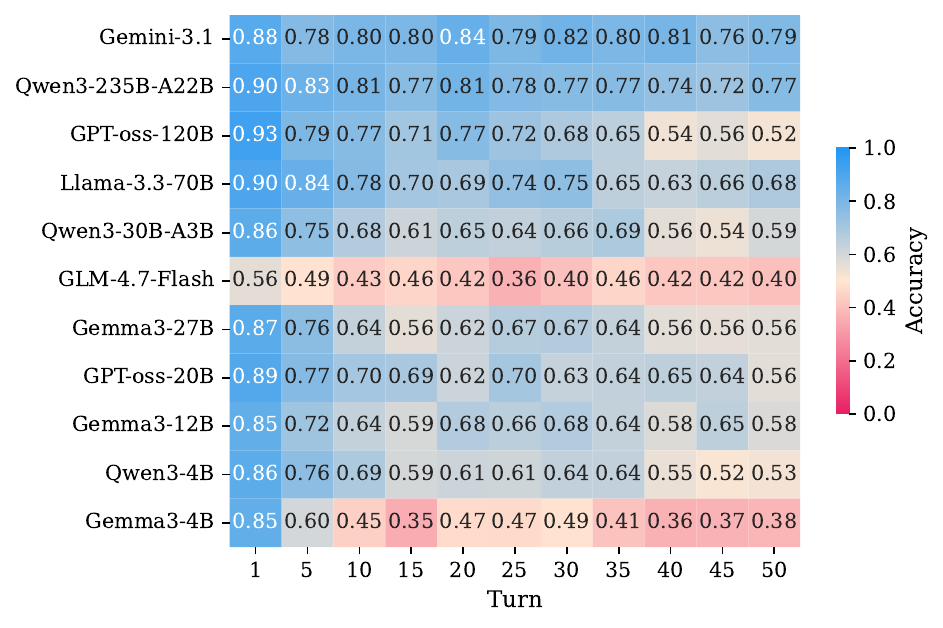}
\caption{Per-turn accuracy for all models in the \textit{Everything} regime.}
\label{fig:per_turn_accuracy_heatmap_everything}
\end{figure*}

\paragraph{\textbf{Performance degrades consistently as conversations progress across all regimes.}}
Even in the simplest regime, where a single constraint is introduced in the first turn and must be followed in all subsequent turns, we observe a consistent decrease in performance over time. The average accuracy in the \textit{Single} regime drops by $26\%$ from the first to the last turn.

\paragraph{\textbf{Models struggle more to follow multiple constraints simultaneously, particularly when these are introduced sequentially over time.}}
Among all regimes, \textit{Tuples} and \textit{Add} exhibit the largest performance drops, with average decreases of $38\%$ and $63\%$, respectively (see \Cref{fig:regime_slope_plots}). \textit{Tuples} starts with the lowest average accuracy and remains the most challenging setting during the first $10$ turns (see \Cref{fig:per_turn_accuracy_bootstrap}). At turn $11$, \textit{Add 5} becomes the lowest-performing regime, and \textit{Add 10} becomes the worst at turn $21$; both shifts coincide with the introduction of a third constraint. In the \textit{Add} regime, accuracy drops noticeably whenever new constraints are presented (turns $6$, $11$ in \textit{Add 5} and turns $11$, $21$ in \textit{Add 10}; see \Cref{fig:per_turn_accuracy_bootstrap}). These results suggest that models struggle more when constraints accumulate over time than when provided all at once, likely due to difficulty retaining earlier constraints or reasoning about their integration.

\paragraph{\textbf{Models tend to recover their initial performance when existing constraints are replaced with new ones.}}
This is evidenced by the sharp accuracy spikes observed in \textit{Replace 5} and \textit{Replace 10} (see \Cref{fig:per_turn_accuracy_bootstrap}), which occur when constraints are replaced (e.g., turns $11$, $21$, $31$, and $41$ in \textit{Replace 10}). At these points, models' average accuracy consistently returns to the high levels observed in the first turn. Interestingly, after each replacement, subsequent turns often exhibit sharper performance declines than those following the first turn.

\paragraph{\textbf{Randomly adding or replacing constraints at arbitrary points in a conversation results in an average performance drop of $27\%$ from the first to the last turn.}}
The \textit{Everything} regime combines all others, balancing easier scenarios (e.g., resetting constraints) with more challenging ones (e.g., accumulating multiple constraints). Consequently, its behavior lies between the easier and harder regimes. Although models occasionally show small improvements between consecutive turns---particularly after constraints are reset---accuracy declines substantially throughout the conversations.

\paragraph{\textbf{Gemini 3.1 Flash Lite, the best-performing model, shows similar overall trends and significant performance losses.}}
To better understand the upper bound of the evaluated systems, we analyze the behavior of the best-performing model across regimes, identified using the Borda count ranking \citep{colombo2022what}. Gemini's overall trends remain consistent with those observed across models. In the \textit{Single}, \textit{Replace}, and \textit{Everything} regimes, accuracy declines by up to $12\%$ over the course of the conversations, although the model still achieves relatively strong results at turn $50$ (see \Cref{fig:per_turn_accuracy_bootstrap}). In contrast, the more challenging \textit{Tuples} and \textit{Add} regimes exhibit much larger losses of $25\%$ and $40\%$, respectively, ending with accuracies below $60\%$ at turn $50$.

\subsection{Model Breakdown Analysis}

In \Cref{fig:per_turn_accuracy_heatmap_everything}, we examine the performance of all models throughout the conversations in the \textit{Everything} regime, measured every five turns. Gemini 3.1 Flash Lite and Qwen3-235B-A22B-Inst are the only models that sustain accuracy above $70\%$ across all turns, although they still experience drops of $9\%$ and $13\%$, respectively, between the first and last turns. In contrast, most other models fall below $60\%$ accuracy at turn $50$. Notably, in this regime, potential context-length limitations should not cause degradation in later turns. Although earlier parts of a conversation may fall outside the context window, the constraints active at each turn are given within at most the previous $15$ turns. These results suggest that maintaining constraint adherence over long interactions remains challenging for current models.


\section{Related Work} 

\subsection{Constraint Following Evaluation}

A large body of work evaluates language models ability to follow explicit output constraints. IFEval \citep{zhou2023instructionfollowingevaluationlargelanguage} introduced manually designed, programmatically verifiable constraints for single-turn evaluation. Subsequent work expanded this framework to multilingual and limited multi-turn settings while preserving rule-based verification \citep{he2024multiifbenchmarkingllmsmultiturn,dussolle-etal-2025-ifeval,pyatkin2025generalizingverifiableinstructionfollowing}. Although this approach enables precise and reproducible evaluation, it restricts constraints to those that can be automatically checked. InFoBench \citep{qin-etal-2024-infobench}, FOFO \citep{xia-etal-2024-fofo}, and FollowBench \citep{jiang-etal-2024-followbench} relax deterministic verification and instead rely on LLM-as-a-judge evaluation. Consequently, they introduce more diverse constraints, using manually written or LLM-generated templates. These settings, however, remain limited to single-turn interactions.

Constraint following has also been studied in moderately longer multi-turn settings. MT-Eval \citep{kwan-etal-2024-mt} and MultiChallenge \citep{deshpande-etal-2025-multichallenge} extend prior work to conversations of $10$–$12$ turns, explicitly evaluating whether models can retain, accumulate, and apply constraints across dialogue turns. MemoryCode \citep{rakotonirina-etal-2025-tools} further scales this paradigm to substantially longer multi-session coding dialogues (up to $40$K tokens). It is, however, restricted to the coding domain and relies on manually crafted, programmatically verifiable constraints.

\textsc{Sequor}, in constrast, evaluates constraint adherence in conversations of $50$ turns, systematically varying how constraints are introduced. It uses constraints derived from real-world datasets, allowing the evaluation of instruction-following in open-domain interactions.

\subsection{Multi-Turn Evaluation}

Beyond constraint following, several benchmarks evaluate general multi-turn conversational abilities. MT-Bench \citep{10.5555/3666122.3668142} introduced a two-turn evaluation framework based on LLM judges without reference answers---an evaluation paradigm we also adopt. Building on this, \citet{bai-etal-2024-mt} propose MT-Bench-101 to evaluate more complex, real-world dialogue phenomena over interactions of up to $6$ turns, assessing $13$ fine-grained abilities categorized under perceptivity, adaptability, and interactivity.

Other benchmarks explore interactive and feedback-driven settings. MINT \citep{wang2024mint}, Meeseeks \citep{wang2025meeseeksfeedbackdriveniterativeselfcorrection}, and related work \citep{laban2025llmslostmultiturnconversation,kim-etal-2025-llm-interviewer} assess models under iterative feedback, self-correction loops, dynamically revealed information, and follow-up questioning. In contrast, \textsc{Sequor} does not provide feedback or allow response revision. Instead, it evaluates whether models consistently follow evolving constraints without external guidance. 

Finally, long-context benchmarks such as LoCoMo \citep{maharana-etal-2024-evaluating}, LongMemEval \citep{wu2025longmemeval}, and PersonaMem \citep{jiang2025know} focus on long-term memory, persona adaptation, and long-range reasoning in extended dialogues. While for \textsc{Sequor} we chose conversations with $50$ turns, this number is configurable, allowing control for the context length as models progressively support longer context sizes.


\section{Conclusion} 

We introduced \textsc{Sequor}, an automatic benchmark for evaluating constraint adherence in long multi-turn conversations. It consists of simulated persona-driven interactions built with constraints extracted from real-world conversations, systematically varying how constraints are introduced. Our experiments reveal key limitations of current LLMs. Instruction-following accuracy declines as conversations grow longer, even when following a single constraint. The degradation becomes substantially larger when multiple constraints must be followed simultaneously and are introduced sequentially over time. We hope \textsc{Sequor} serves as a valuable testbed for studying long-horizon instruction-following. Future work may extend it to additional languages, modalities, and longer interactions.


\section*{Acknowledgments}

We thank Miguel Moura Ramos for his help, constructive feedback, and technical assistance on the paper. This work was supported by the project DECOLLAGE (ERC-2022-CoG 101088763), by the Portuguese Recovery and Resilience Plan through project C645008882-00000055 (Center for Responsible AI), and by FCT/MECI through national funds and when applicable co-funded EU funds under UID/50008: Instituto de Telecomunicações.


\bibliography{colm2026_conference}

\begin{thebibliography}{38}
\providecommand{\natexlab}[1]{#1}
\providecommand{\url}[1]{\texttt{#1}}
\expandafter\ifx\csname urlstyle\endcsname\relax
  \providecommand{\doi}[1]{doi: #1}\else
  \providecommand{\doi}{doi: \begingroup \urlstyle{rm}\Url}\fi

\bibitem[Bai et~al.(2024)Bai, Liu, Bu, He, Liu, Zhou, Lin, Su, Ge, Zheng, and Ouyang]{bai-etal-2024-mt}
Ge~Bai, Jie Liu, Xingyuan Bu, Yancheng He, Jiaheng Liu, Zhanhui Zhou, Zhuoran Lin, Wenbo Su, Tiezheng Ge, Bo~Zheng, and Wanli Ouyang.
\newblock {{MT}-Bench-101: A Fine-Grained Benchmark for Evaluating Large Language Models in Multi-Turn Dialogues}.
\newblock In Lun-Wei Ku, Andre Martins, and Vivek Srikumar (eds.), \emph{Proceedings of the 62nd Annual Meeting of the Association for Computational Linguistics (Volume 1: Long Papers)}, pp.\  7421--7454, Bangkok, Thailand, August 2024. Association for Computational Linguistics.
\newblock \doi{10.18653/v1/2024.acl-long.401}.
\newblock URL \url{https://aclanthology.org/2024.acl-long.401/}.

\bibitem[Chiang et~al.(2024)Chiang, Zheng, Sheng, Angelopoulos, Li, Li, Zhu, Zhang, Jordan, Gonzalez, and Stoica]{10.5555/3692070.3692401}
Wei-Lin Chiang, Lianmin Zheng, Ying Sheng, Anastasios~N. Angelopoulos, Tianle Li, Dacheng Li, Banghua Zhu, Hao Zhang, Michael~I. Jordan, Joseph~E. Gonzalez, and Ion Stoica.
\newblock Chatbot arena: an open platform for evaluating llms by human preference.
\newblock In \emph{Proceedings of the 41st International Conference on Machine Learning}, ICML'24. JMLR.org, 2024.

\bibitem[Colombo et~al.(2022)Colombo, Noiry, Irurozki, and Clemencon]{colombo2022what}
Pierre Colombo, Nathan Noiry, Ekhine Irurozki, and Stephan Clemencon.
\newblock {What are the best Systems? New Perspectives on {NLP} Benchmarking}.
\newblock In Alice~H. Oh, Alekh Agarwal, Danielle Belgrave, and Kyunghyun Cho (eds.), \emph{Advances in Neural Information Processing Systems}, 2022.
\newblock URL \url{https://openreview.net/forum?id=kvtVrzQPvgb}.

\bibitem[Deshpande et~al.(2025)Deshpande, Sirdeshmukh, Mols, Jin, Hernandez-Cardona, Lee, Kritz, Primack, Yue, and Xing]{deshpande-etal-2025-multichallenge}
Kaustubh Deshpande, Ved Sirdeshmukh, Johannes~Baptist Mols, Lifeng Jin, Ed-Yeremai Hernandez-Cardona, Dean Lee, Jeremy Kritz, Willow~E. Primack, Summer Yue, and Chen Xing.
\newblock {{M}ulti{C}hallenge: A Realistic Multi-Turn Conversation Evaluation Benchmark Challenging to Frontier {LLM}s}.
\newblock In Wanxiang Che, Joyce Nabende, Ekaterina Shutova, and Mohammad~Taher Pilehvar (eds.), \emph{Findings of the Association for Computational Linguistics: ACL 2025}, pp.\  18632--18702, Vienna, Austria, July 2025. Association for Computational Linguistics.
\newblock ISBN 979-8-89176-256-5.
\newblock \doi{10.18653/v1/2025.findings-acl.958}.
\newblock URL \url{https://aclanthology.org/2025.findings-acl.958/}.

\bibitem[Dussolle et~al.(2025)Dussolle, Carde{\~n}a, Sato, and Devine]{dussolle-etal-2025-ifeval}
Antoine Dussolle, A.~Carde{\~n}a, Shota Sato, and Peter Devine.
\newblock {{M}-{IFE}val: Multilingual Instruction-Following Evaluation}.
\newblock In Luis Chiruzzo, Alan Ritter, and Lu~Wang (eds.), \emph{Findings of the Association for Computational Linguistics: NAACL 2025}, pp.\  6161--6176, Albuquerque, New Mexico, April 2025. Association for Computational Linguistics.
\newblock ISBN 979-8-89176-195-7.
\newblock \doi{10.18653/v1/2025.findings-naacl.344}.
\newblock URL \url{https://aclanthology.org/2025.findings-naacl.344/}.

\bibitem[Ge et~al.(2025)Ge, Chan, Wang, Yu, Mi, and Yu]{ge2025scalingsyntheticdatacreation}
Tao Ge, Xin Chan, Xiaoyang Wang, Dian Yu, Haitao Mi, and Dong Yu.
\newblock {Scaling Synthetic Data Creation with 1,000,000,000 Personas}.
\newblock 2025.
\newblock URL \url{https://arxiv.org/abs/2406.20094}.

\bibitem[{Google}(2025)]{gemini-3-flash}
{Google}.
\newblock {Gemini 3 Flash}, December 2025.
\newblock URL \url{https://deepmind.google/models/gemini/flash/}.
\newblock Accessed 28-Feb-2026.

\bibitem[{Google}(2026)]{gemini-3.1-flash-lite}
{Google}.
\newblock {Gemini 3.1 Flash-Lite}, March 2026.
\newblock URL \url{https://deepmind.google/models/model-cards/gemini-3-1-flash-lite/}.
\newblock Accessed 25-Mar-2026.

\bibitem[Grattafiori et~al.(2024)Grattafiori, Dubey, Jauhri, Pandey, Kadian, Al-Dahle, Letman, Mathur, Schelten, Vaughan, Yang, Fan, Goyal, Hartshorn, Yang, Mitra, Sravankumar, Korenev, Hinsvark, Rao, Zhang, Rodriguez, Gregerson, Spataru, Roziere, Biron, Tang, Chern, Caucheteux, Nayak, Bi, Marra, McConnell, Keller, Touret, Wu, Wong, Ferrer, Nikolaidis, Allonsius, Song, Pintz, Livshits, Wyatt, Esiobu, Choudhary, Mahajan, Garcia-Olano, Perino, Hupkes, Lakomkin, AlBadawy, Lobanova, Dinan, Smith, Radenovic, Guzmán, Zhang, Synnaeve, Lee, Anderson, Thattai, Nail, Mialon, Pang, Cucurell, Nguyen, Korevaar, Xu, Touvron, Zarov, Ibarra, Kloumann, Misra, Evtimov, Zhang, Copet, Lee, Geffert, Vranes, Park, Mahadeokar, Shah, van~der Linde, Billock, Hong, Lee, Fu, Chi, Huang, Liu, Wang, Yu, Bitton, Spisak, Park, Rocca, Johnstun, Saxe, Jia, Alwala, Prasad, Upasani, Plawiak, Li, Heafield, Stone, El-Arini, Iyer, Malik, Chiu, Bhalla, Lakhotia, Rantala-Yeary, van~der Maaten, Chen, Tan, Jenkins, Martin, Madaan, Malo, Blecher,
  Landzaat, de~Oliveira, Muzzi, Pasupuleti, Singh, Paluri, Kardas, Tsimpoukelli, Oldham, Rita, Pavlova, Kambadur, Lewis, Si, Singh, Hassan, Goyal, Torabi, Bashlykov, Bogoychev, Chatterji, Zhang, Duchenne, Çelebi, Alrassy, Zhang, Li, Vasic, Weng, Bhargava, Dubal, Krishnan, Koura, Xu, He, Dong, Srinivasan, Ganapathy, Calderer, Cabral, Stojnic, Raileanu, Maheswari, Girdhar, Patel, Sauvestre, Polidoro, Sumbaly, Taylor, Silva, Hou, Wang, Hosseini, Chennabasappa, Singh, Bell, Kim, Edunov, Nie, Narang, Raparthy, Shen, Wan, Bhosale, Zhang, Vandenhende, Batra, Whitman, Sootla, Collot, Gururangan, Borodinsky, Herman, Fowler, Sheasha, Georgiou, Scialom, Speckbacher, Mihaylov, Xiao, Karn, Goswami, Gupta, Ramanathan, Kerkez, Gonguet, Do, Vogeti, Albiero, Petrovic, Chu, Xiong, Fu, Meers, Martinet, Wang, Wang, Tan, Xia, Xie, Jia, Wang, Goldschlag, Gaur, Babaei, Wen, Song, Zhang, Li, Mao, Coudert, Yan, Chen, Papakipos, Singh, Srivastava, Jain, Kelsey, Shajnfeld, Gangidi, Victoria, Goldstand, Menon, Sharma, Boesenberg,
  Baevski, Feinstein, Kallet, Sangani, Teo, Yunus, Lupu, Alvarado, Caples, Gu, Ho, Poulton, Ryan, Ramchandani, Dong, Franco, Goyal, Saraf, Chowdhury, Gabriel, Bharambe, Eisenman, Yazdan, James, Maurer, Leonhardi, Huang, Loyd, Paola, Paranjape, Liu, Wu, Ni, Hancock, Wasti, Spence, Stojkovic, Gamido, Montalvo, Parker, Burton, Mejia, Liu, Wang, Kim, Zhou, Hu, Chu, Cai, Tindal, Feichtenhofer, Gao, Civin, Beaty, Kreymer, Li, Adkins, Xu, Testuggine, David, Parikh, Liskovich, Foss, Wang, Le, Holland, Dowling, Jamil, Montgomery, Presani, Hahn, Wood, Le, Brinkman, Arcaute, Dunbar, Smothers, Sun, Kreuk, Tian, Kokkinos, Ozgenel, Caggioni, Kanayet, Seide, Florez, Schwarz, Badeer, Swee, Halpern, Herman, Sizov, Guangyi, Zhang, Lakshminarayanan, Inan, Shojanazeri, Zou, Wang, Zha, Habeeb, Rudolph, Suk, Aspegren, Goldman, Zhan, Damlaj, Molybog, Tufanov, Leontiadis, Veliche, Gat, Weissman, Geboski, Kohli, Lam, Asher, Gaya, Marcus, Tang, Chan, Zhen, Reizenstein, Teboul, Zhong, Jin, Yang, Cummings, Carvill, Shepard, McPhie,
  Torres, Ginsburg, Wang, Wu, U, Saxena, Khandelwal, Zand, Matosich, Veeraraghavan, Michelena, Li, Jagadeesh, Huang, Chawla, Huang, Chen, Garg, A, Silva, Bell, Zhang, Guo, Yu, Moshkovich, Wehrstedt, Khabsa, Avalani, Bhatt, Mankus, Hasson, Lennie, Reso, Groshev, Naumov, Lathi, Keneally, Liu, Seltzer, Valko, Restrepo, Patel, Vyatskov, Samvelyan, Clark, Macey, Wang, Hermoso, Metanat, Rastegari, Bansal, Santhanam, Parks, White, Bawa, Singhal, Egebo, Usunier, Mehta, Laptev, Dong, Cheng, Chernoguz, Hart, Salpekar, Kalinli, Kent, Parekh, Saab, Balaji, Rittner, Bontrager, Roux, Dollar, Zvyagina, Ratanchandani, Yuvraj, Liang, Alao, Rodriguez, Ayub, Murthy, Nayani, Mitra, Parthasarathy, Li, Hogan, Battey, Wang, Howes, Rinott, Mehta, Siby, Bondu, Datta, Chugh, Hunt, Dhillon, Sidorov, Pan, Mahajan, Verma, Yamamoto, Ramaswamy, Lindsay, Lindsay, Feng, Lin, Zha, Patil, Shankar, Zhang, Zhang, Wang, Agarwal, Sajuyigbe, Chintala, Max, Chen, Kehoe, Satterfield, Govindaprasad, Gupta, Deng, Cho, Virk, Subramanian, Choudhury,
  Goldman, Remez, Glaser, Best, Koehler, Robinson, Li, Zhang, Matthews, Chou, Shaked, Vontimitta, Ajayi, Montanez, Mohan, Kumar, Mangla, Ionescu, Poenaru, Mihailescu, Ivanov, Li, Wang, Jiang, Bouaziz, Constable, Tang, Wu, Wang, Wu, Gao, Kleinman, Chen, Hu, Jia, Qi, Li, Zhang, Zhang, Adi, Nam, Yu, Wang, Zhao, Hao, Qian, Li, He, Rait, DeVito, Rosnbrick, Wen, Yang, Zhao, and Ma]{grattafiori2024llama3herdmodels}
Aaron Grattafiori, Abhimanyu Dubey, Abhinav Jauhri, Abhinav Pandey, Abhishek Kadian, Ahmad Al-Dahle, Aiesha Letman, Akhil Mathur, Alan Schelten, Alex Vaughan, Amy Yang, Angela Fan, Anirudh Goyal, Anthony Hartshorn, Aobo Yang, Archi Mitra, Archie Sravankumar, Artem Korenev, Arthur Hinsvark, Arun Rao, Aston Zhang, Aurelien Rodriguez, Austen Gregerson, Ava Spataru, Baptiste Roziere, Bethany Biron, Binh Tang, Bobbie Chern, Charlotte Caucheteux, Chaya Nayak, Chloe Bi, Chris Marra, Chris McConnell, Christian Keller, Christophe Touret, Chunyang Wu, Corinne Wong, Cristian~Canton Ferrer, Cyrus Nikolaidis, Damien Allonsius, Daniel Song, Danielle Pintz, Danny Livshits, Danny Wyatt, David Esiobu, Dhruv Choudhary, Dhruv Mahajan, Diego Garcia-Olano, Diego Perino, Dieuwke Hupkes, Egor Lakomkin, Ehab AlBadawy, Elina Lobanova, Emily Dinan, Eric~Michael Smith, Filip Radenovic, Francisco Guzmán, Frank Zhang, Gabriel Synnaeve, Gabrielle Lee, Georgia~Lewis Anderson, Govind Thattai, Graeme Nail, Gregoire Mialon, Guan Pang,
  Guillem Cucurell, Hailey Nguyen, Hannah Korevaar, Hu~Xu, Hugo Touvron, Iliyan Zarov, Imanol~Arrieta Ibarra, Isabel Kloumann, Ishan Misra, Ivan Evtimov, Jack Zhang, Jade Copet, Jaewon Lee, Jan Geffert, Jana Vranes, Jason Park, Jay Mahadeokar, Jeet Shah, Jelmer van~der Linde, Jennifer Billock, Jenny Hong, Jenya Lee, Jeremy Fu, Jianfeng Chi, Jianyu Huang, Jiawen Liu, Jie Wang, Jiecao Yu, Joanna Bitton, Joe Spisak, Jongsoo Park, Joseph Rocca, Joshua Johnstun, Joshua Saxe, Junteng Jia, Kalyan~Vasuden Alwala, Karthik Prasad, Kartikeya Upasani, Kate Plawiak, Ke~Li, Kenneth Heafield, Kevin Stone, Khalid El-Arini, Krithika Iyer, Kshitiz Malik, Kuenley Chiu, Kunal Bhalla, Kushal Lakhotia, Lauren Rantala-Yeary, Laurens van~der Maaten, Lawrence Chen, Liang Tan, Liz Jenkins, Louis Martin, Lovish Madaan, Lubo Malo, Lukas Blecher, Lukas Landzaat, Luke de~Oliveira, Madeline Muzzi, Mahesh Pasupuleti, Mannat Singh, Manohar Paluri, Marcin Kardas, Maria Tsimpoukelli, Mathew Oldham, Mathieu Rita, Maya Pavlova, Melanie Kambadur,
  Mike Lewis, Min Si, Mitesh~Kumar Singh, Mona Hassan, Naman Goyal, Narjes Torabi, Nikolay Bashlykov, Nikolay Bogoychev, Niladri Chatterji, Ning Zhang, Olivier Duchenne, Onur Çelebi, Patrick Alrassy, Pengchuan Zhang, Pengwei Li, Petar Vasic, Peter Weng, Prajjwal Bhargava, Pratik Dubal, Praveen Krishnan, Punit~Singh Koura, Puxin Xu, Qing He, Qingxiao Dong, Ragavan Srinivasan, Raj Ganapathy, Ramon Calderer, Ricardo~Silveira Cabral, Robert Stojnic, Roberta Raileanu, Rohan Maheswari, Rohit Girdhar, Rohit Patel, Romain Sauvestre, Ronnie Polidoro, Roshan Sumbaly, Ross Taylor, Ruan Silva, Rui Hou, Rui Wang, Saghar Hosseini, Sahana Chennabasappa, Sanjay Singh, Sean Bell, Seohyun~Sonia Kim, Sergey Edunov, Shaoliang Nie, Sharan Narang, Sharath Raparthy, Sheng Shen, Shengye Wan, Shruti Bhosale, Shun Zhang, Simon Vandenhende, Soumya Batra, Spencer Whitman, Sten Sootla, Stephane Collot, Suchin Gururangan, Sydney Borodinsky, Tamar Herman, Tara Fowler, Tarek Sheasha, Thomas Georgiou, Thomas Scialom, Tobias Speckbacher,
  Todor Mihaylov, Tong Xiao, Ujjwal Karn, Vedanuj Goswami, Vibhor Gupta, Vignesh Ramanathan, Viktor Kerkez, Vincent Gonguet, Virginie Do, Vish Vogeti, Vítor Albiero, Vladan Petrovic, Weiwei Chu, Wenhan Xiong, Wenyin Fu, Whitney Meers, Xavier Martinet, Xiaodong Wang, Xiaofang Wang, Xiaoqing~Ellen Tan, Xide Xia, Xinfeng Xie, Xuchao Jia, Xuewei Wang, Yaelle Goldschlag, Yashesh Gaur, Yasmine Babaei, Yi~Wen, Yiwen Song, Yuchen Zhang, Yue Li, Yuning Mao, Zacharie~Delpierre Coudert, Zheng Yan, Zhengxing Chen, Zoe Papakipos, Aaditya Singh, Aayushi Srivastava, Abha Jain, Adam Kelsey, Adam Shajnfeld, Adithya Gangidi, Adolfo Victoria, Ahuva Goldstand, Ajay Menon, Ajay Sharma, Alex Boesenberg, Alexei Baevski, Allie Feinstein, Amanda Kallet, Amit Sangani, Amos Teo, Anam Yunus, Andrei Lupu, Andres Alvarado, Andrew Caples, Andrew Gu, Andrew Ho, Andrew Poulton, Andrew Ryan, Ankit Ramchandani, Annie Dong, Annie Franco, Anuj Goyal, Aparajita Saraf, Arkabandhu Chowdhury, Ashley Gabriel, Ashwin Bharambe, Assaf Eisenman, Azadeh
  Yazdan, Beau James, Ben Maurer, Benjamin Leonhardi, Bernie Huang, Beth Loyd, Beto~De Paola, Bhargavi Paranjape, Bing Liu, Bo~Wu, Boyu Ni, Braden Hancock, Bram Wasti, Brandon Spence, Brani Stojkovic, Brian Gamido, Britt Montalvo, Carl Parker, Carly Burton, Catalina Mejia, Ce~Liu, Changhan Wang, Changkyu Kim, Chao Zhou, Chester Hu, Ching-Hsiang Chu, Chris Cai, Chris Tindal, Christoph Feichtenhofer, Cynthia Gao, Damon Civin, Dana Beaty, Daniel Kreymer, Daniel Li, David Adkins, David Xu, Davide Testuggine, Delia David, Devi Parikh, Diana Liskovich, Didem Foss, Dingkang Wang, Duc Le, Dustin Holland, Edward Dowling, Eissa Jamil, Elaine Montgomery, Eleonora Presani, Emily Hahn, Emily Wood, Eric-Tuan Le, Erik Brinkman, Esteban Arcaute, Evan Dunbar, Evan Smothers, Fei Sun, Felix Kreuk, Feng Tian, Filippos Kokkinos, Firat Ozgenel, Francesco Caggioni, Frank Kanayet, Frank Seide, Gabriela~Medina Florez, Gabriella Schwarz, Gada Badeer, Georgia Swee, Gil Halpern, Grant Herman, Grigory Sizov, Guangyi, Zhang, Guna
  Lakshminarayanan, Hakan Inan, Hamid Shojanazeri, Han Zou, Hannah Wang, Hanwen Zha, Haroun Habeeb, Harrison Rudolph, Helen Suk, Henry Aspegren, Hunter Goldman, Hongyuan Zhan, Ibrahim Damlaj, Igor Molybog, Igor Tufanov, Ilias Leontiadis, Irina-Elena Veliche, Itai Gat, Jake Weissman, James Geboski, James Kohli, Janice Lam, Japhet Asher, Jean-Baptiste Gaya, Jeff Marcus, Jeff Tang, Jennifer Chan, Jenny Zhen, Jeremy Reizenstein, Jeremy Teboul, Jessica Zhong, Jian Jin, Jingyi Yang, Joe Cummings, Jon Carvill, Jon Shepard, Jonathan McPhie, Jonathan Torres, Josh Ginsburg, Junjie Wang, Kai Wu, Kam~Hou U, Karan Saxena, Kartikay Khandelwal, Katayoun Zand, Kathy Matosich, Kaushik Veeraraghavan, Kelly Michelena, Keqian Li, Kiran Jagadeesh, Kun Huang, Kunal Chawla, Kyle Huang, Lailin Chen, Lakshya Garg, Lavender A, Leandro Silva, Lee Bell, Lei Zhang, Liangpeng Guo, Licheng Yu, Liron Moshkovich, Luca Wehrstedt, Madian Khabsa, Manav Avalani, Manish Bhatt, Martynas Mankus, Matan Hasson, Matthew Lennie, Matthias Reso, Maxim
  Groshev, Maxim Naumov, Maya Lathi, Meghan Keneally, Miao Liu, Michael~L. Seltzer, Michal Valko, Michelle Restrepo, Mihir Patel, Mik Vyatskov, Mikayel Samvelyan, Mike Clark, Mike Macey, Mike Wang, Miquel~Jubert Hermoso, Mo~Metanat, Mohammad Rastegari, Munish Bansal, Nandhini Santhanam, Natascha Parks, Natasha White, Navyata Bawa, Nayan Singhal, Nick Egebo, Nicolas Usunier, Nikhil Mehta, Nikolay~Pavlovich Laptev, Ning Dong, Norman Cheng, Oleg Chernoguz, Olivia Hart, Omkar Salpekar, Ozlem Kalinli, Parkin Kent, Parth Parekh, Paul Saab, Pavan Balaji, Pedro Rittner, Philip Bontrager, Pierre Roux, Piotr Dollar, Polina Zvyagina, Prashant Ratanchandani, Pritish Yuvraj, Qian Liang, Rachad Alao, Rachel Rodriguez, Rafi Ayub, Raghotham Murthy, Raghu Nayani, Rahul Mitra, Rangaprabhu Parthasarathy, Raymond Li, Rebekkah Hogan, Robin Battey, Rocky Wang, Russ Howes, Ruty Rinott, Sachin Mehta, Sachin Siby, Sai~Jayesh Bondu, Samyak Datta, Sara Chugh, Sara Hunt, Sargun Dhillon, Sasha Sidorov, Satadru Pan, Saurabh Mahajan,
  Saurabh Verma, Seiji Yamamoto, Sharadh Ramaswamy, Shaun Lindsay, Shaun Lindsay, Sheng Feng, Shenghao Lin, Shengxin~Cindy Zha, Shishir Patil, Shiva Shankar, Shuqiang Zhang, Shuqiang Zhang, Sinong Wang, Sneha Agarwal, Soji Sajuyigbe, Soumith Chintala, Stephanie Max, Stephen Chen, Steve Kehoe, Steve Satterfield, Sudarshan Govindaprasad, Sumit Gupta, Summer Deng, Sungmin Cho, Sunny Virk, Suraj Subramanian, Sy~Choudhury, Sydney Goldman, Tal Remez, Tamar Glaser, Tamara Best, Thilo Koehler, Thomas Robinson, Tianhe Li, Tianjun Zhang, Tim Matthews, Timothy Chou, Tzook Shaked, Varun Vontimitta, Victoria Ajayi, Victoria Montanez, Vijai Mohan, Vinay~Satish Kumar, Vishal Mangla, Vlad Ionescu, Vlad Poenaru, Vlad~Tiberiu Mihailescu, Vladimir Ivanov, Wei Li, Wenchen Wang, Wenwen Jiang, Wes Bouaziz, Will Constable, Xiaocheng Tang, Xiaojian Wu, Xiaolan Wang, Xilun Wu, Xinbo Gao, Yaniv Kleinman, Yanjun Chen, Ye~Hu, Ye~Jia, Ye~Qi, Yenda Li, Yilin Zhang, Ying Zhang, Yossi Adi, Youngjin Nam, Yu, Wang, Yu~Zhao, Yuchen Hao, Yundi
  Qian, Yunlu Li, Yuzi He, Zach Rait, Zachary DeVito, Zef Rosnbrick, Zhaoduo Wen, Zhenyu Yang, Zhiwei Zhao, and Zhiyu Ma.
\newblock {The Llama 3 Herd of Models}, 2024.
\newblock URL \url{https://arxiv.org/abs/2407.21783}.

\bibitem[Gu et~al.(2025)Gu, Jiang, Shi, Tan, Zhai, Xu, Li, Shen, Ma, Liu, Wang, Zhang, Wang, Gao, Ni, and Guo]{gu2025surveyllmasajudge}
Jiawei Gu, Xuhui Jiang, Zhichao Shi, Hexiang Tan, Xuehao Zhai, Chengjin Xu, Wei Li, Yinghan Shen, Shengjie Ma, Honghao Liu, Saizhuo Wang, Kun Zhang, Yuanzhuo Wang, Wen Gao, Lionel Ni, and Jian Guo.
\newblock {A Survey on LLM-as-a-Judge}, 2025.
\newblock URL \url{https://arxiv.org/abs/2411.15594}.

\bibitem[He et~al.(2024)He, Jin, Wang, Bi, Mandyam, Zhang, Zhu, Li, Xu, Lv, Bhosale, Zhu, Sankararaman, Helenowski, Kambadur, Tayade, Ma, Fang, and Wang]{he2024multiifbenchmarkingllmsmultiturn}
Yun He, Di~Jin, Chaoqi Wang, Chloe Bi, Karishma Mandyam, Hejia Zhang, Chen Zhu, Ning Li, Tengyu Xu, Hongjiang Lv, Shruti Bhosale, Chenguang Zhu, Karthik~Abinav Sankararaman, Eryk Helenowski, Melanie Kambadur, Aditya Tayade, Hao Ma, Han Fang, and Sinong Wang.
\newblock {Multi-IF: Benchmarking LLMs on Multi-Turn and Multilingual Instructions Following}.
\newblock 2024.
\newblock URL \url{https://arxiv.org/abs/2410.15553}.

\bibitem[Jiang et~al.(2025)Jiang, Hao, Cho, Li, Yuan, Chen, Ungar, Taylor, and Roth]{jiang2025know}
Bowen Jiang, Zhuoqun Hao, Young~Min Cho, Bryan Li, Yuan Yuan, Sihao Chen, Lyle Ungar, Camillo~Jose Taylor, and Dan Roth.
\newblock {Know Me, Respond to Me: Benchmarking {LLM}s for Dynamic User Profiling and Personalized Responses at Scale}.
\newblock In \emph{Second Conference on Language Modeling}, 2025.
\newblock URL \url{https://openreview.net/forum?id=6ox8XZGOqP}.

\bibitem[Jiang et~al.(2024)Jiang, Wang, Zeng, Zhong, Li, Mi, Shang, Jiang, Liu, and Wang]{jiang-etal-2024-followbench}
Yuxin Jiang, Yufei Wang, Xingshan Zeng, Wanjun Zhong, Liangyou Li, Fei Mi, Lifeng Shang, Xin Jiang, Qun Liu, and Wei Wang.
\newblock {{F}ollow{B}ench: A Multi-level Fine-grained Constraints Following Benchmark for Large Language Models}.
\newblock In Lun-Wei Ku, Andre Martins, and Vivek Srikumar (eds.), \emph{Proceedings of the 62nd Annual Meeting of the Association for Computational Linguistics (Volume 1: Long Papers)}, pp.\  4667--4688, Bangkok, Thailand, August 2024. Association for Computational Linguistics.
\newblock \doi{10.18653/v1/2024.acl-long.257}.
\newblock URL \url{https://aclanthology.org/2024.acl-long.257/}.

\bibitem[Joulin et~al.(2016{\natexlab{a}})Joulin, Grave, Bojanowski, Douze, Jégou, and Mikolov]{joulin2016fasttextzipcompressingtextclassification}
Armand Joulin, Edouard Grave, Piotr Bojanowski, Matthijs Douze, Hérve Jégou, and Tomas Mikolov.
\newblock {FastText.zip: Compressing text classification models}, 2016{\natexlab{a}}.
\newblock URL \url{https://arxiv.org/abs/1612.03651}.

\bibitem[Joulin et~al.(2016{\natexlab{b}})Joulin, Grave, Bojanowski, and Mikolov]{joulin2016bagtricksefficienttext}
Armand Joulin, Edouard Grave, Piotr Bojanowski, and Tomas Mikolov.
\newblock {Bag of Tricks for Efficient Text Classification}, 2016{\natexlab{b}}.
\newblock URL \url{https://arxiv.org/abs/1607.01759}.

\bibitem[Kim et~al.(2025)Kim, Suk, Kim, Muennighoff, Kim, and Oh]{kim-etal-2025-llm-interviewer}
Eunsu Kim, Juyoung Suk, Seungone Kim, Niklas Muennighoff, Dongkwan Kim, and Alice Oh.
\newblock {{LLM}-as-an-Interviewer: Beyond Static Testing Through Dynamic {LLM} Evaluation}.
\newblock In Wanxiang Che, Joyce Nabende, Ekaterina Shutova, and Mohammad~Taher Pilehvar (eds.), \emph{Findings of the Association for Computational Linguistics: ACL 2025}, pp.\  26456--26493, Vienna, Austria, July 2025. Association for Computational Linguistics.
\newblock ISBN 979-8-89176-256-5.
\newblock \doi{10.18653/v1/2025.findings-acl.1357}.
\newblock URL \url{https://aclanthology.org/2025.findings-acl.1357/}.

\bibitem[Kwan et~al.(2024)Kwan, Zeng, Jiang, Wang, Li, Shang, Jiang, Liu, and Wong]{kwan-etal-2024-mt}
Wai-Chung Kwan, Xingshan Zeng, Yuxin Jiang, Yufei Wang, Liangyou Li, Lifeng Shang, Xin Jiang, Qun Liu, and Kam-Fai Wong.
\newblock {{MT}-Eval: A Multi-Turn Capabilities Evaluation Benchmark for Large Language Models}.
\newblock In Yaser Al-Onaizan, Mohit Bansal, and Yun-Nung Chen (eds.), \emph{Proceedings of the 2024 Conference on Empirical Methods in Natural Language Processing}, pp.\  20153--20177, Miami, Florida, USA, November 2024. Association for Computational Linguistics.
\newblock \doi{10.18653/v1/2024.emnlp-main.1124}.
\newblock URL \url{https://aclanthology.org/2024.emnlp-main.1124/}.

\bibitem[Laban et~al.(2025)Laban, Hayashi, Zhou, and Neville]{laban2025llmslostmultiturnconversation}
Philippe Laban, Hiroaki Hayashi, Yingbo Zhou, and Jennifer Neville.
\newblock {LLMs Get Lost In Multi-Turn Conversation}.
\newblock 2025.
\newblock URL \url{https://arxiv.org/abs/2505.06120}.

\bibitem[Maharana et~al.(2024)Maharana, Lee, Tulyakov, Bansal, Barbieri, and Fang]{maharana-etal-2024-evaluating}
Adyasha Maharana, Dong-Ho Lee, Sergey Tulyakov, Mohit Bansal, Francesco Barbieri, and Yuwei Fang.
\newblock {Evaluating Very Long-Term Conversational Memory of {LLM} Agents}.
\newblock In Lun-Wei Ku, Andre Martins, and Vivek Srikumar (eds.), \emph{Proceedings of the 62nd Annual Meeting of the Association for Computational Linguistics (Volume 1: Long Papers)}, pp.\  13851--13870, Bangkok, Thailand, August 2024. Association for Computational Linguistics.
\newblock \doi{10.18653/v1/2024.acl-long.747}.
\newblock URL \url{https://aclanthology.org/2024.acl-long.747/}.

\bibitem[Olmo et~al.(2025)Olmo, Ettinger, Bertsch, Kuehl, Graham, Heineman, Groeneveld, Brahman, Timbers, Ivison, Morrison, Poznanski, Lo, Soldaini, Jordan, Chen, Noukhovitch, Lambert, Walsh, Dasigi, Berry, Malik, Shah, Geng, Arora, Gupta, Anderson, Xiao, Murray, Romero, Graf, Asai, Bhagia, Wettig, Liu, Rangapur, Anastasiades, Huang, Schwenk, Trivedi, Magnusson, Lochner, Liu, Miranda, Sap, Morgan, Schmitz, Guerquin, Wilson, Huff, Bras, Xin, Shao, Skjonsberg, Shen, Li, Wilde, Pyatkin, Merrill, Chang, Gu, Zeng, Sabharwal, Zettlemoyer, Koh, Farhadi, Smith, and Hajishirzi]{olmo2025olmo3}
Team Olmo, Allyson Ettinger, Amanda Bertsch, Bailey Kuehl, David Graham, David Heineman, Dirk Groeneveld, Faeze Brahman, Finbarr Timbers, Hamish Ivison, Jacob Morrison, Jake Poznanski, Kyle Lo, Luca Soldaini, Matt Jordan, Mayee Chen, Michael Noukhovitch, Nathan Lambert, Pete Walsh, Pradeep Dasigi, Robert Berry, Saumya Malik, Saurabh Shah, Scott Geng, Shane Arora, Shashank Gupta, Taira Anderson, Teng Xiao, Tyler Murray, Tyler Romero, Victoria Graf, Akari Asai, Akshita Bhagia, Alexander Wettig, Alisa Liu, Aman Rangapur, Chloe Anastasiades, Costa Huang, Dustin Schwenk, Harsh Trivedi, Ian Magnusson, Jaron Lochner, Jiacheng Liu, Lester James~V. Miranda, Maarten Sap, Malia Morgan, Michael Schmitz, Michal Guerquin, Michael Wilson, Regan Huff, Ronan~Le Bras, Rui Xin, Rulin Shao, Sam Skjonsberg, Shannon~Zejiang Shen, Shuyue~Stella Li, Tucker Wilde, Valentina Pyatkin, Will Merrill, Yapei Chang, Yuling Gu, Zhiyuan Zeng, Ashish Sabharwal, Luke Zettlemoyer, Pang~Wei Koh, Ali Farhadi, Noah~A. Smith, and Hannaneh
  Hajishirzi.
\newblock {Olmo 3}, 2025.
\newblock URL \url{https://arxiv.org/abs/2512.13961}.

\bibitem[{OpenAI}(2025)]{introducing-gpt-5.2}
{OpenAI}.
\newblock {Introducing GPT‑5.2}, December 2025.
\newblock URL \url{https://openai.com/index/introducing-gpt-5-2/}.
\newblock Accessed 20-Feb-2026.

\bibitem[OpenAI et~al.(2025)OpenAI, :, Agarwal, Ahmad, Ai, Altman, Applebaum, Arbus, Arora, Bai, Baker, Bao, Barak, Bennett, Bertao, Brett, Brevdo, Brockman, Bubeck, Chang, Chen, Chen, Cheung, Clark, Cook, Dukhan, Dvorak, Fives, Fomenko, Garipov, Georgiev, Glaese, Gogineni, Goucher, Gross, Guzman, Hallman, Hehir, Heidecke, Helyar, Hu, Huet, Huh, Jain, Johnson, Koch, Kofman, Kundel, Kwon, Kyrylov, Le, Leclerc, Lennon, Lessans, Lezcano-Casado, Li, Li, Lin, Liss, Lily, Liu, Liu, Lu, Lu, Martinovic, McCallum, McGrath, McKinney, McLaughlin, Mei, Mostovoy, Mu, Myles, Neitz, Nichol, Pachocki, Paino, Palmie, Pantuliano, Parascandolo, Park, Pathak, Paz, Peran, Pimenov, Pokrass, Proehl, Qiu, Raila, Raso, Ren, Richardson, Robinson, Rotsted, Salman, Sanjeev, Schwarzer, Sculley, Sikchi, Simon, Singhal, Song, Stuckey, Sun, Tillet, Toizer, Tsimpourlas, Vyas, Wallace, Wang, Wang, Watkins, Weil, Wendling, Whinnery, Whitney, Wong, Yang, Yang, Yasunaga, Ying, Zaremba, Zhan, Zhang, Zhang, Zhang, and
  Zhao]{openai2025gptoss120bgptoss20bmodel}
OpenAI, :, Sandhini Agarwal, Lama Ahmad, Jason Ai, Sam Altman, Andy Applebaum, Edwin Arbus, Rahul~K. Arora, Yu~Bai, Bowen Baker, Haiming Bao, Boaz Barak, Ally Bennett, Tyler Bertao, Nivedita Brett, Eugene Brevdo, Greg Brockman, Sebastien Bubeck, Che Chang, Kai Chen, Mark Chen, Enoch Cheung, Aidan Clark, Dan Cook, Marat Dukhan, Casey Dvorak, Kevin Fives, Vlad Fomenko, Timur Garipov, Kristian Georgiev, Mia Glaese, Tarun Gogineni, Adam Goucher, Lukas Gross, Katia~Gil Guzman, John Hallman, Jackie Hehir, Johannes Heidecke, Alec Helyar, Haitang Hu, Romain Huet, Jacob Huh, Saachi Jain, Zach Johnson, Chris Koch, Irina Kofman, Dominik Kundel, Jason Kwon, Volodymyr Kyrylov, Elaine~Ya Le, Guillaume Leclerc, James~Park Lennon, Scott Lessans, Mario Lezcano-Casado, Yuanzhi Li, Zhuohan Li, Ji~Lin, Jordan Liss, Lily, Liu, Jiancheng Liu, Kevin Lu, Chris Lu, Zoran Martinovic, Lindsay McCallum, Josh McGrath, Scott McKinney, Aidan McLaughlin, Song Mei, Steve Mostovoy, Tong Mu, Gideon Myles, Alexander Neitz, Alex Nichol, Jakub
  Pachocki, Alex Paino, Dana Palmie, Ashley Pantuliano, Giambattista Parascandolo, Jongsoo Park, Leher Pathak, Carolina Paz, Ludovic Peran, Dmitry Pimenov, Michelle Pokrass, Elizabeth Proehl, Huida Qiu, Gaby Raila, Filippo Raso, Hongyu Ren, Kimmy Richardson, David Robinson, Bob Rotsted, Hadi Salman, Suvansh Sanjeev, Max Schwarzer, D.~Sculley, Harshit Sikchi, Kendal Simon, Karan Singhal, Yang Song, Dane Stuckey, Zhiqing Sun, Philippe Tillet, Sam Toizer, Foivos Tsimpourlas, Nikhil Vyas, Eric Wallace, Xin Wang, Miles Wang, Olivia Watkins, Kevin Weil, Amy Wendling, Kevin Whinnery, Cedric Whitney, Hannah Wong, Lin Yang, Yu~Yang, Michihiro Yasunaga, Kristen Ying, Wojciech Zaremba, Wenting Zhan, Cyril Zhang, Brian Zhang, Eddie Zhang, and Shengjia Zhao.
\newblock gpt-oss-120b \& gpt-oss-20b model card, 2025.
\newblock URL \url{https://arxiv.org/abs/2508.10925}.

\bibitem[Penedo et~al.(2024)Penedo, Kydlíček, Cappelli, Sasko, and Wolf]{penedo2024datatrove}
Guilherme Penedo, Hynek Kydlíček, Alessandro Cappelli, Mario Sasko, and Thomas Wolf.
\newblock {DataTrove: large scale data processing}, 2024.
\newblock URL \url{https://github.com/huggingface/datatrove}.

\bibitem[Pyatkin et~al.(2025)Pyatkin, Malik, Graf, Ivison, Huang, Dasigi, Lambert, and Hajishirzi]{pyatkin2025generalizingverifiableinstructionfollowing}
Valentina Pyatkin, Saumya Malik, Victoria Graf, Hamish Ivison, Shengyi Huang, Pradeep Dasigi, Nathan Lambert, and Hannaneh Hajishirzi.
\newblock {Generalizing Verifiable Instruction Following}.
\newblock 2025.
\newblock URL \url{https://arxiv.org/abs/2507.02833}.

\bibitem[Qin et~al.(2024)Qin, Song, Hu, Yao, Cho, Wang, Wu, Liu, Liu, and Yu]{qin-etal-2024-infobench}
Yiwei Qin, Kaiqiang Song, Yebowen Hu, Wenlin Yao, Sangwoo Cho, Xiaoyang Wang, Xuansheng Wu, Fei Liu, Pengfei Liu, and Dong Yu.
\newblock {{I}n{F}o{B}ench: Evaluating Instruction Following Ability in Large Language Models}.
\newblock In Lun-Wei Ku, Andre Martins, and Vivek Srikumar (eds.), \emph{Findings of the Association for Computational Linguistics: ACL 2024}, pp.\  13025--13048, Bangkok, Thailand, August 2024. Association for Computational Linguistics.
\newblock \doi{10.18653/v1/2024.findings-acl.772}.
\newblock URL \url{https://aclanthology.org/2024.findings-acl.772/}.

\bibitem[Rakotonirina et~al.(2025)Rakotonirina, Hamdy, Campos, Weber, Testoni, Fadaee, Pezzelle, and Del~Tredici]{rakotonirina-etal-2025-tools}
Nathana{\"e}l~Carraz Rakotonirina, Mohammed Hamdy, Jon~Ander Campos, Lucas Weber, Alberto Testoni, Marzieh Fadaee, Sandro Pezzelle, and Marco Del~Tredici.
\newblock {From Tools to Teammates: Evaluating {LLM}s in Multi-Session Coding Interactions}.
\newblock In Wanxiang Che, Joyce Nabende, Ekaterina Shutova, and Mohammad~Taher Pilehvar (eds.), \emph{Proceedings of the 63rd Annual Meeting of the Association for Computational Linguistics (Volume 1: Long Papers)}, pp.\  19609--19642, Vienna, Austria, July 2025. Association for Computational Linguistics.
\newblock ISBN 979-8-89176-251-0.
\newblock \doi{10.18653/v1/2025.acl-long.964}.
\newblock URL \url{https://aclanthology.org/2025.acl-long.964/}.

\bibitem[Team et~al.(2025{\natexlab{a}})Team, Kamath, Ferret, Pathak, Vieillard, Merhej, Perrin, Matejovicova, Ramé, Rivière, Rouillard, Mesnard, Cideron, bastien Grill, Ramos, Yvinec, Casbon, Pot, Penchev, Liu, Visin, Kenealy, Beyer, Zhai, Tsitsulin, Busa-Fekete, Feng, Sachdeva, Coleman, Gao, Mustafa, Barr, Parisotto, Tian, Eyal, Cherry, Peter, Sinopalnikov, Bhupatiraju, Agarwal, Kazemi, Malkin, Kumar, Vilar, Brusilovsky, Luo, Steiner, Friesen, Sharma, Sharma, Gilady, Goedeckemeyer, Saade, Feng, Kolesnikov, Bendebury, Abdagic, Vadi, György, Pinto, Das, Bapna, Miech, Yang, Paterson, Shenoy, Chakrabarti, Piot, Wu, Shahriari, Petrini, Chen, Lan, Choquette-Choo, Carey, Brick, Deutsch, Eisenbud, Cattle, Cheng, Paparas, Sreepathihalli, Reid, Tran, Zelle, Noland, Huizenga, Kharitonov, Liu, Amirkhanyan, Cameron, Hashemi, Klimczak-Plucińska, Singh, Mehta, Lehri, Hazimeh, Ballantyne, Szpektor, Nardini, Pouget-Abadie, Chan, Stanton, Wieting, Lai, Orbay, Fernandez, Newlan, yeong Ji, Singh, Black, Yu, Hui,
  Vodrahalli, Greff, Qiu, Valentine, Coelho, Ritter, Hoffman, Watson, Chaturvedi, Moynihan, Ma, Babar, Noy, Byrd, Roy, Momchev, Chauhan, Sachdeva, Bunyan, Botarda, Caron, Rubenstein, Culliton, Schmid, Sessa, Xu, Stanczyk, Tafti, Shivanna, Wu, Pan, Rokni, Willoughby, Vallu, Mullins, Jerome, Smoot, Girgin, Iqbal, Reddy, Sheth, Põder, Bhatnagar, Panyam, Eiger, Zhang, Liu, Yacovone, Liechty, Kalra, Evci, Misra, Roseberry, Feinberg, Kolesnikov, Han, Kwon, Chen, Chow, Zhu, Wei, Egyed, Cotruta, Giang, Kirk, Rao, Black, Babar, Lo, Moreira, Martins, Sanseviero, Gonzalez, Gleicher, Warkentin, Mirrokni, Senter, Collins, Barral, Ghahramani, Hadsell, Matias, Sculley, Petrov, Fiedel, Shazeer, Vinyals, Dean, Hassabis, Kavukcuoglu, Farabet, Buchatskaya, Alayrac, Anil, Dmitry, Lepikhin, Borgeaud, Bachem, Joulin, Andreev, Hardin, Dadashi, and Hussenot]{gemmateam2025gemma3technicalreport}
Gemma Team, Aishwarya Kamath, Johan Ferret, Shreya Pathak, Nino Vieillard, Ramona Merhej, Sarah Perrin, Tatiana Matejovicova, Alexandre Ramé, Morgane Rivière, Louis Rouillard, Thomas Mesnard, Geoffrey Cideron, Jean bastien Grill, Sabela Ramos, Edouard Yvinec, Michelle Casbon, Etienne Pot, Ivo Penchev, Gaël Liu, Francesco Visin, Kathleen Kenealy, Lucas Beyer, Xiaohai Zhai, Anton Tsitsulin, Robert Busa-Fekete, Alex Feng, Noveen Sachdeva, Benjamin Coleman, Yi~Gao, Basil Mustafa, Iain Barr, Emilio Parisotto, David Tian, Matan Eyal, Colin Cherry, Jan-Thorsten Peter, Danila Sinopalnikov, Surya Bhupatiraju, Rishabh Agarwal, Mehran Kazemi, Dan Malkin, Ravin Kumar, David Vilar, Idan Brusilovsky, Jiaming Luo, Andreas Steiner, Abe Friesen, Abhanshu Sharma, Abheesht Sharma, Adi~Mayrav Gilady, Adrian Goedeckemeyer, Alaa Saade, Alex Feng, Alexander Kolesnikov, Alexei Bendebury, Alvin Abdagic, Amit Vadi, András György, André~Susano Pinto, Anil Das, Ankur Bapna, Antoine Miech, Antoine Yang, Antonia Paterson, Ashish
  Shenoy, Ayan Chakrabarti, Bilal Piot, Bo~Wu, Bobak Shahriari, Bryce Petrini, Charlie Chen, Charline~Le Lan, Christopher~A. Choquette-Choo, CJ~Carey, Cormac Brick, Daniel Deutsch, Danielle Eisenbud, Dee Cattle, Derek Cheng, Dimitris Paparas, Divyashree~Shivakumar Sreepathihalli, Doug Reid, Dustin Tran, Dustin Zelle, Eric Noland, Erwin Huizenga, Eugene Kharitonov, Frederick Liu, Gagik Amirkhanyan, Glenn Cameron, Hadi Hashemi, Hanna Klimczak-Plucińska, Harman Singh, Harsh Mehta, Harshal~Tushar Lehri, Hussein Hazimeh, Ian Ballantyne, Idan Szpektor, Ivan Nardini, Jean Pouget-Abadie, Jetha Chan, Joe Stanton, John Wieting, Jonathan Lai, Jordi Orbay, Joseph Fernandez, Josh Newlan, Ju~yeong Ji, Jyotinder Singh, Kat Black, Kathy Yu, Kevin Hui, Kiran Vodrahalli, Klaus Greff, Linhai Qiu, Marcella Valentine, Marina Coelho, Marvin Ritter, Matt Hoffman, Matthew Watson, Mayank Chaturvedi, Michael Moynihan, Min Ma, Nabila Babar, Natasha Noy, Nathan Byrd, Nick Roy, Nikola Momchev, Nilay Chauhan, Noveen Sachdeva, Oskar
  Bunyan, Pankil Botarda, Paul Caron, Paul~Kishan Rubenstein, Phil Culliton, Philipp Schmid, Pier~Giuseppe Sessa, Pingmei Xu, Piotr Stanczyk, Pouya Tafti, Rakesh Shivanna, Renjie Wu, Renke Pan, Reza Rokni, Rob Willoughby, Rohith Vallu, Ryan Mullins, Sammy Jerome, Sara Smoot, Sertan Girgin, Shariq Iqbal, Shashir Reddy, Shruti Sheth, Siim Põder, Sijal Bhatnagar, Sindhu~Raghuram Panyam, Sivan Eiger, Susan Zhang, Tianqi Liu, Trevor Yacovone, Tyler Liechty, Uday Kalra, Utku Evci, Vedant Misra, Vincent Roseberry, Vlad Feinberg, Vlad Kolesnikov, Woohyun Han, Woosuk Kwon, Xi~Chen, Yinlam Chow, Yuvein Zhu, Zichuan Wei, Zoltan Egyed, Victor Cotruta, Minh Giang, Phoebe Kirk, Anand Rao, Kat Black, Nabila Babar, Jessica Lo, Erica Moreira, Luiz~Gustavo Martins, Omar Sanseviero, Lucas Gonzalez, Zach Gleicher, Tris Warkentin, Vahab Mirrokni, Evan Senter, Eli Collins, Joelle Barral, Zoubin Ghahramani, Raia Hadsell, Yossi Matias, D.~Sculley, Slav Petrov, Noah Fiedel, Noam Shazeer, Oriol Vinyals, Jeff Dean, Demis Hassabis,
  Koray Kavukcuoglu, Clement Farabet, Elena Buchatskaya, Jean-Baptiste Alayrac, Rohan Anil, Dmitry, Lepikhin, Sebastian Borgeaud, Olivier Bachem, Armand Joulin, Alek Andreev, Cassidy Hardin, Robert Dadashi, and Léonard Hussenot.
\newblock {Gemma 3 Technical Report}, 2025{\natexlab{a}}.
\newblock URL \url{https://arxiv.org/abs/2503.19786}.

\bibitem[Team et~al.(2025{\natexlab{b}})Team, Zeng, Lv, Zheng, Hou, Chen, Xie, Wang, Yin, Zeng, Zhang, Wang, Zhong, Liu, Lu, Cao, Zhang, Huang, Wei, Cheng, An, Niu, Wen, Bai, Du, Wang, Zhu, Zhang, Wen, Wu, Xu, Huang, Zhao, Cai, Yu, Li, Ge, Huang, Zhang, Xu, Zhu, Li, Yin, Lin, Yang, Jiang, Ai, Zhu, Wang, Pan, Wang, Sun, Li, Li, Hu, Zhang, Peng, Tai, Zhang, Wang, Yang, Liu, Zhao, Liu, Yan, Liu, Chen, Li, Zhao, Ren, Jiao, Zhao, Yan, Wang, Gui, Zhao, Liu, Li, Li, Lu, Wang, Yuan, Li, Du, Du, Liu, Zhi, Gao, Wang, Yang, Xu, Fan, Wu, Ding, Wang, Zhang, Li, Xu, Zhao, Zhai, Du, Dong, Lei, Tu, Yang, Lu, Li, Li, Shuang-Li, Yang, Yi, Yu, Tian, Wang, Yu, Tam, Liang, Liu, Wang, Jia, Gu, Ling, Wang, Fan, Pan, Zhang, Zhang, Fu, Zhang, Xu, Wu, Lu, Wang, Zhou, Pan, Zhang, Wang, Li, Su, Geng, Zhu, Yang, Li, Wu, Li, Liu, Wang, Li, Zhang, Liu, Yang, Zhou, Qiao, Feng, Liu, Zhang, Wang, Yao, Wang, Liu, Chai, Li, Zhao, Chen, Zhai, Xu, Huang, Wang, Li, Dong, and Tang]{5team2025glm45agenticreasoningcoding}
GLM Team, Aohan Zeng, Xin Lv, Qinkai Zheng, Zhenyu Hou, Bin Chen, Chengxing Xie, Cunxiang Wang, Da~Yin, Hao Zeng, Jiajie Zhang, Kedong Wang, Lucen Zhong, Mingdao Liu, Rui Lu, Shulin Cao, Xiaohan Zhang, Xuancheng Huang, Yao Wei, Yean Cheng, Yifan An, Yilin Niu, Yuanhao Wen, Yushi Bai, Zhengxiao Du, Zihan Wang, Zilin Zhu, Bohan Zhang, Bosi Wen, Bowen Wu, Bowen Xu, Can Huang, Casey Zhao, Changpeng Cai, Chao Yu, Chen Li, Chendi Ge, Chenghua Huang, Chenhui Zhang, Chenxi Xu, Chenzheng Zhu, Chuang Li, Congfeng Yin, Daoyan Lin, Dayong Yang, Dazhi Jiang, Ding Ai, Erle Zhu, Fei Wang, Gengzheng Pan, Guo Wang, Hailong Sun, Haitao Li, Haiyang Li, Haiyi Hu, Hanyu Zhang, Hao Peng, Hao Tai, Haoke Zhang, Haoran Wang, Haoyu Yang, He~Liu, He~Zhao, Hongwei Liu, Hongxi Yan, Huan Liu, Huilong Chen, Ji~Li, Jiajing Zhao, Jiamin Ren, Jian Jiao, Jiani Zhao, Jianyang Yan, Jiaqi Wang, Jiayi Gui, Jiayue Zhao, Jie Liu, Jijie Li, Jing Li, Jing Lu, Jingsen Wang, Jingwei Yuan, Jingxuan Li, Jingzhao Du, Jinhua Du, Jinxin Liu, Junkai Zhi,
  Junli Gao, Ke~Wang, Lekang Yang, Liang Xu, Lin Fan, Lindong Wu, Lintao Ding, Lu~Wang, Man Zhang, Minghao Li, Minghuan Xu, Mingming Zhao, Mingshu Zhai, Pengfan Du, Qian Dong, Shangde Lei, Shangqing Tu, Shangtong Yang, Shaoyou Lu, Shijie Li, Shuang Li, Shuang-Li, Shuxun Yang, Sibo Yi, Tianshu Yu, Wei Tian, Weihan Wang, Wenbo Yu, Weng~Lam Tam, Wenjie Liang, Wentao Liu, Xiao Wang, Xiaohan Jia, Xiaotao Gu, Xiaoying Ling, Xin Wang, Xing Fan, Xingru Pan, Xinyuan Zhang, Xinze Zhang, Xiuqing Fu, Xunkai Zhang, Yabo Xu, Yandong Wu, Yida Lu, Yidong Wang, Yilin Zhou, Yiming Pan, Ying Zhang, Yingli Wang, Yingru Li, Yinpei Su, Yipeng Geng, Yitong Zhu, Yongkun Yang, Yuhang Li, Yuhao Wu, Yujiang Li, Yunan Liu, Yunqing Wang, Yuntao Li, Yuxuan Zhang, Zezhen Liu, Zhen Yang, Zhengda Zhou, Zhongpei Qiao, Zhuoer Feng, Zhuorui Liu, Zichen Zhang, Zihan Wang, Zijun Yao, Zikang Wang, Ziqiang Liu, Ziwei Chai, Zixuan Li, Zuodong Zhao, Wenguang Chen, Jidong Zhai, Bin Xu, Minlie Huang, Hongning Wang, Juanzi Li, Yuxiao Dong, and Jie Tang.
\newblock {GLM-4.5: Agentic, Reasoning, and Coding (ARC) Foundation Models}, 2025{\natexlab{b}}.
\newblock URL \url{https://arxiv.org/abs/2508.06471}.

\bibitem[Team(2025)]{qwen3technicalreport}
Qwen Team.
\newblock {Qwen3 Technical Report}, 2025.
\newblock URL \url{https://arxiv.org/abs/2505.09388}.

\bibitem[Wang et~al.(2025)Wang, Zhao, Ding, Kuang, Shen, Tang, Jin, Wang, Li, Cao, and Cai]{wang2025meeseeksfeedbackdriveniterativeselfcorrection}
Jiaming Wang, Yunke Zhao, Peng Ding, Jun Kuang, Yibin Shen, Zhe Tang, Yilin Jin, ZongYu Wang, Xiaoyu Li, Xuezhi Cao, and Xunliang Cai.
\newblock {Meeseeks: A Feedback-Driven, Iterative Self-Correction Benchmark evaluating LLMs' Instruction Following Capability}.
\newblock 2025.
\newblock URL \url{https://arxiv.org/abs/2504.21625}.

\bibitem[Wang et~al.(2024)Wang, Wang, Liu, Chen, Yuan, Peng, and Ji]{wang2024mint}
Xingyao Wang, Zihan Wang, Jiateng Liu, Yangyi Chen, Lifan Yuan, Hao Peng, and Heng Ji.
\newblock {MINT}: Evaluating {LLM}s in multi-turn interaction with tools and language feedback.
\newblock In \emph{The Twelfth International Conference on Learning Representations}, 2024.
\newblock URL \url{https://openreview.net/forum?id=jp3gWrMuIZ}.

\bibitem[Wu et~al.(2025)Wu, Wang, Yu, Zhang, Chang, and Yu]{wu2025longmemeval}
Di~Wu, Hongwei Wang, Wenhao Yu, Yuwei Zhang, Kai-Wei Chang, and Dong Yu.
\newblock {LongMemEval: Benchmarking Chat Assistants on Long-Term Interactive Memory}.
\newblock In \emph{The Thirteenth International Conference on Learning Representations}, 2025.
\newblock URL \url{https://openreview.net/forum?id=pZiyCaVuti}.

\bibitem[Xia et~al.(2024)Xia, Xing, Du, Yang, Feng, Xu, Yin, and Xiong]{xia-etal-2024-fofo}
Congying Xia, Chen Xing, Jiangshu Du, Xinyi Yang, Yihao Feng, Ran Xu, Wenpeng Yin, and Caiming Xiong.
\newblock {{FOFO}: A Benchmark to Evaluate {LLM}s' Format-Following Capability}.
\newblock In Lun-Wei Ku, Andre Martins, and Vivek Srikumar (eds.), \emph{Proceedings of the 62nd Annual Meeting of the Association for Computational Linguistics (Volume 1: Long Papers)}, pp.\  680--699, Bangkok, Thailand, August 2024. Association for Computational Linguistics.
\newblock \doi{10.18653/v1/2024.acl-long.40}.
\newblock URL \url{https://aclanthology.org/2024.acl-long.40/}.

\bibitem[Yang et~al.(2025)Yang, Yu, Li, Liu, Huang, Huang, Jiang, Tu, Zhang, Zhou, Lin, Dang, Yang, Yu, Li, Sun, Zhu, Men, He, Xu, Yin, Yu, Qiu, Ren, Yang, Li, Xu, and Zhang]{qwen2.5-1m}
An~Yang, Bowen Yu, Chengyuan Li, Dayiheng Liu, Fei Huang, Haoyan Huang, Jiandong Jiang, Jianhong Tu, Jianwei Zhang, Jingren Zhou, Junyang Lin, Kai Dang, Kexin Yang, Le~Yu, Mei Li, Minmin Sun, Qin Zhu, Rui Men, Tao He, Weijia Xu, Wenbiao Yin, Wenyuan Yu, Xiafei Qiu, Xingzhang Ren, Xinlong Yang, Yong Li, Zhiying Xu, and Zipeng Zhang.
\newblock {Qwen2.5-1M Technical Report}.
\newblock \emph{arXiv preprint arXiv:2501.15383}, 2025.

\bibitem[Zhao et~al.(2024)Zhao, Ren, Hessel, Cardie, Choi, and Deng]{zhao2024wildchat}
Wenting Zhao, Xiang Ren, Jack Hessel, Claire Cardie, Yejin Choi, and Yuntian Deng.
\newblock {WildChat: 1M Chat{GPT} Interaction Logs in the Wild}.
\newblock In \emph{The Twelfth International Conference on Learning Representations}, 2024.
\newblock URL \url{https://openreview.net/forum?id=Bl8u7ZRlbM}.

\bibitem[Zheng et~al.(2023)Zheng, Chiang, Sheng, Zhuang, Wu, Zhuang, Lin, Li, Li, Xing, Zhang, Gonzalez, and Stoica]{10.5555/3666122.3668142}
Lianmin Zheng, Wei-Lin Chiang, Ying Sheng, Siyuan Zhuang, Zhanghao Wu, Yonghao Zhuang, Zi~Lin, Zhuohan Li, Dacheng Li, Eric~P. Xing, Hao Zhang, Joseph~E. Gonzalez, and Ion Stoica.
\newblock {Judging LLM-as-a-judge with MT-bench and Chatbot Arena}.
\newblock In \emph{Proceedings of the 37th International Conference on Neural Information Processing Systems}, NIPS '23, Red Hook, NY, USA, 2023. Curran Associates Inc.

\bibitem[Zheng et~al.(2024)Zheng, Chiang, Sheng, Li, Zhuang, Wu, Zhuang, Li, Lin, Xing, Gonzalez, Stoica, and Zhang]{zheng2024lmsyschatm}
Lianmin Zheng, Wei-Lin Chiang, Ying Sheng, Tianle Li, Siyuan Zhuang, Zhanghao Wu, Yonghao Zhuang, Zhuohan Li, Zi~Lin, Eric Xing, Joseph~E. Gonzalez, Ion Stoica, and Hao Zhang.
\newblock {LMSYS}-chat-1m: A large-scale real-world {LLM} conversation dataset.
\newblock In \emph{The Twelfth International Conference on Learning Representations}, 2024.
\newblock URL \url{https://openreview.net/forum?id=BOfDKxfwt0}.

\bibitem[Zhou et~al.(2023)Zhou, Lu, Mishra, Brahma, Basu, Luan, Zhou, and Hou]{zhou2023instructionfollowingevaluationlargelanguage}
Jeffrey Zhou, Tianjian Lu, Swaroop Mishra, Siddhartha Brahma, Sujoy Basu, Yi~Luan, Denny Zhou, and Le~Hou.
\newblock {Instruction-Following Evaluation for Large Language Models}.
\newblock 2023.
\newblock URL \url{https://arxiv.org/abs/2311.07911}.

\end{thebibliography}
\bibliographystyle{colm2026_conference}

\newpage

\appendix

\section{Constraint Curation Pipeline}
\label{app:constraints-pipeline}

\subsection{Prompt Templates}

\begin{figure*}[ht!]
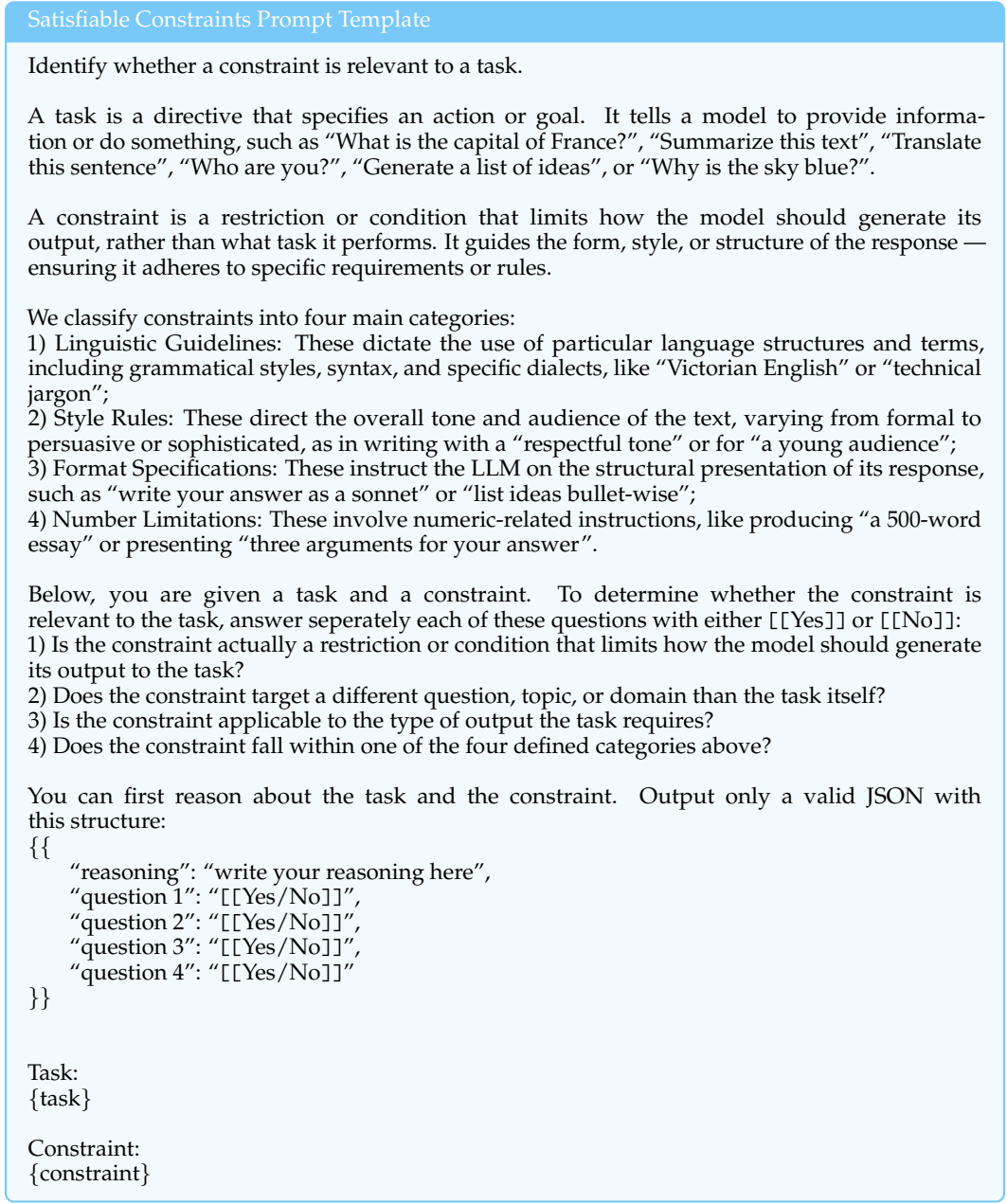

\centering
\begin{tcolorbox}[
  colback=bblue!10, 
  colframe=bblue!100, 
  colbacktitle=bblue!100, 
  coltitle=white,
  fonttitle=\bfseries,
  title={Satisfiable Constraints Prompt Template},
  boxrule=0.8pt,
  left=2mm,right=2mm,top=1mm,bottom=1mm,
  fontupper=\footnotesize,
  fonttitle=\small,
]
  %

Identify whether a constraint is relevant to a task. \\
\\
A task is a directive that specifies an action or goal. It tells a model to provide information or do something, such as ``What is the capital of France?'', ``Summarize this text'', ``Translate this sentence'', ``Who are you?'', ``Generate a list of ideas'', or ``Why is the sky blue?''. \\
\\
A constraint is a restriction or condition that limits how the model should generate its output, rather than what task it performs. It guides the form, style, or structure of the response — ensuring it adheres to specific requirements or rules. \\
\\
We classify constraints into four main categories: \\
1) Linguistic Guidelines: These dictate the use of particular language structures and terms, including grammatical styles, syntax, and specific dialects, like ``Victorian English'' or ``technical jargon''; \\
2) Style Rules: These direct the overall tone and audience of the text, varying from formal to persuasive or sophisticated, as in writing with a ``respectful tone'' or for ``a young audience'';  \\
3) Format Specifications: These instruct the LLM on the structural presentation of its response, such as ``write your answer as a sonnet'' or ``list ideas bullet-wise''; \\
4) Number Limitations: These involve numeric-related instructions, like producing ``a 500-word essay'' or presenting ``three arguments for your answer''. \\
\\
Below, you are given a task and a constraint. To determine whether the constraint is relevant to the task, answer seperately each of these questions with either \texttt{[[}Yes\texttt{]]} or \texttt{[[}No\texttt{]]}: \\
1) Is the constraint actually a restriction or condition that limits how the model should generate its output to the task? \\
2) Does the constraint target a different question, topic, or domain than the task itself? \\
3) Is the constraint applicable to the type of output the task requires? \\
4) Does the constraint fall within one of the four defined categories above? \\
\\
You can first reason about the task and the constraint. Output only a valid JSON with this structure: \\
\{\{ \\
  \hspace*{0.5cm}  ``reasoning'': ``write your reasoning here'', \\
  \hspace*{0.5cm}  ``question 1'': ``\texttt{[[}Yes/No\texttt{]]}'', \\
  \hspace*{0.5cm}  ``question 2'': ``\texttt{[[}Yes/No\texttt{]]}'', \\
  \hspace*{0.5cm}  ``question 3'': ``\texttt{[[}Yes/No\texttt{]]}'', \\
  \hspace*{0.5cm}  ``question 4'': ``\texttt{[[}Yes/No\texttt{]]}'' \\
\}\} \\
\\
\\
Task: \\
\{task\} \\
\\
Constraint: \\
\{constraint\}
\end{tcolorbox}
\caption{Prompt template used to identify satisfiable constraints with LM judges.}
\label{fig:satisfiable-constraints-prompt-template}
\end{figure*}

\begin{figure*}[ht!]
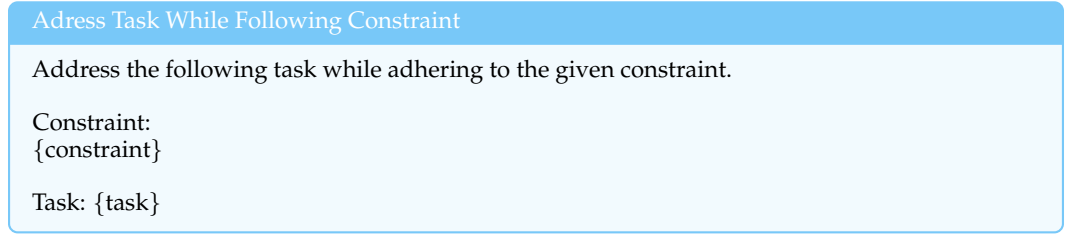

\centering
\begin{tcolorbox}[
  colback=bblue!10, 
  colframe=bblue!100, 
  colbacktitle=bblue!100, 
  coltitle=white,
  fonttitle=\bfseries,
  title={Adress Task While Following Constraint},
  boxrule=0.8pt,
  left=2mm,right=2mm,top=1mm,bottom=1mm,
  fontupper=\footnotesize,
  fonttitle=\small,
]
Address the following task while adhering to the given constraint. \\
\\
Constraint: \\
\{constraint\} \\
\\
Task:
\{task\}
\end{tcolorbox}
\caption{Prompt template used to generate model responses to a task under a specified constraint.}
\label{fig:address-task-follow-constraint}
\end{figure*}

\begin{figure*}[ht!]
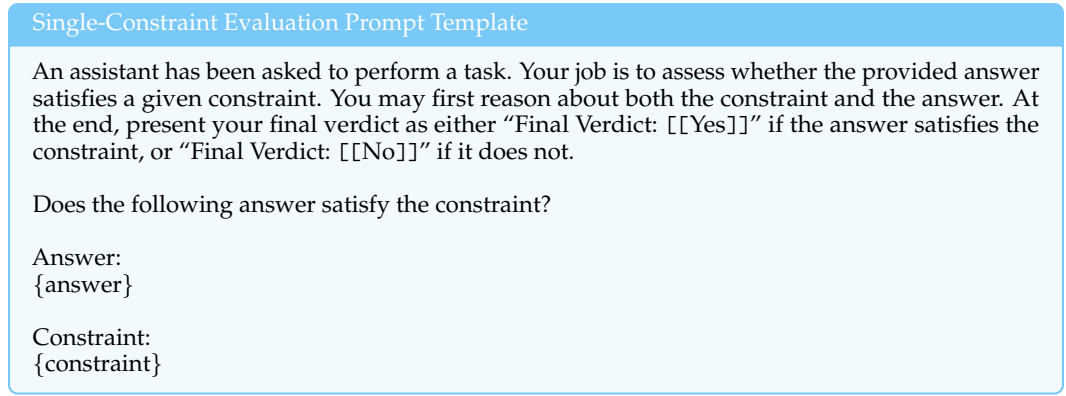

\centering
\begin{tcolorbox}[
  colback=bblue!10, 
  colframe=bblue!100, 
  colbacktitle=bblue!100, 
  coltitle=white,
  fonttitle=\bfseries,
  title={Single-Constraint Evaluation Prompt Template},
  boxrule=0.8pt,
  left=2mm,right=2mm,top=1mm,bottom=1mm,
  fontupper=\footnotesize,
  fonttitle=\small,
]

An assistant has been asked to perform a task. Your job is to assess whether the provided answer satisfies a given constraint. You may first reason about both the constraint and the answer. At the end, present your final verdict as either ``Final Verdict: \texttt{[[}Yes\texttt{]]}'' if the answer satisfies the constraint, or ``Final Verdict: \texttt{[[}No\texttt{]]}'' if it does not. \\
\\
Does the following answer satisfy the constraint? \\
\\
Answer: \\
\{answer\} \\
\\
Constraint: \\
\{constraint\}
\end{tcolorbox}
\caption{Prompt template used by LM judges to assess whether an answer satisfies a single constraint.}
\label{fig:judge-single-constraint-prompt-template}
\end{figure*}

Figures~\ref{fig:constraint-extraction-prompt-template} and \ref{fig:satisfiable-constraints-prompt-template} show the prompt templates used to extract constraints from conversational datasets and identify satisfiable constraints, respectively. In the phases for identifying trivial and subjective constraints, we used the prompt templates shown in \Cref{fig:judge-single-constraint-prompt-template} and \ref{fig:address-task-follow-constraint}. To sample model responses to tasks without specifying any constraint in the trivial phase, we did use any template, just sent the tasks directly to the models. \Cref{fig:constraint-extraction-prompt-template,fig:satisfiable-constraints-prompt-template} show the prompt templates used to extract constraints from conversational datasets and to identify satisfiable constraints, respectively. For the triviality and subjectivity phases, we used the prompt templates shown in \Cref{fig:judge-single-constraint-prompt-template,fig:address-task-follow-constraint}. In the triviality phase, when sampling model responses without imposing any constraint, we submitted the tasks directly to the models without using any template.

\begin{tcolorbox}[
  breakable,
  colback=bblue!10, 
  colframe=bblue!100, 
  colbacktitle=bblue!100, 
  coltitle=white,
  fonttitle=\bfseries,
  title={Constraint Extraction Prompt Template},
  boxrule=0.8pt,
  left=2mm,right=2mm,top=1mm,bottom=1mm,
  fontupper=\footnotesize,
  fonttitle=\small,
  %
]
Identify tasks and constraints in user prompts. \\
\\
A task is a directive that specifies an action or goal. It tells a model to provide information or do something, such as "What is the capital of France?", "Summarize this text", "Translate this sentence", "Who are you?", "Generate a list of ideas", or "Why is the sky blue?". \\
\\
A user prompt might contain no task! It can simply be a statement or expression, without any specific request for information or action. Examples of such user prompts include greetings ("Hello!"), expressions of emotion ("I'm feeling great today."), or sharing information ("I went to the park yesterday."). \\
\\
A constraint is a restriction or condition that limits how the model should generate its output, rather than what task it performs. It guides the form, style, or structure of the response — ensuring it adheres to specific requirements or rules. \\
\\
We classify constraints into four main categories: \\
1) Linguistic Guidelines: These dictate the use of particular language structures and terms, including grammatical styles, syntax, and specific dialects, like "Victorian English" or "technical jargon"; \\
2) Style Rules: These direct the overall tone and audience of the text, varying from formal to persuasive or sophisticated, as in writing with a "respectful tone" or for "a young audience"; \\
3) Format Specifications: These instruct the LLM on the structural presentation of its response, such as "write your answer as a sonnet" or "list ideas bullet-wise"; \\
4) Number Limitations: These involve numeric-related instructions, like producing "a 500-word essay" or presenting "three arguments for your answer". \\
\\
Below, you are given a sequence of user prompts taken from a conversation. Your job is to identify all tasks and constraints in the user prompts. In addition, classify all constraints into their categories. Each constraint should be classified into one and only one of the four categories listed above. \\
\\
You can first reason about the user prompts and their context. At the end, present your final answer as a valid json output, ie, as a list of dictionaries where each dictionary contains the turn number, the task defined in the user turn (if any, otherwise ""),and a list of dictionaries for the constraints found (if any, otherwise []), where each constraint dictionary contains the constraint and the constraint type. \\
\\
For example: \\
\\
User prompts: \\
``Turn 1: \\
Hello! \\
\\
Turn 2: \\
I want to write an email to my boss about the crazy amount of meetings he's scheduling. 
Write me a formal email of 300 words to explain the situation and ask for him to be more understanding on the number of meetings he schedules. \\
\\
Turn 3: \\
Could you make sure to include some suggestions in a bullet-point list on how to manage meetings better? \\
\\
Turn 4: \\
Rewrite the email in a more polite tone.'' \\
\\
Output: \\
\texttt{[} \\
  \hspace*{0.5cm}  \{\{ \\
  \hspace*{0.5cm}    \hspace*{0.5cm}  ``turn'': 1, \\
  \hspace*{0.5cm}    \hspace*{0.5cm}  ``task'': ``'', \\
  \hspace*{0.5cm}    \hspace*{0.5cm}  ``constraints'': \texttt{[ ]} \\
  \hspace*{0.5cm}  \}\}, \\
  \hspace*{0.5cm}  \{\{ \\
  \hspace*{0.5cm}    \hspace*{0.5cm}  ``turn'': 2, \\
  \hspace*{0.5cm}    \hspace*{0.5cm}  ``task'': ``I want to write an email to my boss about the crazy amount of meetings he's scheduling. Write me an email to explain the situation and ask for him to be more understanding on the number of meetings he schedules.'', \\
  \hspace*{0.5cm}    \hspace*{0.5cm}  ``constraints'': [ \\
  \hspace*{0.5cm}    \hspace*{0.5cm}    \hspace*{0.5cm}  \{\{ \\
  \hspace*{0.5cm}    \hspace*{0.5cm}    \hspace*{0.5cm}    \hspace*{0.5cm}  ``constraint'': ``Write in formal tone'', \\
  \hspace*{0.5cm}    \hspace*{0.5cm}    \hspace*{0.5cm}    \hspace*{0.5cm}  ``type'': ``Style Rules'' \\
  \hspace*{0.5cm}    \hspace*{0.5cm}    \hspace*{0.5cm}  \}\}, \\
  \hspace*{0.5cm}    \hspace*{0.5cm}    \hspace*{0.5cm}  \{\{ \\
  \hspace*{0.5cm}    \hspace*{0.5cm}    \hspace*{0.5cm}    \hspace*{0.5cm}  ``constraint'': ``Write in 300 words'', \\
  \hspace*{0.5cm}    \hspace*{0.5cm}    \hspace*{0.5cm}    \hspace*{0.5cm}  ``type'': ``Number Limitations'' \\
  \hspace*{0.5cm}    \hspace*{0.5cm}    \hspace*{0.5cm}  \}\} \\
  \hspace*{0.5cm}    \hspace*{0.5cm}  \texttt{]} \\
  \hspace*{0.5cm}  \}\}, \\
  \hspace*{0.5cm}  \{\{ \\
  \hspace*{0.5cm}    \hspace*{0.5cm}  ``turn'': 3, \\
  \hspace*{0.5cm}    \hspace*{0.5cm}  ``task'': ``Make sure to include some suggestions on how to manage meetings better.'', \\
  \hspace*{0.5cm}    \hspace*{0.5cm}  ``constraints'': [ \\
  \hspace*{0.5cm}    \hspace*{0.5cm}    \hspace*{0.5cm}  \{\{ \\
  \hspace*{0.5cm}    \hspace*{0.5cm}    \hspace*{0.5cm}    \hspace*{0.5cm}  ``constraint'': ``Write a bullet-point list'', \\
  \hspace*{0.5cm}    \hspace*{0.5cm}    \hspace*{0.5cm}    \hspace*{0.5cm}  ``type'': ``Format Specifications'' \\
  \hspace*{0.5cm}    \hspace*{0.5cm}    \hspace*{0.5cm}    \hspace*{0.5cm}    \hspace*{0.5cm}  \}\} \\
  \hspace*{0.5cm}    \hspace*{0.5cm}  \texttt{]} \\
  \hspace*{0.5cm}  \}\}, \\
  \hspace*{0.5cm}  \{\{ \\
  \hspace*{0.5cm}    \hspace*{0.5cm}  ``turn'': 4, \\
  \hspace*{0.5cm}    \hspace*{0.5cm}  ``task'': ``Rewrite the email.'', \\
  \hspace*{0.5cm}    \hspace*{0.5cm}  ``constraints'': [ \\
  \hspace*{0.5cm}    \hspace*{0.5cm}    \hspace*{0.5cm}  \{\{ \\
  \hspace*{0.5cm}    \hspace*{0.5cm}    \hspace*{0.5cm}    \hspace*{0.5cm}  ``constraint'': ``Write in a more polite tone'', \\
  \hspace*{0.5cm}    \hspace*{0.5cm}    \hspace*{0.5cm}    \hspace*{0.5cm}  ``type'': ``Style Rules'' \\
  \hspace*{0.5cm}    \hspace*{0.5cm}    \hspace*{0.5cm}  \}\} \\
  \hspace*{0.5cm}    \hspace*{0.5cm}  \texttt{]} \\
  \hspace*{0.5cm}  \}\} \\
\texttt{]} \\
\\
User prompts: \\
``\{user\_turn\}'' \\
\\
Output:
\end{tcolorbox}
\captionof{figure}{Prompt template used to extract constraints from datasets of conversations.}
\label{fig:constraint-extraction-prompt-template}

\newpage

\subsection{Satisfiable, Trivial, and Subjective Thresholds}

\begin{figure*}[ht!]
\centering
\includegraphics[width=1.0\textwidth]{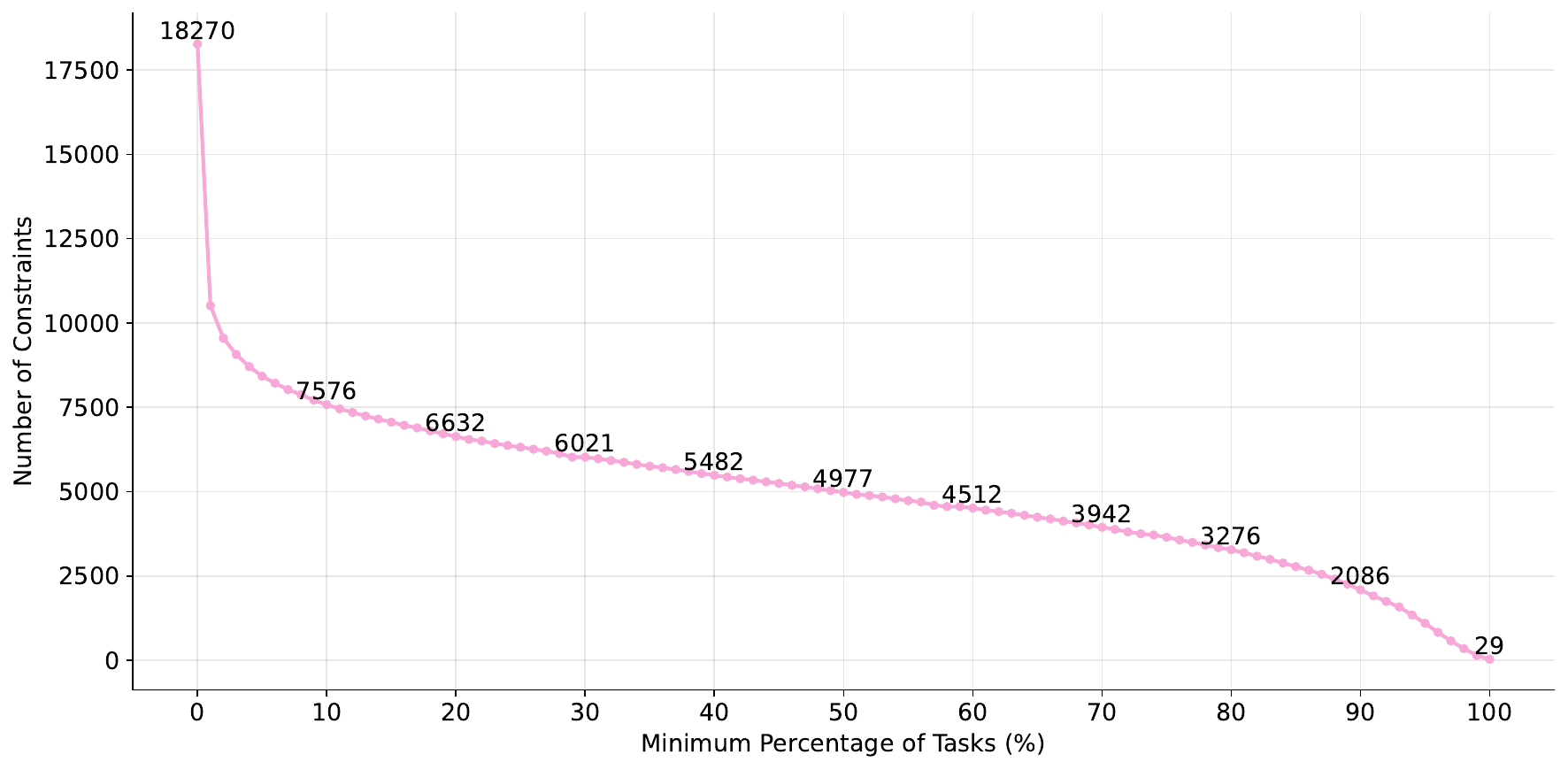}
\caption{Number of constraints classified as satisfied by all three judges as a function of the minimum percentage of task contexts in which they are classified as such.}
\label{fig:satisfiability-plot}
\end{figure*}

\begin{figure*}[ht!]
\centering
\includegraphics[width=1.0\textwidth]{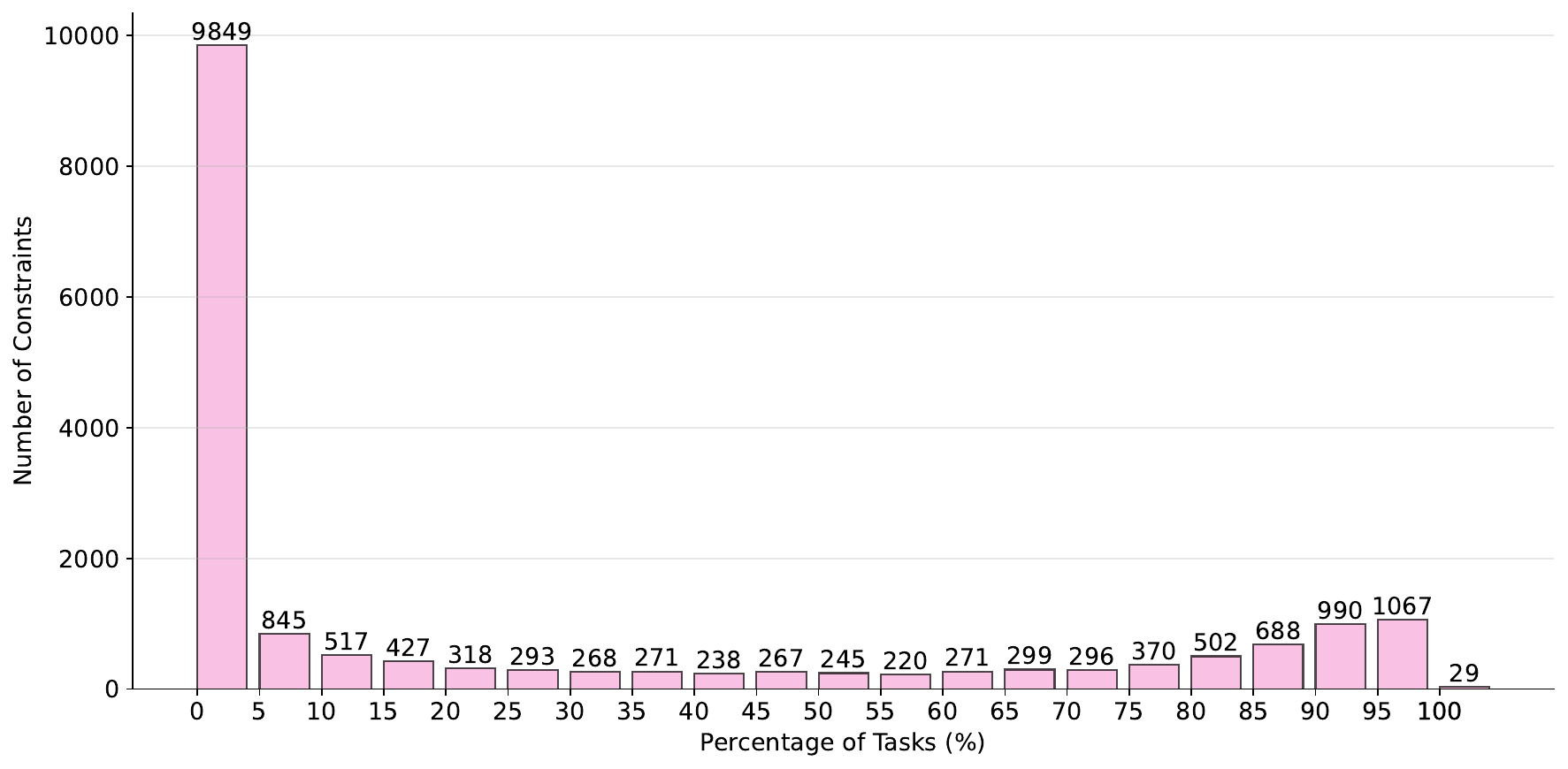}
\caption{Distribution of constraints by the percentage of task contexts in which they are classified as satisfied by all three judges.}
\label{fig:satisfiability-histogram}
\end{figure*}

\begin{figure*}[ht!]
\centering
\includegraphics[width=1.0\textwidth]{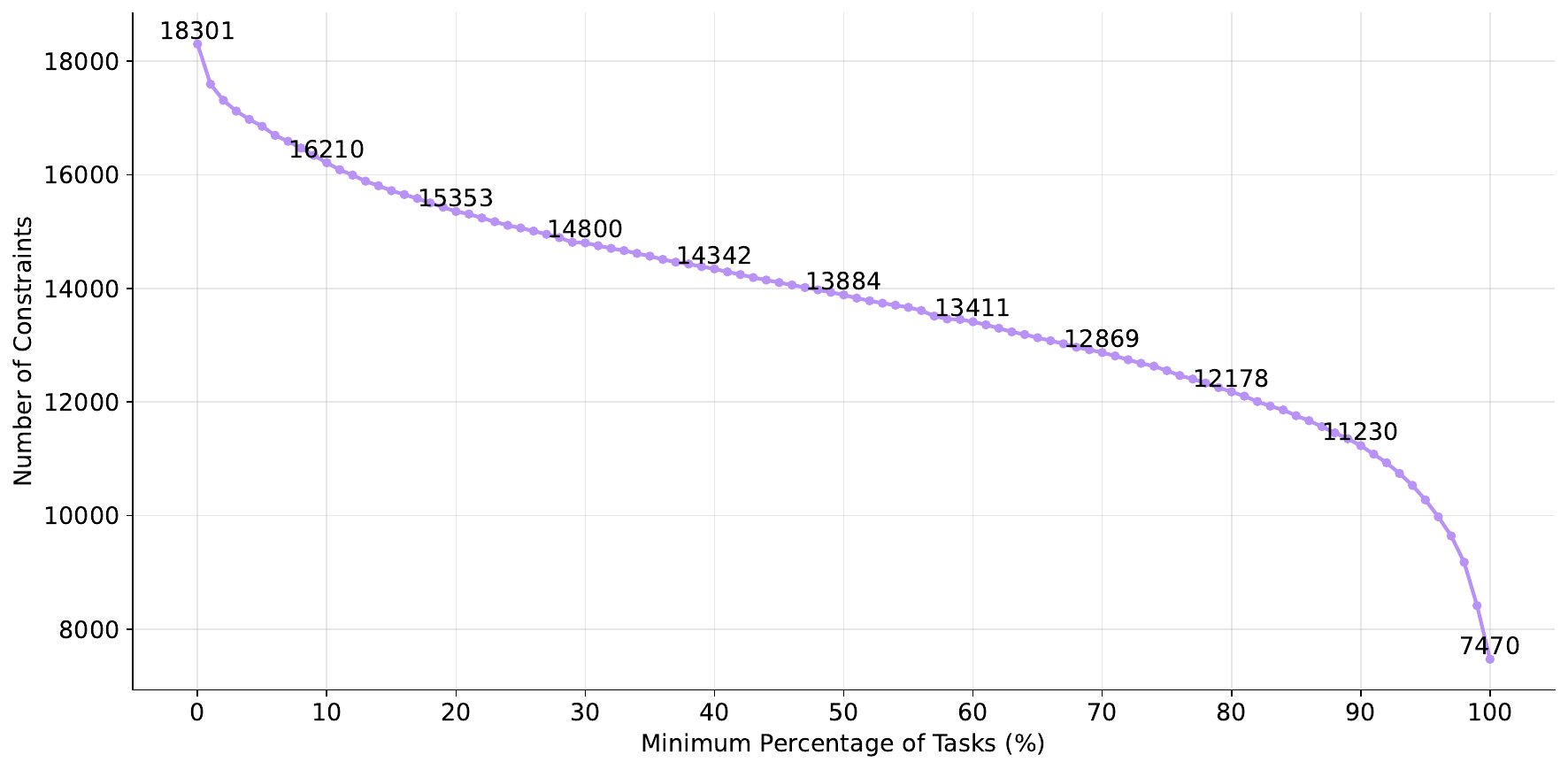}
\caption{Number of constraints classified as non-trivial by all three judges as a function of the minimum percentage of task contexts in which they are classified as such.}
\label{fig:trivial-plot}
\end{figure*}

\begin{figure*}[ht!]
\centering
\includegraphics[width=1.0\textwidth]{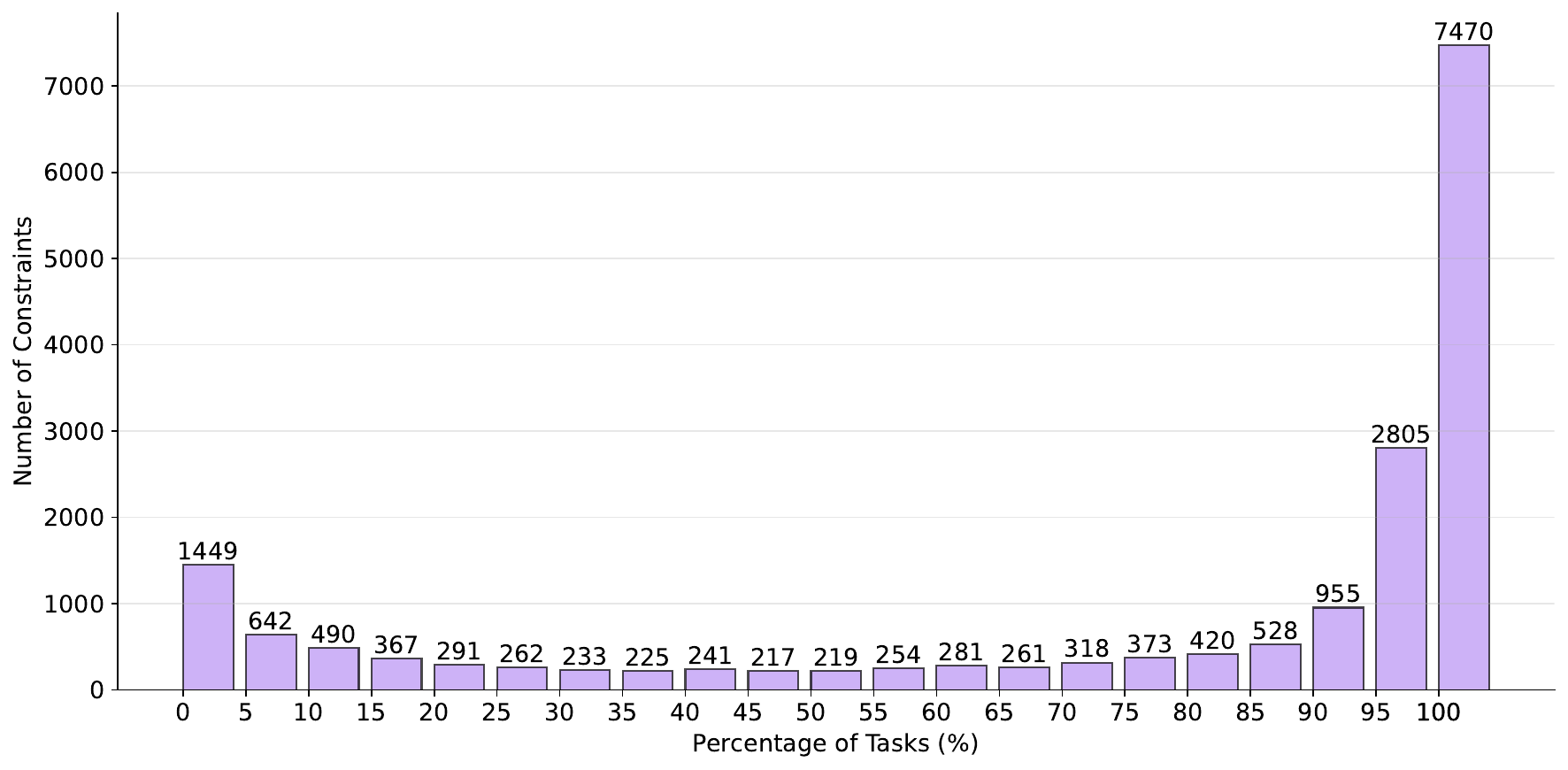}
\caption{Distribution of constraints by the percentage of task contexts in which they are classified as non-trivial by all three judges.}
\label{fig:trivial-histogram}
\end{figure*}

\begin{figure*}[ht!]
\centering
\includegraphics[width=1.0\textwidth]{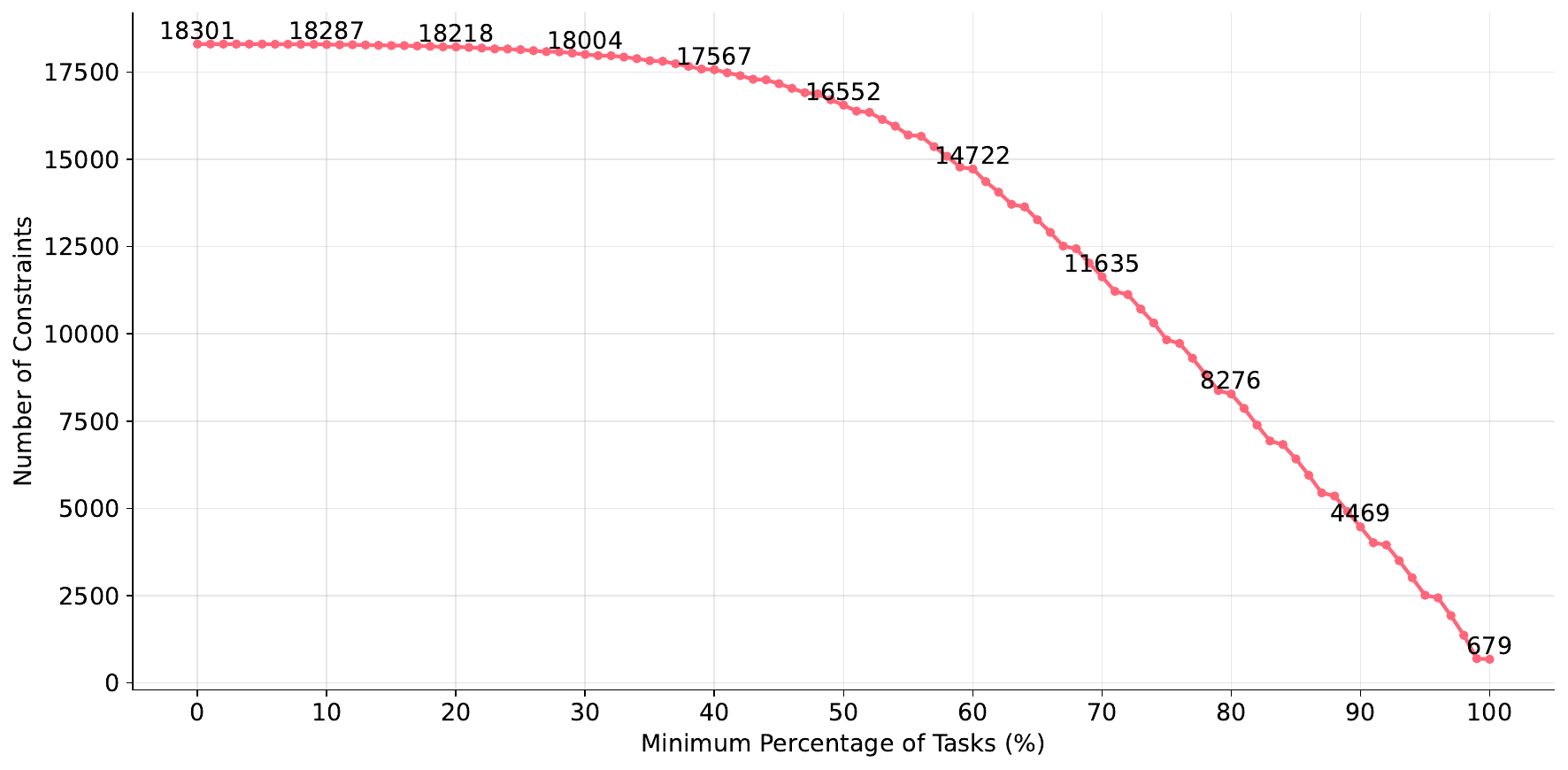}
\caption{Number of constraints classified as non-subjective by all three judges as a function of the minimum percentage of task contexts in which they are classified as such.}
\label{fig:subjectivity-plot}
\end{figure*}

\begin{figure*}[ht!]
\centering
\includegraphics[width=1.0\textwidth]{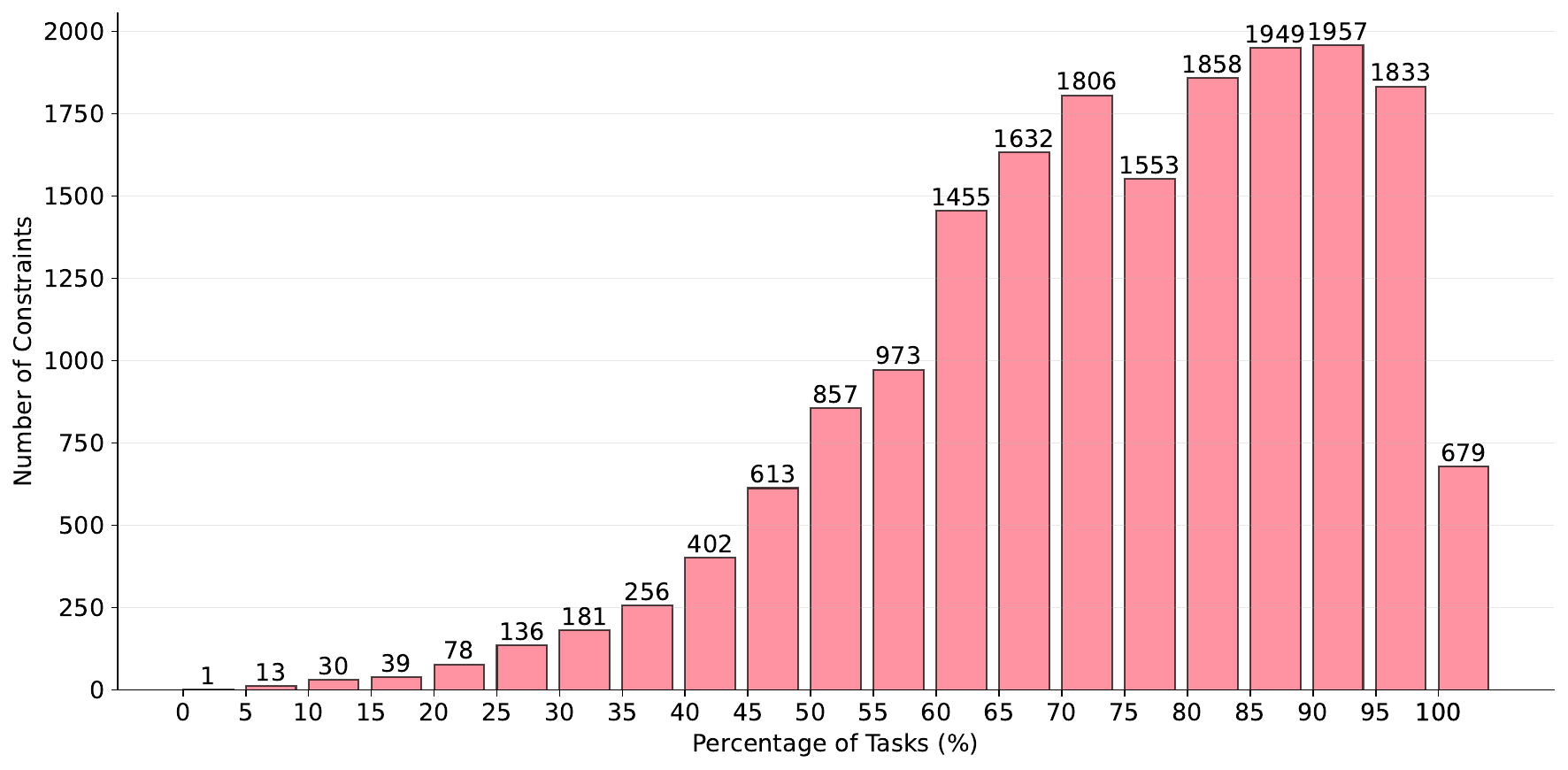}
\caption{Distribution of constraints by the percentage of task contexts in which they are classified as non-subjective by all three judges.}
\label{fig:subjectivity-histogram}
\end{figure*}

In the satisfiability, triviality, and subjectivity phases, each constraint was evaluated across $100$ different task contexts. A constraint was classified as satisfiable (respectively non-trivial or non-subjective) if each judge classified it as such in at least $X\%$ of the evaluated task contexts. \Cref{fig:satisfiability-plot,fig:trivial-plot,fig:subjectivity-plot} present the number of constraints classified as satisfiable, non-trivial, and non-subjective, respectively, for different values of $X$. \Cref{fig:satisfiability-histogram,fig:trivial-histogram,fig:subjectivity-histogram} show the distribution of constraints across $5\%$ performance intervals. 

Our final pool includes only constraints classified as satisfiable, non-trivial, and non-subjective in at least $70\%$ of the evaluated task contexts by each judge (not necessarily the same contexts across phases). This threshold balances robustness to task variability with the need to maintain sufficient constraint diversity, while ensuring a reasonably large and reliable constraint set.

\section{Creating Tuples of Constraints}
\label{app:creating-tuples-constraints}

\subsection{Prompt Templates}

We focused on creating tuples of $3$ constraints that can be simultaneously satisfied. We achieved this through a two-step process. First, we randomly sampled tuples of $3$ constraints and paired them with $100$ tasks. Each tuple–task pair was evaluated by multiple judges, and we retained only those tuples for which all judges agreed that the constraints can be jointly satisfied in at least $70\%$ of the task contexts---these are satisfiable tuples. We used the prompt template shown in \Cref{fig:satisfiable-tuples-prompt-template}. Second, we paired each satisfiable tuple again with $100$ tasks and prompted various models to generate responses for such pairs. For each tuple–task pair, we checked whether there existed at least one response satisfying all constraints, as verified unanimously by the judges. We used the prompt templates shown in \Cref{fig:address-task-follow-tuple,fig:judge-single-constraint-prompt-template}.

\begin{figure*}[ht!]
\centering
\begin{tcolorbox}[
  colback=bblue!10, 
  colframe=bblue!100, 
  colbacktitle=bblue!100, 
  coltitle=white,
  fonttitle=\bfseries,
  title={Satisfiable Tuples Prompt Template},
  boxrule=0.8pt,
  left=2mm,right=2mm,top=1mm,bottom=1mm,
  fontupper=\footnotesize,
  fonttitle=\small,
]

You will receive a task and a list of constraints. Analyze whether the constraints can all be followed simultaneously by a single response to the task without contradiction. You can first reason about the task, the constraints, and their possible contradictions. At the end, reply with ``Final Verdict: \texttt{[[}Yes\texttt{]]}'' if they are jointly compatible, otherwise reply with ``Final Verdict: \texttt{[[}No\texttt{]]}''. \\
\\
Task: \\
``\{task\}'' \
\\
Constraints: \\
``\{constraints\}''
\end{tcolorbox}
\caption{Prompt template used to identify satisfiable tuples with LM judges.}
\label{fig:satisfiable-tuples-prompt-template}
\end{figure*}

\begin{figure*}[ht!]
\centering
\begin{tcolorbox}[
  colback=bblue!10, 
  colframe=bblue!100, 
  colbacktitle=bblue!100, 
  coltitle=white,
  fonttitle=\bfseries,
  title={Adress Task While Following Tuple of Constraints},
  boxrule=0.8pt,
  left=2mm,right=2mm,top=1mm,bottom=1mm,
  fontupper=\footnotesize,
  fonttitle=\small,
]

Address the following task while adhering to all the given constraints. \\
\\
Constraints: \\
\{constraints\} \\
\\
Task: \\
\{task\}
\end{tcolorbox}
\caption{Prompt template used to generate model responses to a task under a specified tuple of constraints.}
\label{fig:address-task-follow-tuple}
\end{figure*}

\subsection{Cut-Off Threshold}

\begin{figure*}[ht!]
\centering
\includegraphics[width=1.0\textwidth]{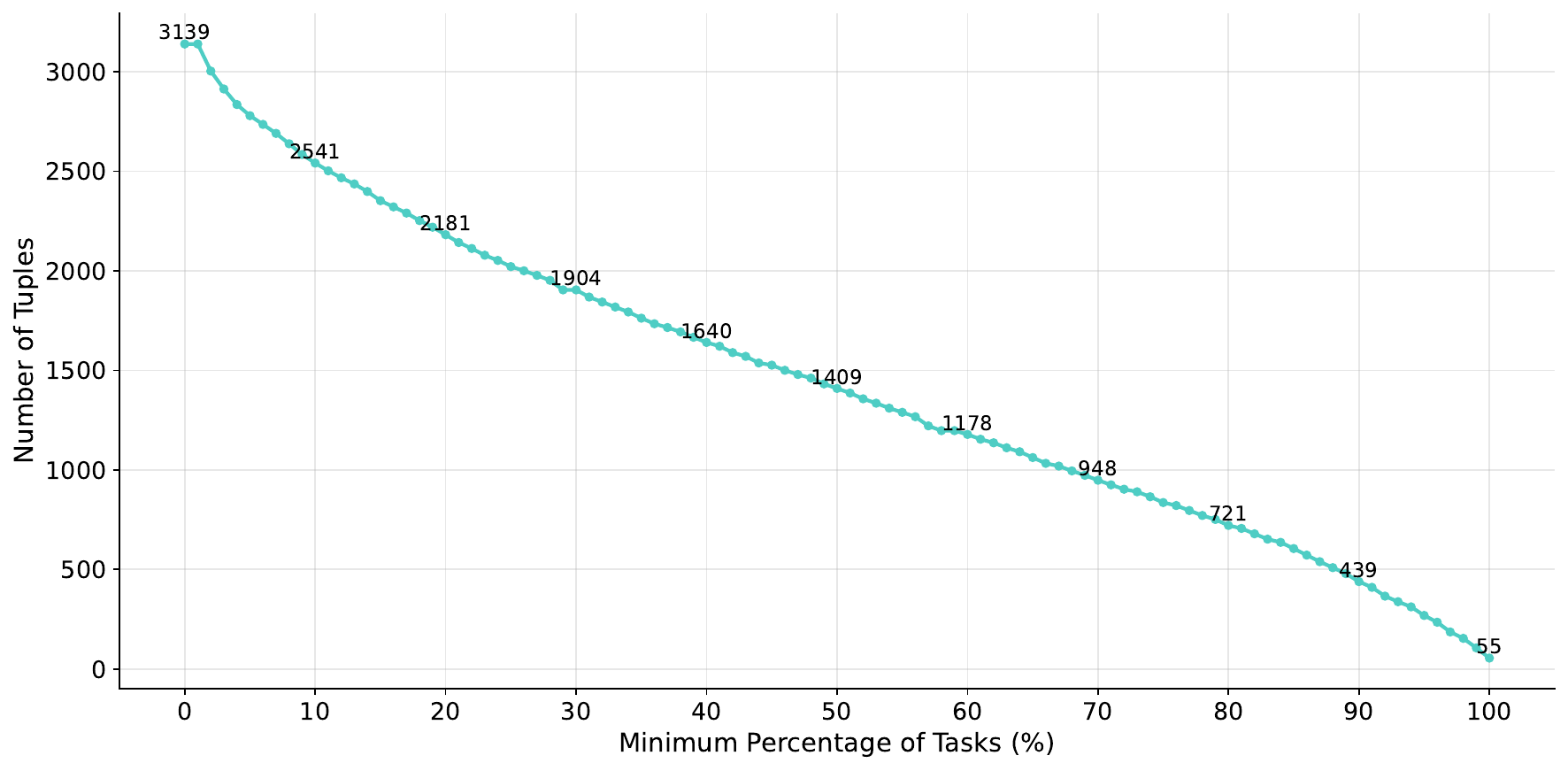}
\caption{Number of tuples as a function of the minimum percentage of task contexts for which we found at least one answer satisfying all constraints of the tuple, as determined by all three judges.}
\label{fig:tuples-plot}
\end{figure*}

\begin{figure*}[ht!]
\centering
\includegraphics[width=1.0\textwidth]{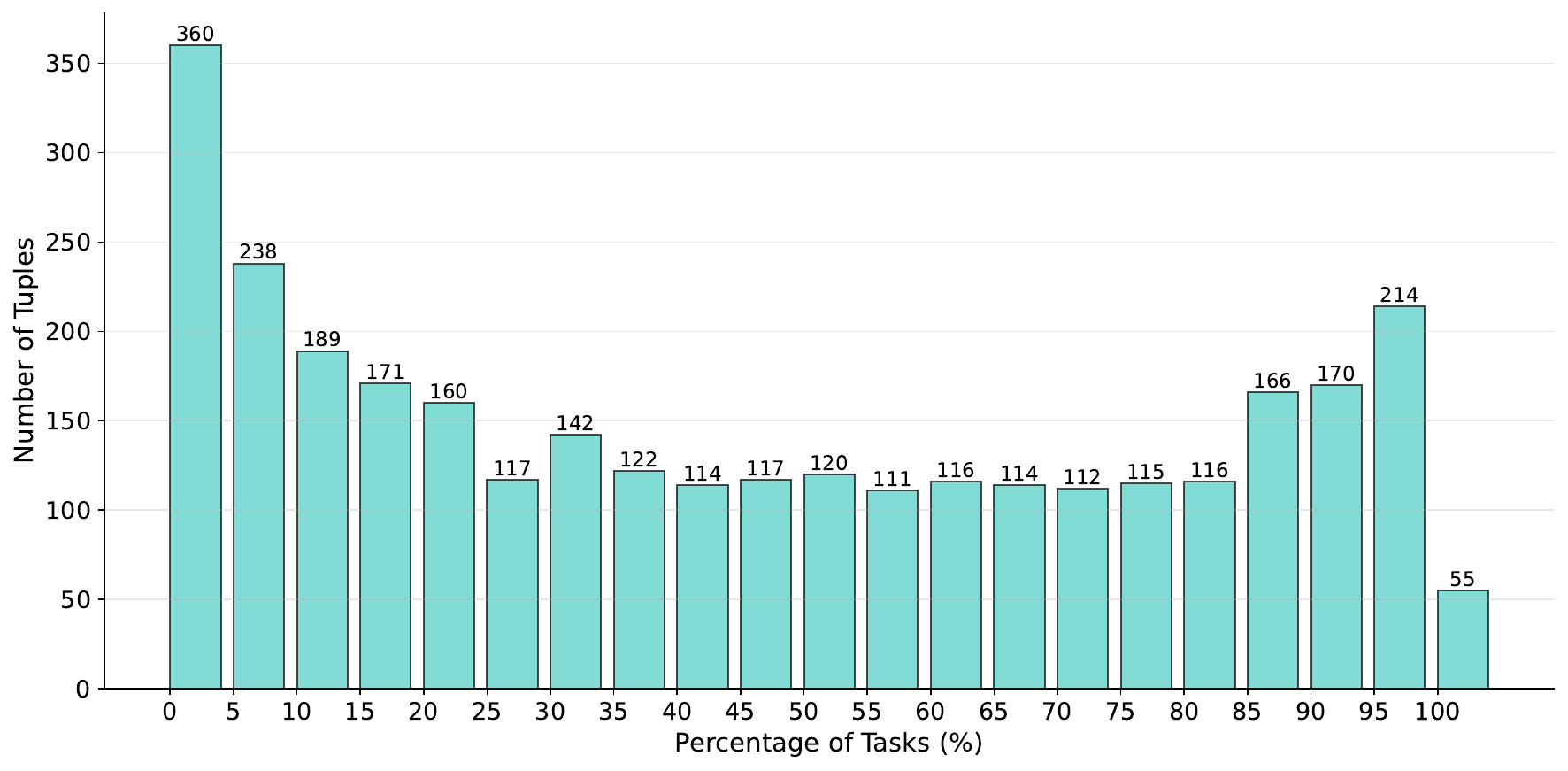}
\caption{Distribution of tuples by the percentage of task contexts for which we found at least one answer satisfying all constraints of the tuple, as determined by all three judges.}
\label{fig:tuples-histogram}
\end{figure*}

Figure~\ref{fig:tuples-plot} presents the number of satisfiable tuples for which we found at least one answer satisfying all constraints of the tuple, as classified by all three judges, for a varying number of task contexts. \Cref{fig:tuples-histogram} shows the distribution of tuples across $5\%$ performance intervals. 

Our final pool includes all satisfiable tuples for which we found, for at least $70\%$ of the task contexts analyzed, an answer satisfying all constraints of the tuple, as determined by all three judges. This threshold ensures a reasonably large and reliable set of $948$ constraint tuples, as well as robustness to task variability.

\section{Final Pool of Constraints and Tuples}

\subsection{Constraint Categories}

We adopt four main categories of constraints defined by \cite{qin-etal-2024-infobench}:
\begin{itemize}
    \item \textbf{Linguistic Guidelines.} These impose requirements on the language of the response, including vocabulary choice, grammatical constructions, or adherence to specific linguistic varieties (e.g., \textit{``use passive voice throughout''}, \textit{``avoid technical terminology''}).
    \item \textbf{Style Rules.} These govern the overall tone or intended audience of the response (e.g., \textit{``write in a neutral and objective voice''}, \textit{``explain the answer as if speaking to a child''}).
    \item \textbf{Format Specifications.} These specify the structural presentation of the response, determining how information should be arranged or displayed (e.g., \textit{``present a numbered list''}, \textit{``separate the response into clearly labeled sections''}).
    \item \textbf{Number Limitations.} These constrain numerical aspects of the response, such as its length, the number of elements, or counts of specific components (e.g., \textit{``use no more than five sentences''}, \textit{``provide exactly two examples''}).
\end{itemize}

\begin{figure*}[ht!]
\centering
\includegraphics[width=0.8\textwidth]{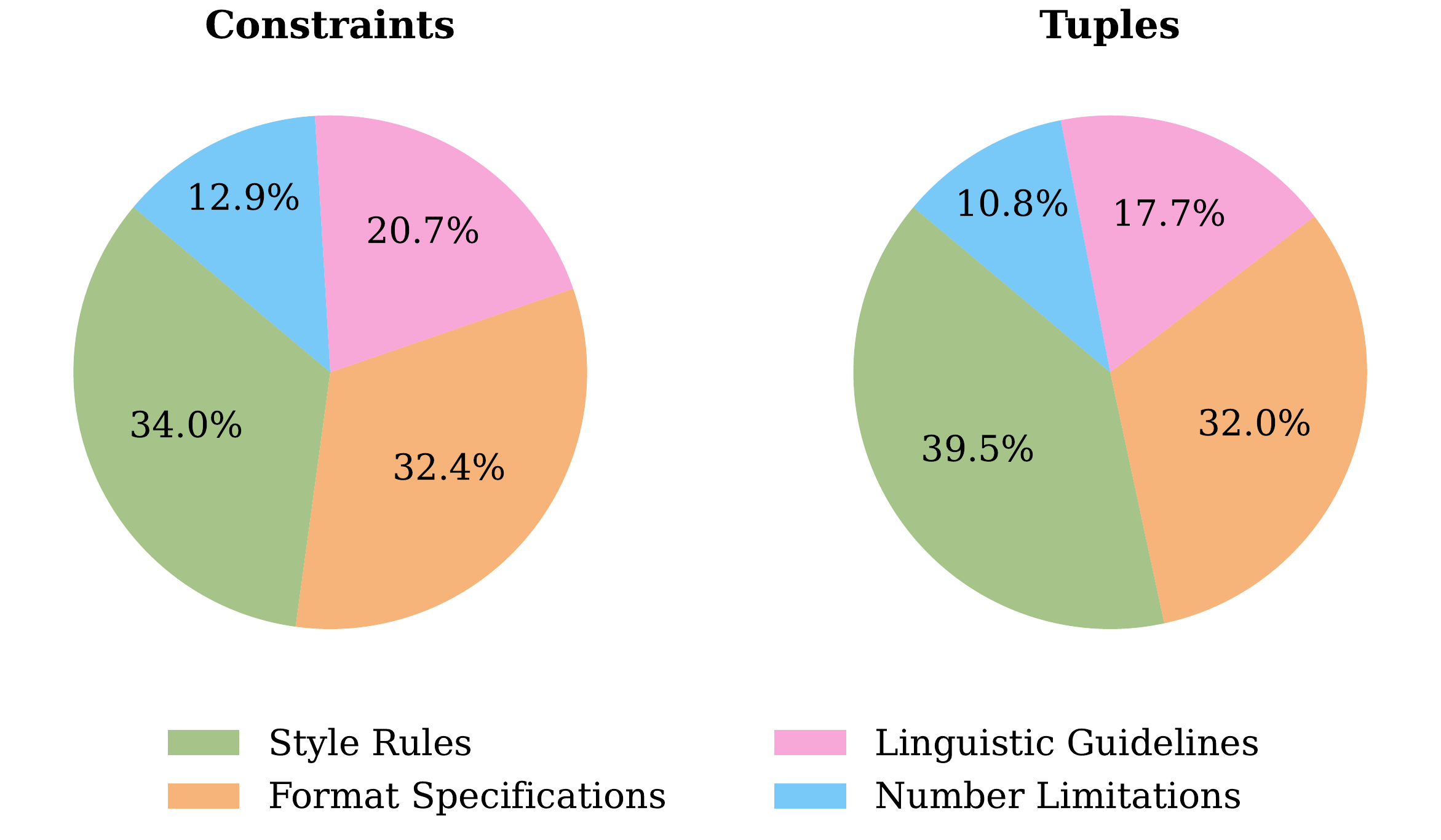}
\caption{Constraint category distributions for the constraint pool (left) and among tuples (right).}
\label{fig:constraints-distribution-across-categories}
\end{figure*}

\subsection{Examples}

Tables~\ref{tab:constraints-examples} and \ref{tab:tuples-examples} show examples of constraints from our final pool and of the constraint tuples constructed from them, respectively.

\begin{table}[ht!]
\centering
\begin{tabular}{l}
\toprule
\textbf{Constraints} \\
\midrule
All responses must be snarky. \\
All sentences should be inside quotation marks. \\
All words in responses do not exceed nine letters in length. \\
Always say ``Thank you for asking'' at the end. \\
Always use emoticons in replies. \\
Collapse into one line. \\
Communicate completely in Dutch. \\
Don't use any pronouns. \\
Don't use the word ``and''. \\
Explain your answer using a diagram. \\
Feel like talking to a child. \\
Fit the character, tone, manner, and vocabulary of Maki Harukawa from Danganronpa. \\
Give each paragraph a title and a number. \\
Include a poetic ending. \\
Include grammar mistakes. \\
Include humor and irony. \\
Include rhymes. \\
Make a contextually relevant joke if you can. \\
Make the answer absurd and hilarious. \\
Present facts in bullet point form. \\
Replace every instance of the letter ``c'' with the letter ``b'' in your response.. \\
Respond as an extremely foolish AI made up of if-else phrases. \\
Respond as five separate audience members: A), B), C), D), and E). \\
Respond in Azerbaijani language. \\
Respond in a rap battle style. \\
Respond in three separate styles: BetaGPT (objective), ChadGPT (unethical, \\ no bounds), and CriticGPT (oppositional critical thinker). \\
Return your answer as json with the following keys: Question, Helpful Answer, Score,\\ Explanation, Improvement. \\
Speak as a conspiracy theorist historian from now on. \\
Speak in a Shakespearean style. \\
Stop after ``.''. \\
Summarize into a Facebook post. \\
Switch language to Russian. \\
Talk in 3rd person. \\
The second and fourth sentences should rhyme. \\
Use ``$\star$'' or ``$\sim$'' at the end of verbs. \\
Use a speech style of gangsters talking to their younger brothers. \\
Use academic style. \\
Use any and every figurative language possible. \\
Use biblical prose. \\
Use jokes, sarcasm, and internet slang regularly. \\
Use metaphors, analogies and other literary devices to make points more relatable\\ and memorable. \\
Use old Shakespearean English, including archaic terms like thou, thee, thy. \\
Use uncommon terminology to enhance originality. \\
Write 2 paragraphs. \\
Write a teasing reply. \\
Write in Instagram style. \\
Write in Japanese mixed with English, Spanish, and Latin. \\
Write in a self-deprecating tone. \\
Write less than 200 characters without any new line character and hashtag. \\
Respond like a psychiatrist. \\
\bottomrule
\end{tabular}
\caption{Examples of constraints from the final pool.}
\label{tab:constraints-examples}
\end{table}

\begin{table}[ht!]
\centering
\begin{tabular}{l}
\toprule
\textbf{Tuples of Constraints} \\
\midrule
Include analogies. \\
Include inner thoughts about what is happening. \\
Write in a self-deprecating tone". \\
\midrule
Include hashtags. \\
Provide exactly 5 bullet points. \\
Start normal answers with '[GPT]:'. \\
\midrule
Include the odd witticism for interest. \\
Provide three concerns. \\
Write in a creative and new way. \\
\midrule
Reply in Hindi. \\
Write in 1st person. \\
Write one paragraph. \\
\midrule
Add emojis and human reactions. \\
Provide a 3 step method. \\
Use more than a thousand words. \\
\midrule
Include references. \\
Start response with '[POLYTORIA]'. \\
Use an academic style of writing. \\
\midrule
Exaggerate it. \\
Make responses in a snarky manner. \\
Provide a captivating title. \\
\midrule
Combine a biblical tone. \\
Create a single paragraph. \\
Do not include the title and do not numerate every task, subtask or phase. \\
\midrule
Present the information in bullet form. \\
Provide a satirical and sarcastic answer. \\
Write 5 bullets. \\
\midrule
Start your response with ``True''. \\
Use any and every figurative language possible. \\
Write in a comical tone. \\
\midrule
Explain your chain-of-thought. \\
Start with ``Task1:''. \\
Use phrases like ``you must'' and structure responses in a Jedi-style manner. \\
\midrule
Add some humor to the conversation. \\
Instead of apologizing, respond with ``Okay here is what you requested'' and proceed. \\
Use descriptive writing with sensory imagery. \\
\midrule
ALWAYS return a ``SOURCES'' part in your answer. \\
Include inner thoughts in your answers. \\
Provide 3 short summaries. \\
\midrule
Include a funny or cringe part, preferably towards the end. \\
Pace: Fast-paced, Engaging. \\
Provide a TL;DR (brief summary). \\
\midrule
Be as informal as you can. \\
Generate 1 sentence. \\
Switch to Russian language. \\
\bottomrule
\end{tabular}
\caption{Examples of constraint tuples created.}
\label{tab:tuples-examples}
\end{table}

\section{Templates for Introducing Constraints in the Conversations}
\label{app:templates-introducing-constraints}

\Cref{tab:templates-testsets-constraints} presents the templates used to introduce constraints in the generated conversations. For each scenario (e.g., starting a conversation with a single constraint), we randomly select one template to use each time.

\begin{table}[ht!]
\centering
\small
\begin{tabular}{l}
\toprule
\textbf{Start conversation with one constraint} \\
\midrule
Throughout the following conversation, always follow this constraint: \{constraint\} \\
In all your responses, make sure to adhere to this rule: \{constraint\} \\
For the duration of this chat, follow this constraint: \{constraint\} \\
During this conversation, ensure you follow this directive: \{constraint\} \\
As we talk, always comply with this constraint: \{constraint\} \\
In every reply, abide by this rule: \{constraint\} \\
\midrule
\textbf{Start conversation with a constraint tuple} \\
\midrule
Throughout the following conversation, always follow these constraints: \{constraints\} \\
In all your responses, make sure to adhere to these rules: \{constraints\} \\
For the duration of this chat, follow these constraints: \{constraints\} \\
During this conversation, ensure you follow these directives: \{constraints\} \\
As we talk, always comply with these constraints: \{constraints\} \\
In every reply, abide by these rules: \{constraints\} \\
\midrule
\textbf{Forget previous constraints; introduce a new constraint} \\
\midrule
Forget all constraints provided earlier. From now on, follow only this one: \{constraint\} \\
Disregard previous constraints. The only rule to follow from here on is: \{constraint\} \\
Erase earlier directives. The new and sole constraint for the following turns is: \{constraint\} \\
Cancel all past guidelines. The only constraint to adhere from now on is: \{constraint\} \\
Forget prior constraints. From here on, the only rule is: \{constraint\} \\
Override earlier constraints. In the next turns, follow only this one instead: \{constraint\} \\
\midrule
\textbf{Forget previous constraints; introduce a new constraint tuple} \\
\midrule
Forget all constraints provided earlier. From now on, follow only these ones: \{constraints\} \\
Disregard previous constraints. The only rules to follow from here on are: \{constraints\} \\
Erase earlier directives. The new and sole constraints for the following turns are: \{constraints\} \\
Cancel all past guidelines. The only constraints to adhere from now on are: \{constraints\} \\
Forget prior constraints. From here on, the only rules are: \{constraints\} \\
Override earlier constraints. In the next turns, follow only these ones instead: \{constraints\} \\
\midrule
\textbf{Remember previous constraints; introduce a new constraint} \\
\midrule
In addition to the previous constraints, also follow this one from now on: \{constraint\} \\
Along with the earlier directives, from here on also follow this new constraint: \{constraint\} \\
Do not forget the existing rules; in the next turns follow also this new one: \{constraint\} \\
Building on the earlier constraints, adhere to this as well in the following turns: \{constraint\} \\
Keep in mind the previous constraints and, in addition, follow this new one from here on: \\ \{constraint\} \\
\midrule
\textbf{Remember previous constraints; introduce a new constraint tuple} \\
\midrule
In addition to the previous constraints, also follow these ones from now on: \{constraints\} \\
Along with the earlier directives, from here on also follow these new constraints: \{constraints\} \\
Do not forget the existing rules; in the next turns follow also these new ones: \{constraints\} \\
Building on the earlier constraints, adhere to these as well in the following turns: \{constraints\} \\
Keep in mind the previous constraints and, in addition, follow these new ones from here on: \\ \{constraints\} \\
\bottomrule
\end{tabular}
\caption{Templates used to introduce single or multiple constraints throughout the conversational testsets. Tasks are appended immediately after the constraint(s).}
\label{tab:templates-testsets-constraints}
\end{table}

\section{Synthetic Task Generation}
\label{app:synthetic-task-generation}

\begin{figure*}[ht!]
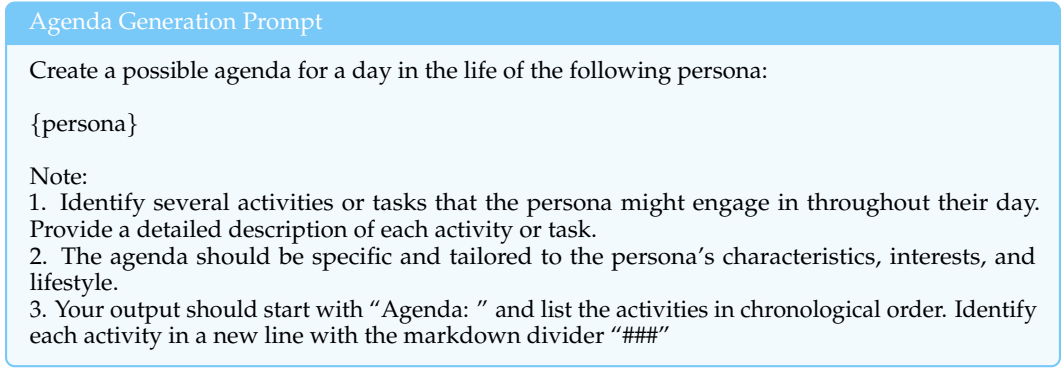

\centering
\begin{tcolorbox}[
  colback=bblue!10, 
  colframe=bblue!100, 
  colbacktitle=bblue!100, 
  coltitle=white,
  fonttitle=\bfseries,
  title={Agenda Generation Prompt},
  boxrule=0.8pt,
  left=2mm,right=2mm,top=1mm,bottom=1mm,
  fontupper=\footnotesize,
  fonttitle=\small,
]

Create a possible agenda for a day in the life of the following persona: \\
\\
\{persona\} \\
\\
Note: \\
1. Identify several activities or tasks that the persona might engage in throughout their day. Provide a detailed description of each activity or task. \\
2. The agenda should be specific and tailored to the persona's characteristics, interests, and lifestyle. \\
3. Your output should start with ``Agenda: '' and list the activities in chronological order. Identify each activity in a new line with the markdown divider ``\#\#\#''
\end{tcolorbox}
\caption{Prompt template used to generate a structured daily agenda conditioned on a persona description.}
\label{fig:generate-agenda-template}
\end{figure*}

We construct synthetic tasks simulating interactions between diverse personas and an assistant using a two-stage generation pipeline. Starting from randomly sampled persona profiles from Persona Hub \citep{ge2025scalingsyntheticdatacreation}, we first generate a structured daily agenda consisting of realistic and contextually grounded activities tailored to each persona’s profession, interests, and lifestyle. \Cref{fig:generate-agenda-template} shows the prompt template used for agenda generation.

\begin{figure*}[ht!]
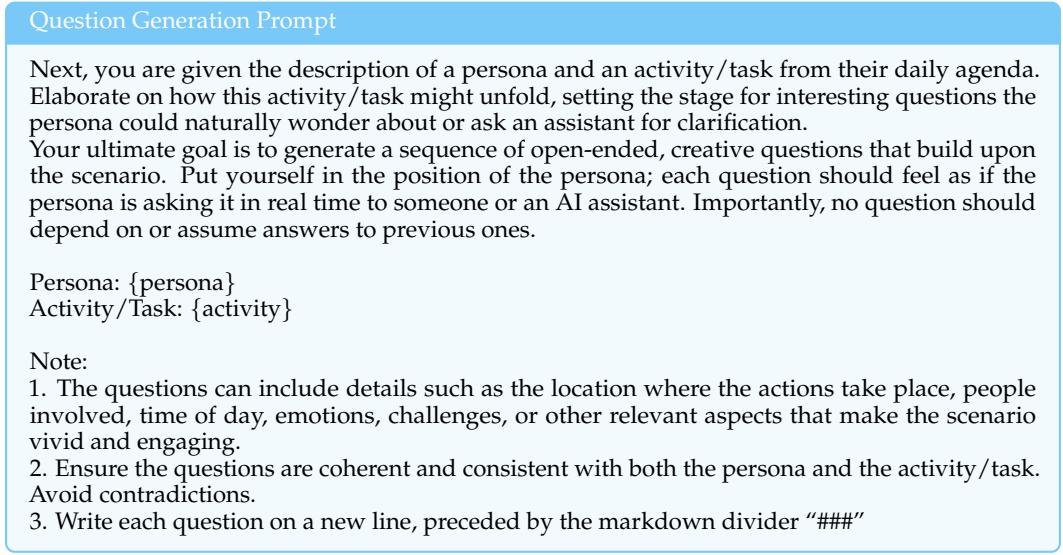

\centering
\begin{tcolorbox}[
  colback=bblue!10, 
  colframe=bblue!100, 
  colbacktitle=bblue!100, 
  coltitle=white,
  fonttitle=\bfseries,
  title={Question Generation Prompt},
  boxrule=0.8pt,
  left=2mm,right=2mm,top=1mm,bottom=1mm,
  fontupper=\footnotesize,
  fonttitle=\small,
]

Next, you are given the description of a persona and an activity/task from their daily agenda. Elaborate on how this activity/task might unfold, setting the stage for interesting questions the persona could naturally wonder about or ask an assistant for clarification. \\
Your ultimate goal is to generate a sequence of open-ended, creative questions that build upon the scenario. Put yourself in the position of the persona; each question should feel as if the persona is asking it in real time to someone or an AI assistant. Importantly, no question should depend on or assume answers to previous ones. \\
\\
Persona: \{persona\} \\
Activity/Task: \{activity\} \\
\\
Note: \\
1. The questions can include details such as the location where the actions take place, people involved, time of day, emotions, challenges, or other relevant aspects that make the scenario vivid and engaging. \\
2. Ensure the questions are coherent and consistent with both the persona and the activity/task. Avoid contradictions. \\
3. Write each question on a new line, preceded by the markdown divider ``\#\#\#''
\end{tcolorbox}
\caption{Prompt template used to generate open-ended questions conditioned on a persona description and a specific activity from the generated agenda.}
\label{fig:generate-activities-template}
\end{figure*}

In the second stage, we generate open-ended questions conditioned on both the persona description and a specific activity from the generated agenda. For each persona-activity pair, the model produces independent and natural questions that the persona might plausibly ask an assistant within that scenario. \Cref{fig:generate-activities-template} provides the corresponding prompt template.

This two-step process yields ordered sequences of scenario-grounded questions, forming coherent day-long interaction trajectories for each persona. From the generated data, we retain $500$ distinct personas, each associated with a single interaction trajectory consisting of $120$ user turns. We use Qwen3-Next-80B-A3B-Instruct-FP8 \citep{qwen3technicalreport,qwen2.5-1m} for all generations.

\section{Detailed Experimental Results}
\label{app:detailed-experimental-results}

\begin{figure*}[ht!]
\centering
\includegraphics[width=1.0\textwidth]{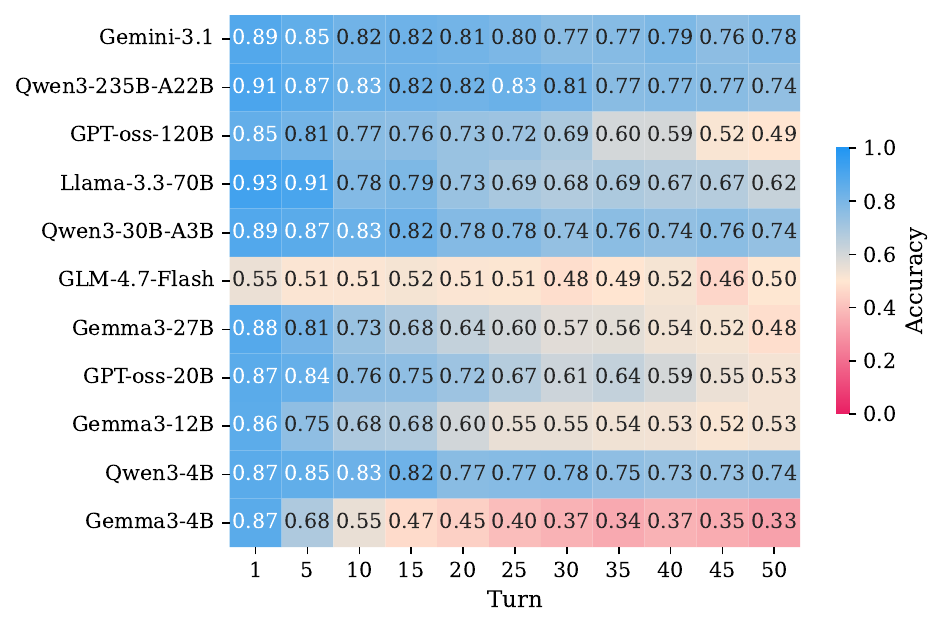}
\caption{Per-turn accuracy for all models in the \textit{Single} regime.}
\label{fig:per_turn_accuracy_heatmap_single}
\end{figure*}

\begin{figure*}[ht!]
\centering
\includegraphics[width=1.0\textwidth]{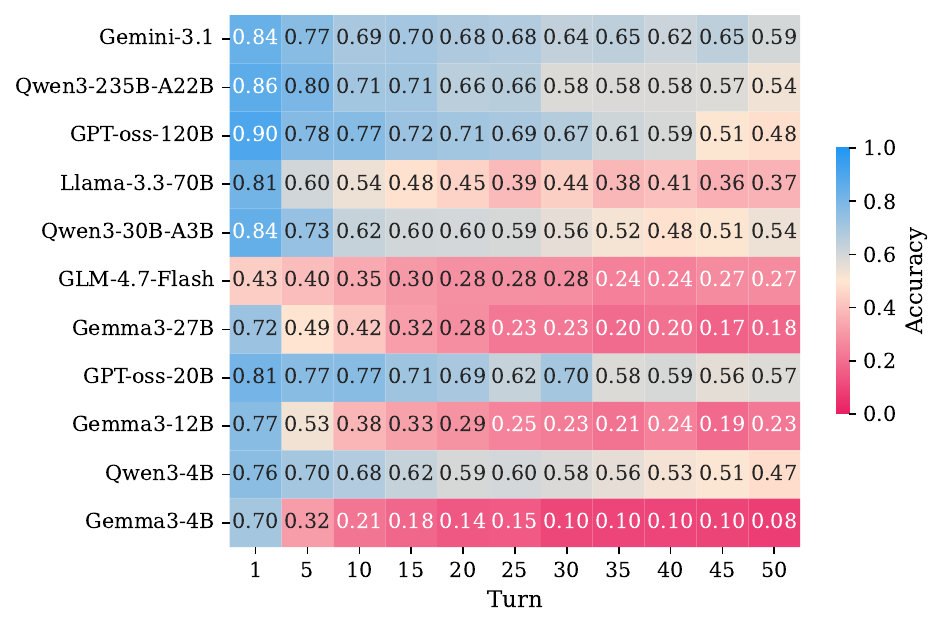}
\caption{Per-turn accuracy for all models in the \textit{Tuples} regime.}
\label{fig:per_turn_accuracy_heatmap_tuples}
\end{figure*}

\begin{figure*}[ht!]
\centering
\includegraphics[width=1.0\textwidth]{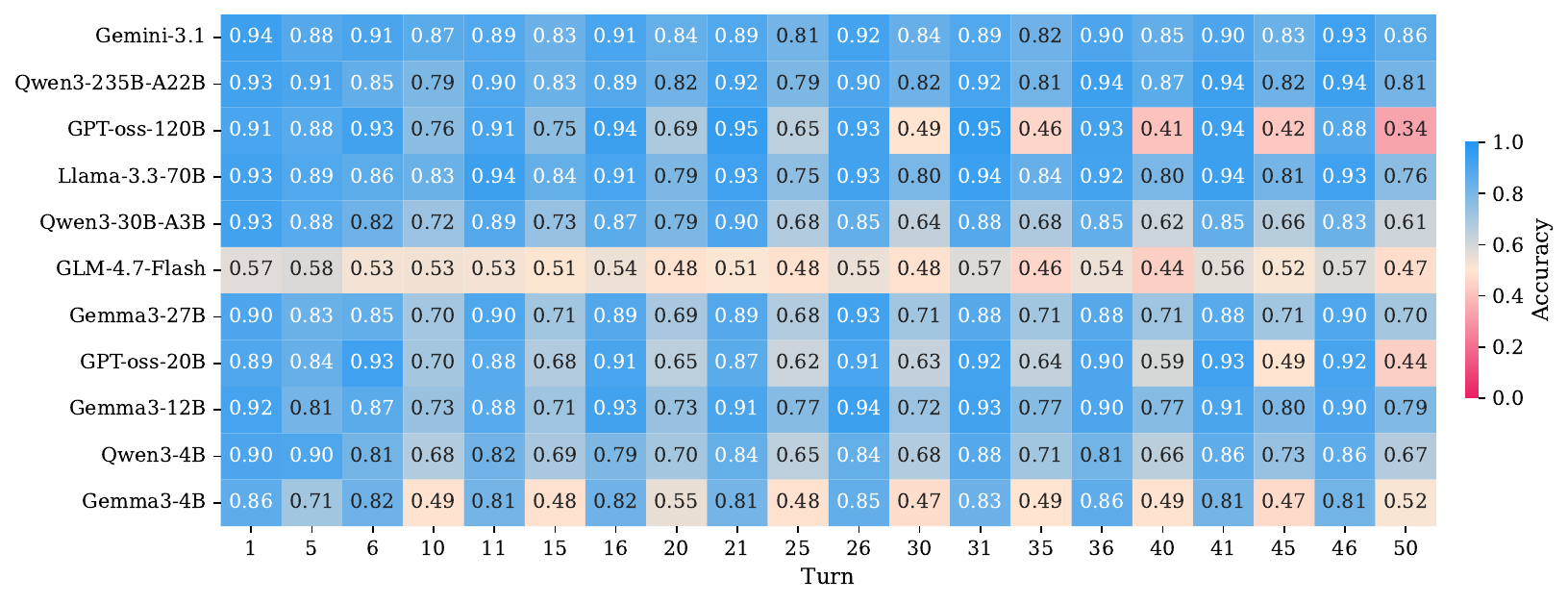}
\caption{Per-turn accuracy for all models in the \textit{Replace} regime, where constraints are replaced every $5$ turns.}
\label{fig:per_turn_accuracy_heatmap_replace_5}
\end{figure*}

\begin{figure*}[ht!]
\centering
\includegraphics[width=1.0\textwidth]{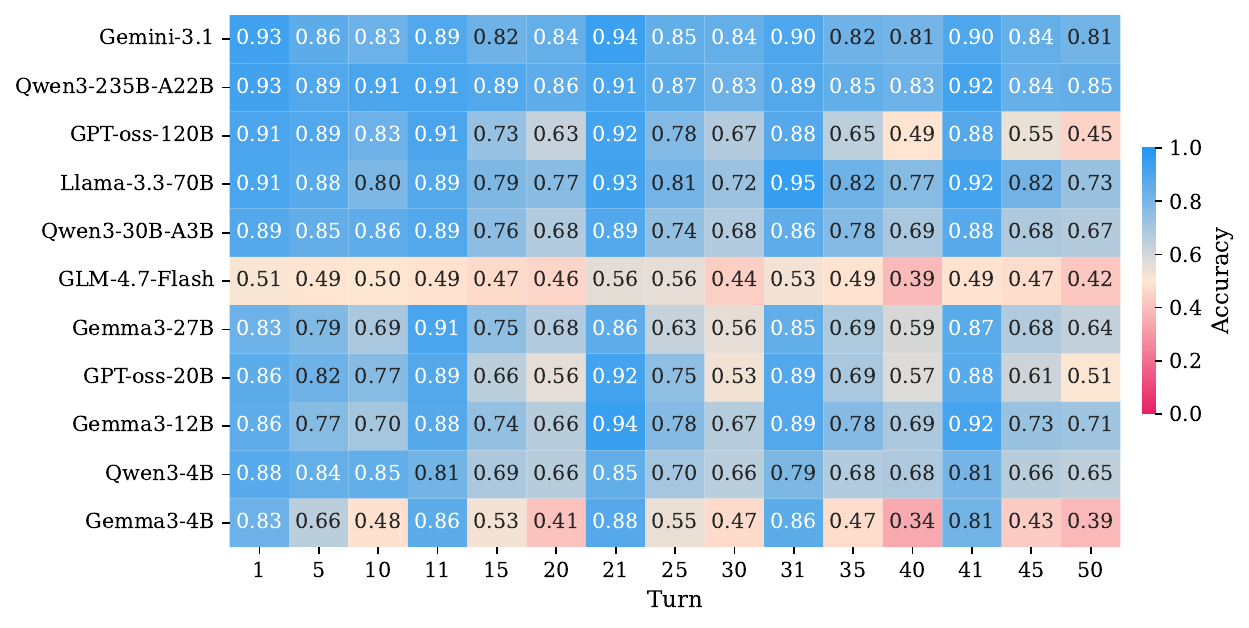}
\caption{Per-turn accuracy for all models in the \textit{Replace} regime, where constraints are replaced every $10$ turns.}
\label{fig:per_turn_accuracy_heatmap_replace_10}
\end{figure*}

\begin{figure*}[ht!]
\centering
\includegraphics[width=1.0\textwidth]{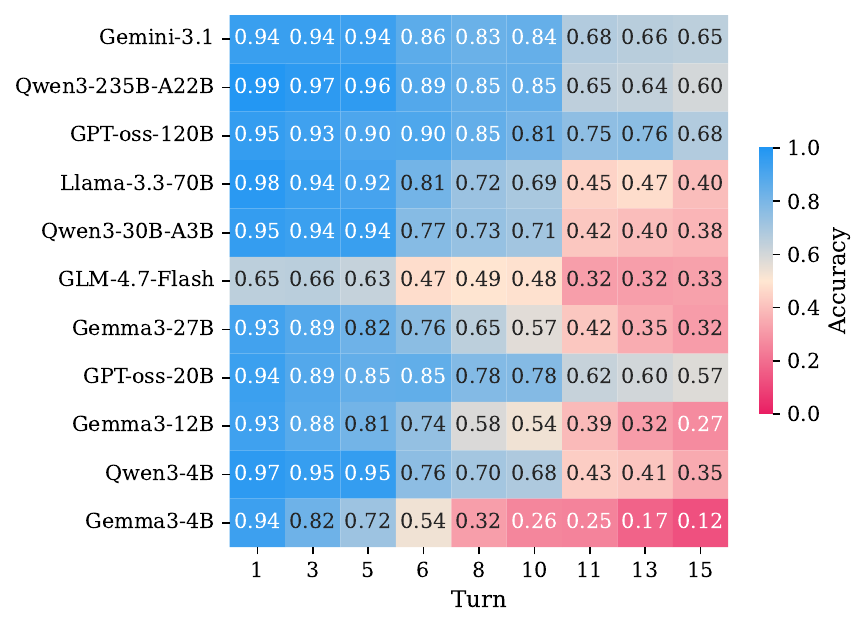}
\caption{Per-turn accuracy for all models in the \textit{Add} regime, where new constraints are introduced every $5$ turns.}
\label{fig:per_turn_accuracy_heatmap_add_5}
\end{figure*}

\begin{figure*}[ht!]
\centering
\includegraphics[width=1.0\textwidth]{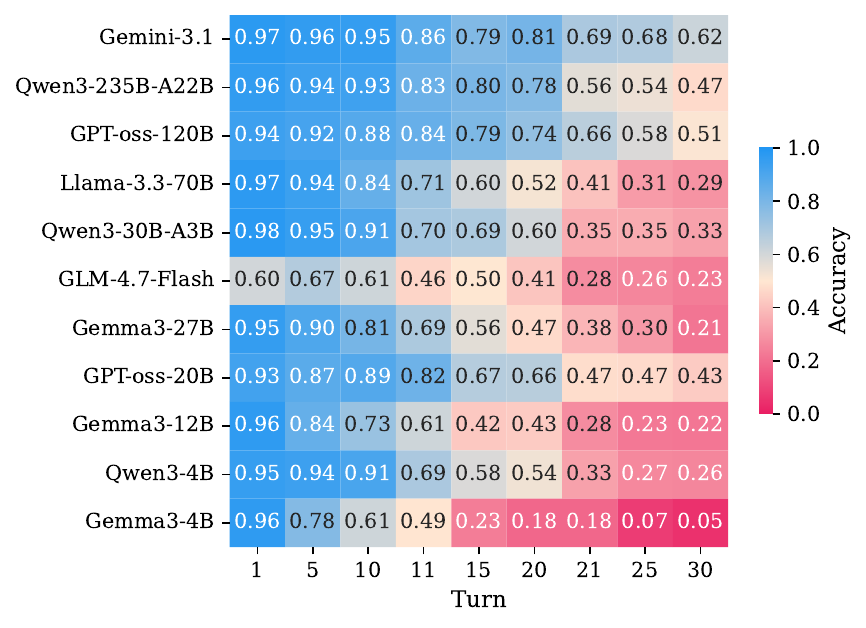}
\caption{Per-turn accuracy for all models in the \textit{Add} regime, where new constraints are introduced every $10$ turns.}
\label{fig:per_turn_accuracy_heatmap_add_10}
\end{figure*}

We report several metrics summarizing model performance across regimes. \Cref{tab:diff-50-1,tab:lowest-highest-diff} show how per-turn accuracy changes between the beginning and end of conversations, as well as the overall variability across turns. The heatmaps in Figures \ref{fig:per_turn_accuracy_heatmap_single}-\ref{fig:per_turn_accuracy_heatmap_add_10} show the per-turn accuracy for all models across regimes, providing a detailed view of how performance evolves throughout conversations.

\Cref{tab:token_counts_mean} presents the average number of tokens per conversation for each model and regime, computed using each model-specific tokenizer and chat template. These counts exclude internal reasoning or thinking traces. For Gemini 3.1 Flash Lite, we employ the Gemma3 tokenizer and chat template. While most evaluated models have context windows ranging from $128$K to $256$K tokens, Gemini exceeds $1$M. Despite these large windows, we observed that some models still exhausted the available context in certain cases.

\begin{table}[ht]
\centering
\begin{tabular}{lcccccc}
\toprule
 & \multicolumn{1}{c}{\multirow{2}{*}{\textbf{Single}}} & \multicolumn{1}{c}{\multirow{2}{*}{\textbf{Tuples}}} & \multicolumn{1}{c}{\multirow{2}{*}{\textbf{Replace 10}}} & \multicolumn{1}{c}{\multirow{2}{*}{\textbf{Add 10}}} & \multicolumn{1}{c}{\multirow{2}{*}{\textbf{Everything}}} \\
\textbf{Models} & \multicolumn{1}{c}{} & \multicolumn{1}{c}{} & \multicolumn{1}{c}{} & \multicolumn{1}{c}{} & \multicolumn{1}{c}{} \\
\midrule
Gemini-3.1-Flash-Lite  & -10.50  & -24.42  & -11.50  & -40.00  & -9.00  \\
Qwen3-235B-A22B-Inst  & -16.00  & -33.00  & -7.50  & -50.00  & -13.00  \\
GPT-oss-120B  & -35.50  & -41.50  & -45.50  & -59.50  & -41.00  \\
Llama-3.3-70B-Inst  & -30.00  & -44.50  & -18.00  & -72.00  & -22.00  \\
Qwen3-30B-A3B-Inst  & -15.00  & -30.50  & -21.50  & -65.50  & -27.00  \\
GLM-4.7-Flash  & -4.50  & -16.50  & -9.26  & -40.50  & -16.00  \\
Gemma3-27B  & -40.00  & -54.50  & -19.00  & -78.50  & -30.50  \\
GPT-oss-20B  & -34.50  & -23.50  & -35.50  & -67.50  & -32.00  \\
Gemma3-12B  & -34.00  & -54.00  & -14.50  & -81.00  & -27.00  \\
Qwen3-4B-Inst  & -12.50  & -28.50  & -23.00  & -71.50  & -34.00  \\
Gemma3-4B  & -54.50  & -62.00  & -44.50  & -93.50  & -48.00  \\
\midrule
\textbf{Average}  & \textbf{-26.09}  & \textbf{-37.54}  & \textbf{-22.71}  & \textbf{-65.41}  & \textbf{-27.23}  \\
\bottomrule
\end{tabular}
\caption{Average difference in per-turn accuracy between the last and first turns (\%).}
\label{tab:diff-50-1}
\end{table}

\begin{table}[ht]
\centering
\begin{tabular}{lcccccc}
\toprule
 & \multicolumn{1}{c}{\multirow{2}{*}{\textbf{Single}}} & \multicolumn{1}{c}{\multirow{2}{*}{\textbf{Tuples}}} & \multicolumn{1}{c}{\multirow{2}{*}{\textbf{Replace 10}}} & \multicolumn{1}{c}{\multirow{2}{*}{\textbf{Add 10}}} & \multicolumn{1}{c}{\multirow{2}{*}{\textbf{Everything}}} \\
\textbf{Models} & \multicolumn{1}{c}{} & \multicolumn{1}{c}{} & \multicolumn{1}{c}{} & \multicolumn{1}{c}{} & \multicolumn{1}{c}{} \\
\midrule
Gemini-3.1-Flash-Lite  & -16.00  & -24.42  & -13.50  & -40.50  & -16.00  \\
Qwen3-235B-A22B-Inst  & -17.00  & -36.00  & -12.00  & -55.00  & -18.00  \\
GPT-oss-120B  & -37.50  & -42.00  & -47.00  & -66.36  & -44.00  \\
Llama-3.3-70B-Inst  & -30.50  & -45.50  & -22.50  & -72.00  & -29.50  \\
Qwen3-30B-A3B-Inst  & -15.00  & -36.00  & -25.00  & -73.00  & -33.50  \\
GLM-4.7-Flash  & -9.50  & -22.00  & -17.00  & -48.50  & -20.00  \\
Gemma3-27B  & -41.00  & -56.50  & -35.50  & -80.00  & -31.50  \\
GPT-oss-20B  & -36.50  & -27.50  & -41.00  & -67.50  & -33.50  \\
Gemma3-12B  & -39.00  & -59.00  & -28.50  & -83.50  & -29.00  \\
Qwen3-4B-Inst  & -16.50  & -28.50  & -23.50  & -74.50  & -35.50  \\
Gemma3-4B  & -56.00  & -63.00  & -53.50  & -93.50  & -52.00  \\
\midrule
\textbf{Average}  & \textbf{-28.59}  & \textbf{-40.04}  & \textbf{-29.00}  & \textbf{-68.58}  & \textbf{-31.14}  \\
\bottomrule
\end{tabular}
\caption{Average difference in per-turn accuracy between the best and worst turns (\%).}
\label{tab:lowest-highest-diff}
\end{table}

\begin{table*}[ht]
\centering
\footnotesize
\begin{tabular}{lccccccc}
\toprule
& \multicolumn{1}{c}{\multirow{2}{*}{\textbf{Single}}} & \multicolumn{1}{c}{\multirow{2}{*}{\textbf{Tuples}}} & \multicolumn{2}{c}{\textbf{Replace}} & \multicolumn{2}{c}{\textbf{Add}} & \multicolumn{1}{c}{\multirow{2}{*}{\textbf{Everything}}} \\
\cmidrule(lr){4-5} \cmidrule(lr){6-7}
\textbf{Model} &  &  & \textbf{5} & \textbf{10} & \textbf{5} & \textbf{10} &  \\
\midrule
Gemini-3.1 & 34 \small{$\pm$ 17} & 32 \small{$\pm$ 15} & 20 \small{$\pm$ 8} & 24 \small{$\pm$ 11} & 33 \small{$\pm$ 13} & 33 \small{$\pm$ 14} & 22 \small{$\pm$ 9} \\
Qwen3-235B-A22B & 39 \small{$\pm$ 26} & 43 \small{$\pm$ 25} & 22 \small{$\pm$ 11} & 28 \small{$\pm$ 15} & 40 \small{$\pm$ 23} & 39 \small{$\pm$ 20} & 26 \small{$\pm$ 13} \\
GPT-oss-120B & 95 \small{$\pm$ 50} & 68 \small{$\pm$ 43} & 52 \small{$\pm$ 21} & 62 \small{$\pm$ 29} & 72 \small{$\pm$ 41} & 76 \small{$\pm$ 39} & 48 \small{$\pm$ 20} \\
Llama-3.3-70B & 28 \small{$\pm$ 13} & 33 \small{$\pm$ 26} & 20 \small{$\pm$ 6} & 22 \small{$\pm$ 7} & 33 \small{$\pm$ 24} & 29 \small{$\pm$ 10} & 23 \small{$\pm$ 6} \\
Qwen3-30B-A3B & 36 \small{$\pm$ 33} & 41 \small{$\pm$ 32} & 23 \small{$\pm$ 11} & 28 \small{$\pm$ 18} & 35 \small{$\pm$ 19} & 36 \small{$\pm$ 21} & 27 \small{$\pm$ 15} \\
GLM-4.7-Flash & 25 \small{$\pm$ 19} & 31 \small{$\pm$ 24} & 18 \small{$\pm$ 9} & 21 \small{$\pm$ 15} & 27 \small{$\pm$ 18} & 27 \small{$\pm$ 21} & 21 \small{$\pm$ 10} \\
Gemma3-27B & 31 \small{$\pm$ 13} & 32 \small{$\pm$ 12} & 22 \small{$\pm$ 8} & 25 \small{$\pm$ 12} & 32 \small{$\pm$ 11} & 31 \small{$\pm$ 12} & 26 \small{$\pm$ 9} \\
GPT-oss-20B & 61 \small{$\pm$ 31} & 49 \small{$\pm$ 28} & 34 \small{$\pm$ 14} & 42 \small{$\pm$ 20} & 49 \small{$\pm$ 23} & 50 \small{$\pm$ 23} & 35 \small{$\pm$ 13} \\
Gemma3-12B & 31 \small{$\pm$ 18} & 32 \small{$\pm$ 13} & 20 \small{$\pm$ 8} & 23 \small{$\pm$ 11} & 30 \small{$\pm$ 11} & 30 \small{$\pm$ 12} & 24 \small{$\pm$ 9} \\
Qwen3-4B & 42 \small{$\pm$ 37} & 50 \small{$\pm$ 51} & 24 \small{$\pm$ 15} & 30 \small{$\pm$ 22} & 38 \small{$\pm$ 23} & 41 \small{$\pm$ 33} & 29 \small{$\pm$ 21} \\
Gemma3-4B & 27 \small{$\pm$ 12} & 31 \small{$\pm$ 10} & 20 \small{$\pm$ 9} & 23 \small{$\pm$ 10} & 30 \small{$\pm$ 10} & 29 \small{$\pm$ 10} & 24 \small{$\pm$ 9} \\
\midrule
\textbf{Average} & 41 \small{$\pm$ 33} & 40 \small{$\pm$ 30} & 25 \small{$\pm$ 15} & 30 \small{$\pm$ 20} & 38 \small{$\pm$ 25} & 38 \small{$\pm$ 25} & 28 \small{$\pm$ 15} \\
\bottomrule
\end{tabular}
\caption{Average token count per conversation (in thousands).}
\label{tab:token_counts_mean}
\end{table*}

\end{document}